\def\year{2022}
\documentclass[letterpaper]{article}
\usepackage{aaai}
\usepackage{array}
\usepackage{times}
\usepackage{color}
\usepackage{hyperref}
\usepackage{helvet}
\usepackage{courier}
\usepackage{epsfig}
\usepackage{graphicx}
\usepackage{amsmath}
\usepackage{amssymb}
\usepackage{caption}
\usepackage{subcaption}
\usepackage{subfiles}
\usepackage{multirow}
\usepackage{caption}
\usepackage{seqsplit}
\usepackage{url}
\usepackage{booktabs}
\usepackage{algorithmic}
\usepackage{algorithm}
\usepackage{threeparttable}
\usepackage{tablefootnote}
\usepackage{caption}
\usepackage{bm}
\usepackage{arydshln}

\begin{document}

\title{\emph{GuidedMix-Net}: Semi-Supervised Semantic Segmentation by Using Labeled Images as Reference}

\author{
    Peng Tu\thanks{Equal Contribution}\textsuperscript{\rm 1, \rm 2},
    Yawen Huang\footnotemark[1]\textsuperscript{\rm 3},
    Zheng Feng\thanks{Corresponding Author}\textsuperscript{\rm 1},
    Zhenyu He\textsuperscript{\rm 4},
    Liujuan Cao\textsuperscript{\rm 5}, 
    Ling Shao\textsuperscript{\rm 6} \\
    \textsuperscript{\rm 1}Southern University of Science and Technolog, Shenzhen, China \quad
    \textsuperscript{\rm 2}Shenzhen Microbt Electronics Technology Co., Ltd, China \\
    \textsuperscript{\rm 3}Tencent Jarvis Lab, Shenzhen, China \quad
    \textsuperscript{\rm 4}Harbin Institute of Technology, Shenzhen, China \quad
    \textsuperscript{\rm 5}Xiamen University, Xiamen, China \\
    \textsuperscript{\rm 6}National Center for Artificial Intelligence, Saudi Data and Artificial Intelligence Authority, Riyadh, Saudi Arabia \\
    \texttt{yh.peng.tu@gmail.com, yawenhuang@tencent.com, zfeng02@gmail.com} \\
    \texttt{zhenyuhe@hit.edu.cn, caoliujuan@xmu.edu.cn, ling.shao@ieee.org} 
    }
\maketitle

\begin{abstract}
\begin{quote}
Semi-supervised learning is a challenging problem which aims to construct a model by learning from limited labeled examples.
Numerous methods for this task focus on utilizing the predictions of unlabeled instances consistency alone to regularize networks.
However, treating labeled and unlabeled data separately often leads to the discarding of mass prior knowledge learned from the labeled examples. 
In this paper, we propose a novel method for semi-supervised semantic segmentation named GuidedMix-Net, by leveraging labeled information to guide the learning of unlabeled instances.
Specifically, GuidedMix-Net employs three operations: 1) interpolation of similar labeled-unlabeled image pairs; 2) transfer of mutual information; 3) generalization of pseudo masks.
It enables segmentation models can learning the higher-quality pseudo masks of unlabeled data by transfer the knowledge from labeled samples to unlabeled data.
Along with supervised learning for labeled data, the prediction of unlabeled data is jointly learned with the generated pseudo masks from the mixed data. 
Extensive experiments on PASCAL VOC 2012, and Cityscapes demonstrate the effectiveness of our GuidedMix-Net, which achieves competitive segmentation accuracy and significantly improves the mIoU by +7$\%$ compared to previous approaches.
\end{quote}
\end{abstract}

\section{Introduction}
The past several years have witnessed the success of convolutional neural networks (CNNs) \cite{long2015fully,ronneberger2015u,huang2020mcmt,chen2017deeplab,huang2021brain} for visual semantic segmentation.
Although data-driven deep learning techniques have benefitted greatly from the availability of large-scale image datasets, they require dense and precise pixel-level annotations for parameter learning. 
Alternative learning strategies, such as semi-supervision, have thus emerged as promising approaches to reduce the need for annotations, requiring simpler or fewer labels for image classification \cite{tarvainen2017mean,2018Virtual,sohn2020fixmatch}.

\begin{figure}[t]
\centering
\subfloat{\includegraphics[width=7.6cm, height=3.5cm]{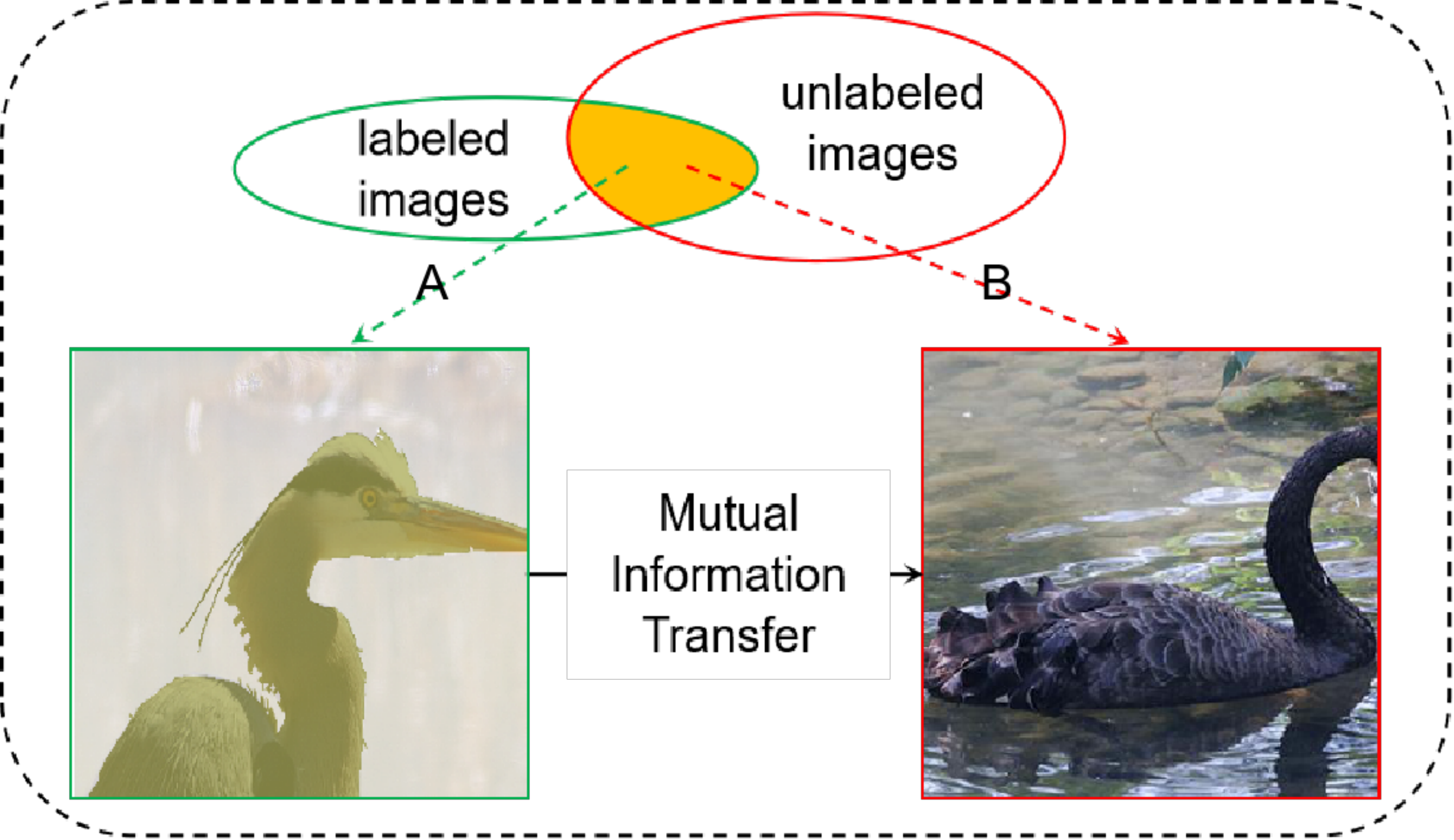}}\\\vspace{0.5mm}
\subfloat{\includegraphics[width=2cm, height=1.3cm]{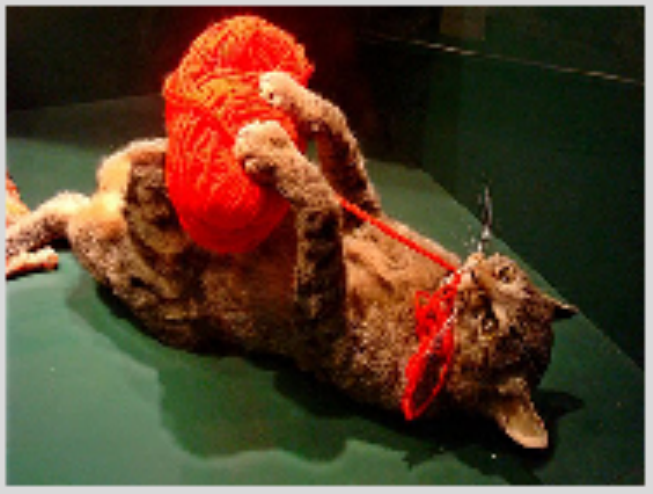}}\hspace{0.1mm}
\subfloat{\includegraphics[width=2cm, height=1.3cm]{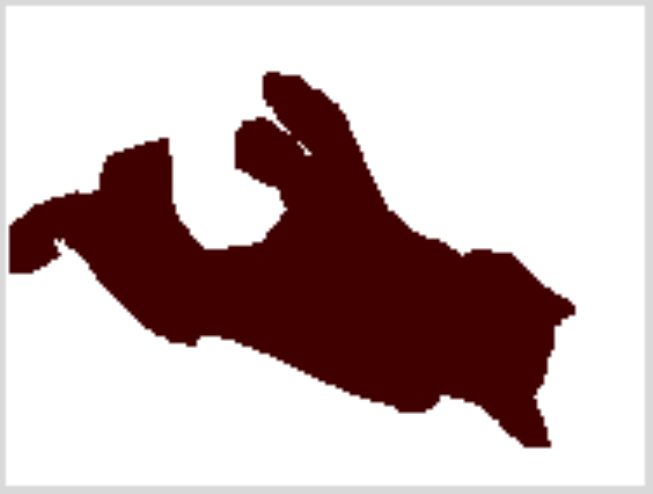}}\hspace{0.1mm}
\subfloat{\includegraphics[width=2cm, height=1.3cm]{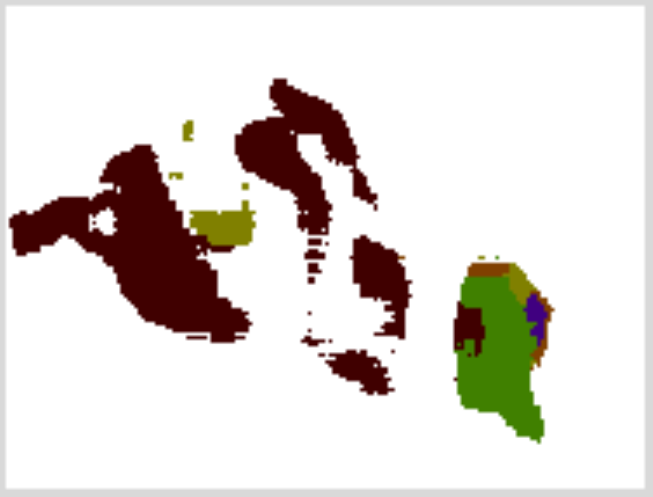}}\hspace{0.1mm}
\subfloat{\includegraphics[width=2cm, height=1.3cm]{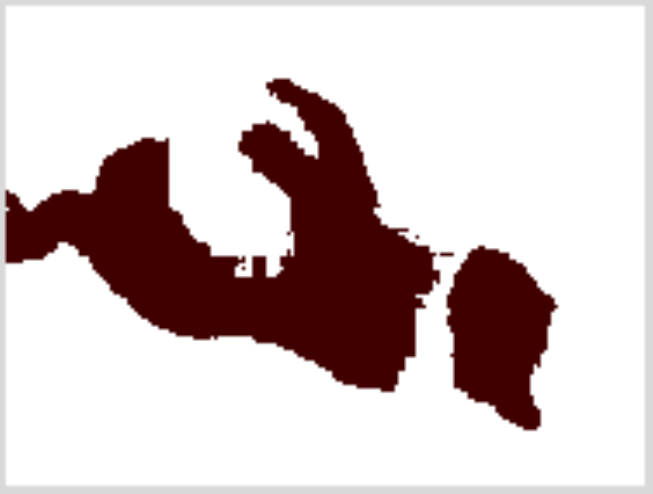}}\\\vspace{0.1mm}
\setcounter{subfigure}{0}
\subfloat[Input]{\includegraphics[width=2cm, height=1.3cm]{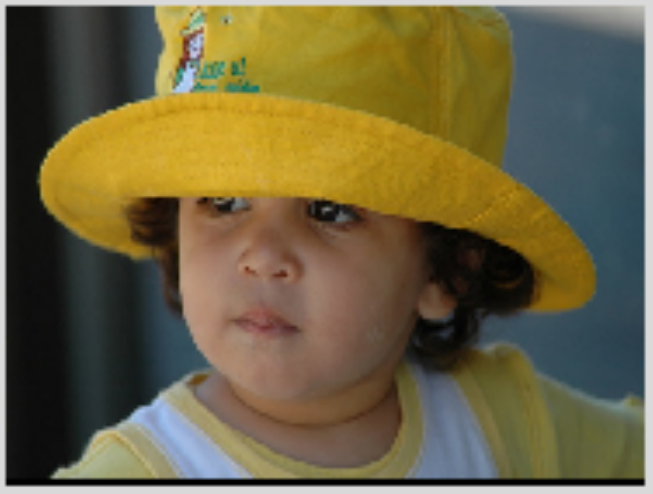}}\hspace{0.1mm}
\subfloat[GTs]{\includegraphics[width=2cm, height=1.3cm]{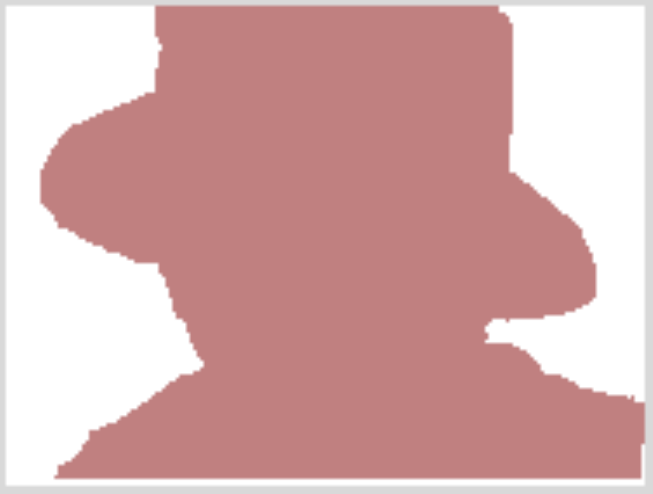}}\hspace{0.1mm}
\subfloat[CCT]{\includegraphics[width=2cm, height=1.3cm]{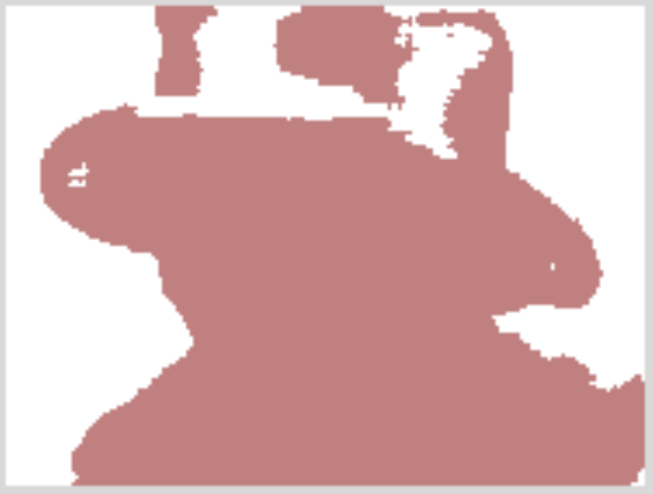}}\hspace{0.1mm}
\subfloat[Ours]{\includegraphics[width=2cm, height=1.3cm]{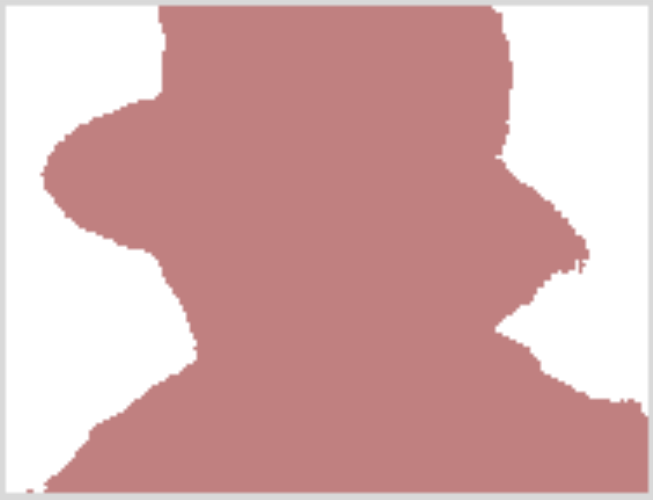}}\vspace{-1mm}
\caption{
\textbf{Upper} (dotted box): The mutual information transfer module selects similar features to transfer knowledge from the labeled samples to unlabeled images. 
\textbf{Lower}: Examples of ground-truths (GTs) (b), the pseudo mask of CCT (c) and ours (d). }
\label{MDExem}
\end{figure}
Recent semi-supervised methods for semantic segmentation, such as \cite{ouali2020semi,luo2020semi,french2020semi,olsson2021classmix}, exploit consistency by perturbing the unlabeled samples to regularize model training. 
Intuitively, the models are expected to manifest the invariance underlying any small perturbations while observing the natural properties of the data, especially for unlabeled data.
To address this, numerous semi-supervised semantic segmentation methods have been proposed against the perturbations when leveraging unlabeled samples. 
For example, CCT \cite{ouali2020semi} introduces random manual perturbations by designing separated, unrelated decoders for each type of perturbation. 
DTC \cite{luo2020semi} builds a task-level regularization rather than data-level perturbation. 
CutMix \cite{french2020semi} and ClassMix \cite{olsson2021classmix} follow MixUp \cite{zhang2017mixup} and achieve semi-supervised segmentation by forcing the predictions for the augmented and original data to be consistent.

Although various approaches have been introduced over the years, a key bottleneck of semi-supervised segmentation models is that they typically treat the labeled and unlabeled samples separately during training. 
Existing methods focus on how to use unlabeled data alone under various manual perturbations. 
On the one hand, although low-level perturbations do improve the robustness slightly, the rich underlying intrinsic information of the unlabeled instances has not yet been fully explored. 
For instance, consistency-based methods rely primarily on the local information of the samples themselves by constraining the local smoothness. 
This strategy cannot comprehensively mine the structural information, especially for unlabeled data, which in turn causes the model to produce a suboptimal solution. 
Another clear weakness of consistency-based methods is that they require abundant and diverse perturbations, which are expensive and time-consuming to obtain. 
For instance, CCT \cite{ouali2020semi} incorporates almost thirty decoders to make the learned model adequately robust. 
The example results shown in Fig. \ref{MDExem} (c) also confirm that generating pseudo masks using simple perturbation consistency training yields significant deviation in the contour and semantic understanding of objects.
In other words, the massive amount of prior information learned from the labeled samples cannot be transferred to the unlabeled data. 
Current semi-supervised semantic segmentation methods provide inconsistent optimization objectives for labeled and unlabeled data in different training stages, where the labeled samples are used to improve the discriminative ability and the unlabeled samples enhance the smoothness of models. 
However, we should be able to train a uniform model by leveraging a large amount of unlabeled data under the guidance of labeled samples.
This will enable the learned representations to be refine, thus facilitating mutual information interaction and transfer. 
We also note that humans can recognize unfamiliar objects subconsciously, by making inferences based on similar or recognizable objects. 
For example, in Fig. \ref{MDExem}, the dotted box provides a labeled image $A$ and an unlabeled image $B$ with similar objects. 
Most people can recognize and segment the object in image $B$ by transferring their knowledge of image $A$ to image $B$. 
In contrast, existing deep models are typically trained with limited labeled samples and most directly generate the pseudo mask of image $B$, making it difficult to produce a high-quality prediction for unseen samples, as shown in the bottom of Fig. \ref{MDExem} (c). 
The example shown in Fig. \ref{MDExem} is a relatively simple scenario; natural images are usually far more complex with, for example, multiple occluded objects, making them even more challenging to segment.
Another observation is that similar objects (${e.g.}$ intra-class objects) often contain common edges and textures.
An intuitive way to improve the segmentation of unlabeled data is therefoce to refer to labeled images, as humans do. 

Motivated by these problems, we propose a novel semi-supervised method for semantic segmentation, named GuidedMix-Net. 
GuidedMix-Net allows knowledge to be transferred from the labeled images to the unlabeled samples, as occur in the human cognitive path. 
To learn from the unlabeled samples, GuidedMix-Net employs three processes, ${i.e.}$, labeled-unlabeled image pair interpolation, mutual information transfer, and pseudo mask generation. 
Specifically, we feed pairs of labeled and unlabeled images as input into the model and carry out a linear interpolation of them to capture pairwise interactions.
Then, we learn the uniform feature vectors from the mixed data to inherit different contexts from the image pairs.
To incorporate non-local blocks \cite{Wang2018NonlocalNN} into the mixed feature layer, long-range dependencies are explored both within images and between them to mine similar object patterns and learn semantic correlations. 
We further select objects with similar features to ensure that the cues will be similar for different image pairs.
Feature selection improves the prediction and mask qualities of the unlabeled images by using the supervised information from the labeled images as reference.
After that, we decouple the hybrid prediction to obtain a pseudo mask for the unlabeled image. 
As a result, the generated pseudo masks are more credible than the direct predictions of unlabeled samples.
Finally, the pairs can be utilized for self-training to explore the rich underlying semantic structures provided by the unlabeled examples and further improve the performance of our model.


\section{Related Work}

\subsection{Semi-Supervised Classification} 
Semi-supervised classification methods \cite{2016Regularization,2017Mean,2018Virtual} typically focus on achieving consistent training by combining a standard supervised loss (\textit{e.g.} cross-entropy loss) and an unsupervised consistency loss to encourage consistent predictions for perturbations on the unlabeled samples. 
Randomness is essential for machine learning to either guarantee the generalization and robustness of the model or provides multiple different predictions for the same input. 
Based on this, Sajjadi et al. \cite{2016Regularization} introduced an unsupervised loss function which leverages the stochastic property of randomized data augmentation, dropout and random max-pooling to minimize the difference between the predictions of multiple passes of a training sample through the network.
Although these random augmentation techniques can improve the performance, they still remain difficulty on providing effective constraints for boundaries.
Miyato et al.~\cite{2018Virtual} after proposed a virtual adversarial training scheme to achieve smooth regularization. 
Their method aims to perturb the decision boundary of the model via a virtual adversarial loss-based regularization to measure the local smoothness of the conditional label distribution.

\subsection{Semi-Supervised Semantic Segmentation} 
Semi-supervised semantic segmentation algorithms have achieved great success in recent years \cite{2016Weakly,2018Revisiting,2017Semi,2019FickleNet,2018Adversarial}. 
For example, EM-Fixed \cite{2016Weakly} provides a novel online expectation-maximization method by training from either weakly annotated data such as bounding boxes, image-level labels, or a combination of a few strongly labeled and many weakly labeled images, sourced from different datasets. 
EM-Fixed benefits from the use of both a small amount of labeled and large amount of unlabeled data, achieving competitive results even against other fully supervised methods. 
From a certain point of view, EM-Fixed involves weak labels for training and hence is not a pure semi-supervised method. 
Spampinato et al. \cite{2017Semi} designed a semi-supervised semantic segmentation method using limited labeled data and abundant unlabeled data in a generative adversarial network (GAN). 
Their model uses discriminators to estimate the quality of predictions for unlabeled data.
If the quality score is high, the pseudo-label generated from the prediction can be regarded as the ground-truth, and the model is optimized by calculating the cross-entropy loss. 
However, models trained on a limited amount of labeled data typically fail in the following ways:
1) they generate inaccurate low-level details; 2) they misinterpret high-level information. 
To address these problems, s4GAN-MLMT \cite{2019Semi} fuses a GAN-based branch and a classifier to discriminate the generated segmentation maps. 
However, considering the intrinsic training difficulty of GANs, some semi-supervised image classification approaches instead adopt a consistent training strategy to ensure similar outputs under small changes, since this is flexible and easy to implement.
CCT \cite{ouali2020semi} employs such a training scheme for semi-supervised semantic segmentation, where the invariance of the predictions is enforced over different perturbations applied to the outputs of the decoder.
Specifically, a shared encoder and a main decoder are trained in a supervised manner using a few labeled examples.
To leverage unlabeled data, CCT enforces consistency between the main decoder predictions and the other set of decoders (for each type of perturbations), and uses different perturbations output from the encoders as input to improve the representations.

Unlike previous methods, which primarily focus on learning unlabeled data, we use the labeled images as references and transfer their knowledge to guide the learning of effective information from unlabeled data.
As a result, the proposed method generate high-quality features from the labeled images, and refines features of unlabeled data via pairwise interaction.

\section{GuidedMix-Net}
\begin{figure*}[t]
    \centering
    \includegraphics[width=11cm, height=6.5cm]{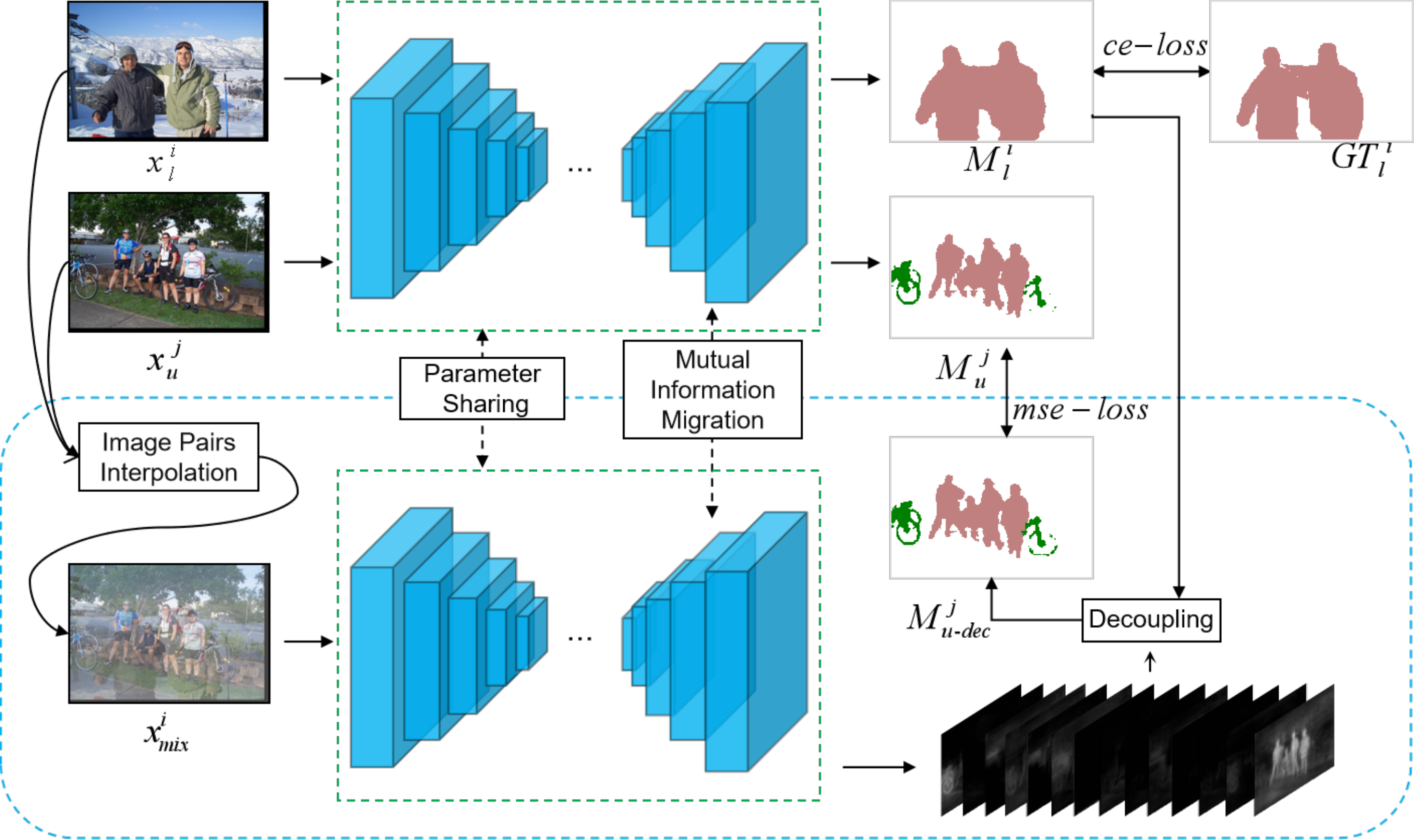} 
    \caption{Overview of our proposed semi-supervised segmentation approach. GuidedMix-Net follows the basic architecture of U-Net, consisting of an encoder-decoder architecture.
    The main decoder is constructed by ResNet, while the decoder is incorporated by our MITrans modules.}
    \label{IMMD}
\end{figure*}

Assume that we have a limited number of labeled images ${S}_{l}$=\{${x}_{l}, {y}_{l}$\}, where ${y}_{l}$ is the ground-truth mask of the image ${x}_{l}$, and a large amount of data without annotations ${S}_{u}$=\{${x}_{u}$\}. 
The image ${x}\in{\mathbb{R}}^{H\times W}$ has spatial dimensions of ${H}\times{W}$ and the masks ${y}\in{\mathbb{R}}^{H\times W\times C}$ have ${C}$ categories.
Fully supervised methods aim to train a CNN $\Gamma(x; \theta)$ that takes image ${x}$ as input, where $\theta$ denotes the parameter of the model, and outputs the segmented mask $\hat{y}$ by minimizing the cross-entropy loss $\mathbb{L}_{ce}$ as follows:
\begin{equation}
\mathbb{L}_{ce}(\hat{y}, y)=-\sum_{i}\hat{y}_{i}log({y}_{i}),
\label{ce_mine}
\end{equation}
where ${i}$ represents the i-th category.
Generally, collecting large-scale labeled training data is time-consuming, costly and sometimes infeasible. 
In contrast, for some computer vision tasks, large amounts of unlabeled data can be collected relatively easily and labeled samples are hard to be obtained.
In this case, fully supervised training scheme cannot achieve good performance when suffering in the presence of a minor data deficiency.
To address this and employ unlabeled examples during training, we propose a novel framework, called GuidedMix-Net, to leverage the limited number of labeled samples to guide the learning of unlabeled data. The overall framework is shown in Fig. \ref{IMMD}.

To leverage the labeled samples to guide the generation of credible pseudo masks for the unlabeled samples,
GuidedMix-Net employs three operations: 1) interpolation of image pairs; 2) transfer of mutual information; 3) generation of pseudo masks, which will be introduced accordingly.

\subsection{Labeled-Unlabeled Image Pair Interpolation}\label{sec:LUPI}
Labeled-unlabeled image pair interpolation (LUPI) applies linear interpolation to formulate a data mixing objective for current unlabeled instances with potentially similar labeled samples to guarantee the cues to be similar, and enable information to flow between them. 
Given a pair of samples $({x}_{l}^{i}, {y}_{l}^{i})$ and ${x}_{u}^{k}$, image-level's interpolation as shown in Eq. \ref{mixup}. 
\begin{equation}
{x}_{mix}({x}_{l}, {x}_{u})=\lambda {x}_{l}+(1-\lambda){x}_{u}.
\label{mixup}
\end{equation}

The output after images interpolation can be expressed as ${x}_{mix}$.
To learn unlabeled data ${x}_{u}$ over the labeled samples ${x}_{l}$, we set $\lambda \leftarrow	 \min(\lambda, 1-\lambda)$, where $\lambda\in(0,1)$ is a hyper-parameter sampled from the $Beta(\alpha, \alpha)$ distribution with $\alpha$.
And here the $\alpha$ is predefined as 1. 

\subsubsection{Similar Image Pair Selection}\label{sec:SPS}
Although LUPI enables models to associate similar cues between the labeled and unlabeled images, randomly selecting an image pair will not allow knowledge to be transferred from labeled to unlabeled data, since the individual pair may not contain enough similar objects.
We overcome this problem by construct similar image pairs in a mini-batch for training. 
Specifically, we add a fully connected layer as a classifier after the encoder to enhance the semantics of the pooled features, and select image pairs with similar features according to the Euclidean distance (shown in Eq. \ref{Euclidean_distance}, where ${\Gamma}_{enc}$ is the encoder module of GuideMix-Net), to conduct image pairs.
Note that the classifier is first trained using labeled data, so it has some recognition ability.
This procedure allows the proposed model to capture the most similar labeled example for each unlabeled image.
Further discussion can be see in Appendices A.
\begin{equation}
{d({x}_{l}^{i}, {x}_{u}^{k})}=\sqrt{\sum_{}^{H}\sum_{}^{W}({\Gamma}_{enc}({x}_{l}^{i})-{\Gamma}_{enc}({x}_{u}^{k}))^2}.
\label{Euclidean_distance}
\end{equation}

\subsection{Mutual Information Transfer}
After mixing the pair of samples, we associate similar cues to enhance the features and generate pseudo masks of the unlabeled samples.
Generally, labeled data corresponds to the credible features, and unlabeled data are treated as poor-quality features, since no supervision signals are present to guide the gradient updates.
This means that in a uniform mixed vector space, for a pair of similar objects from different sources, the poor features can use the credible ones as reference to improve their quality, whether the credible information is located in the short- or long-range.
Although LUPI enables the short-range credible information to flow to the worse features (see the Appendices A), CNNs are ineffective in capturing information from distant spatial locations.
We solve this problem by applying a non-local (NL) block \cite{ouali2020semi} to obtain long-range patches that are similar to a given local region of mixed data.
After mixing labeled-unlabeled images, we obtain a mutual image which contains all the information from the input pairs.
The mixed data ${x}_{mix}$ is then fed to the encoder for segmentation $\Gamma$ until reaching layer ${j}$, providing an intermediate $\mathbf{v}_{j}$ as follows:
\begin{equation}
    \mathbf{v}_{j} = {h}_{j}({x}_{mix}).
    \label{encoder}
\end{equation}

\begin{figure}[t!]
    \centering
    \includegraphics[width=6.5cm, height=3.5cm]{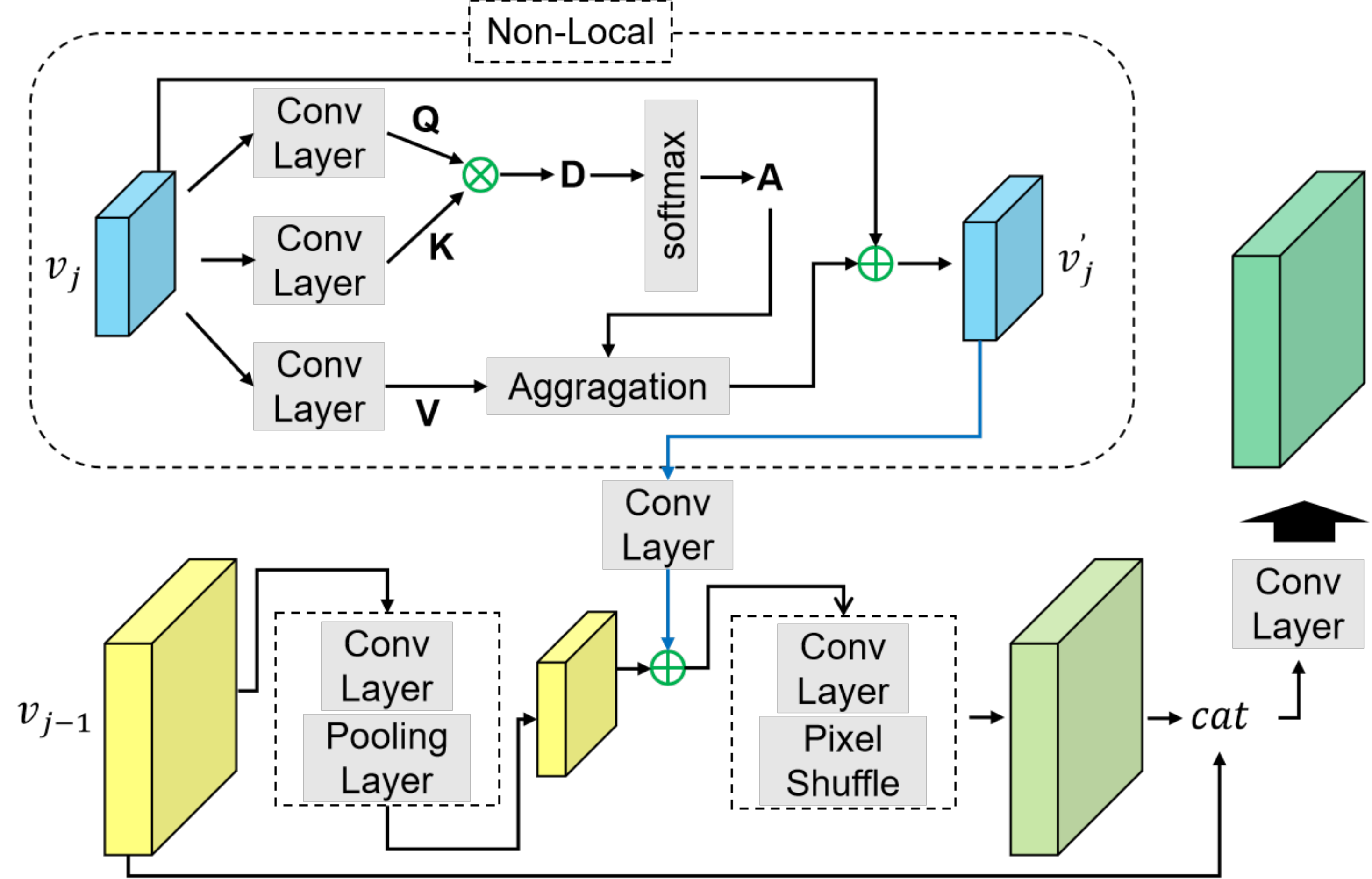}
    \caption{Construction details of the MITrans module.}
    \label{mitrans}
\end{figure}
Another intermediate, $\mathbf{v}_{j-1}$, is produced by the ${j-1}$ layer, where the spatial size of which is double that of $\mathbf{v}_{j}$.
The non-local module collects contextual information from long-range features to enhance local feature representation.
As shown in the dotted box of Fig. \ref{mitrans}, the module uses two convolutional layers of $1\times1$ and filters $\mathbf{v}_{j}$ to map and obtain two features $\textbf{Q}$ and $\textbf{K}$.
The spatial size is the same as for $\mathbf{v}_{j}$, but the channel number is half that of $\mathbf{v}_{j}$ double for dimension reduction.
Then, we generate a correlation matrix $\textbf{D}$ by calculating the correlation between $\textbf{Q}$ and $\textbf{K}$. 
A softmax layer is also utilized on $\textbf{D}$ over the channel dimension to get the attention map $\textbf{A}=f(\textbf{Q}, \textbf{K})$.
To obtain the adapted features of $\mathbf{v}_{j}$, another convolutional layer with a $1\times1$ filter is used to generate $\textbf{V}$ without size change.
The long-range contextual information is captured by an aggregation operation, as follows:
\begin{equation}
\mathbf{v}_{j,n}^{'} = 
\frac{1}{C(x)}\sum\limits_{\mathbf{\forall m}} f(\textbf{Q}_{n}, \textbf{K}_{m}) \mathbf{V}_{\textbf{m}} + \mathbf{v}_{j,n},
\label{Aggregation}
\end{equation}
where ${n}$ and ${m}$ are the indices of the position in the variable space whose response needs to be computed and all other potential positions, respectively. 
Parameters $\mathbf{v}_{j,n}^{'}$ denotes the feature vector in the final output features $\mathbf{v}_{j}^{'}$ at position $\mathbf{n}$, and $\mathbf{v}_{j,n}$ is a feature vector in $\mathbf{v}_{j}$ at position $\mathbf{n}$.
The function ${f}$ is used to represent relationships such as the affinity between $\textbf{Q}_{n}$ and all $\textbf{K}_{m}$.
Finally, ${C(x)}$ is a normalization factor.

After that, several convolutional layers are combined with the PixelShuffle layer \cite{shi2016real} to fuse the low-level features and restore the spatial information (shown in Fig. \ref{mitrans}).
The model is first trained on the labeled samples before mixing. 
Although the number of training samples is small, they provide some recognition ability. 
The non-local blocks use the features of the labeled samples as the novel training signals to correct the feature generation of the unlabeled samples in the mixed data (\textit{i.e.} inter-images).
Unlike previous methods, which primarily focus on intra-image information,  the ``mutual information transfer" module addresses the semantic relations within for images comprehensive object pattern mining.
The proposed module captures semantic similarities for image themselves to build a mutual information transformation mode, and thus improves the prediction of the unlabeled samples.

Note that the non-local blocks in MITrans are used to associate similar patches from the labeled and unlabeled features.
Therefore, they can leverage supervision signals for unlabeled data training. 

\subsection{Pseudo Mask Generation}
To effectively learn from the unlabeled samples, we need to decouple high-quality pseudo masks from the mixed data $x_{mix}$.
According to the translation equivariance \cite{Goodfellow2015DeepL} of the convolution operator, the translation operated on the input image is still detectable on the output features with the corresponding translation.
The translation equivariance can be reflected in the mixed data as well, as shown in Appendices B, the visualization (a), (b), (c), where the activations for the spatial locations of the interesting object, ${e.g.}$, bus and person, are invariant. 
Further, the predicted layer of the segmentation network assigns these activated features to certain category channels, separately.
We focus on this and propose a pseudo mask generation (PMG) to generate the masks for the unlabeled instances by conducting subtraction between the predictions to constrain foreground separation, which comes from labeled and unlabeled data, respectively.
We then train both the labeled and unlabeled data jointly.

In general, semantic segmentation can be regarded as seeking a mapping function $\Gamma$, such that the output $M=\Gamma(x)$ is the desired mask which is close to the ground-truth.
For a pair of labeled and unlabeled images $({x}_{l},{x}_{u})$, the predictions are ${M}_{l} = \Gamma({x}_{l})$ and ${M}_{u} = \Gamma({x}_{u})$, respectively.
After feeding ${x}_{mix}$ from Eq. \ref{mixup} into the segmentation network $\Gamma$, we can obtain the predicted mask ${M}_{mix} = \Gamma({x}_{mix})$, which can be treated as the approximation of directly mixing the masks ${M}_{l}$ and ${M}_{u}$:
\begin{equation}
{M}_{mix}=\Gamma({x}_{mix})\approx{M}_{l}+{M}_{u}.
\label{mixmask}
\end{equation}
We decouple the pseudo masks of ${x}_{u}$ from ${M}_{mix}$, and then leverage them as the target to calculate the mean squared error loss (with the direct output of ${x}_{u}$ from the main decoder).
This procedure ensures the model to be robust and less sensitive to small perturbations.

\subsubsection{Hard Decoupling}
The goal of mask decoupling is to eliminate ${M}_{l}$ from the mixed data and then generate pseudo masks for the unlabeled samples.
Considering that the ground-truth of labeled data are provided for the model in the early training stage, the prediction ${M}_{l}$ has higher probability of being close to the real mask.
Once ${M}_{l}$ is obtained, we can directly decouple the unlabeled data mask ${M}_{u-dec}$ using Eq. \ref{decouple}, which we refer to as hard decoupling:
\begin{equation}
{M}_{u-dec}={M}_{mix}-{M}_{l}.
\label{decouple}
\end{equation}

Hard decoupling is reasonable since the neural network has the ability to separate the corresponding category channels (an example is shown in Appendices B).
Directly subtracting between final predictions can separate and obtain more refined results for the unlabeled samples.

\subsubsection{Soft Decoupling}
The proposed hard decoupling directly performs a subtraction between the prediction of the mixed image and the labeled image, which may counteract the prediction of the overlapping region of objects in the mixed image.
To overcome this problem, we propose a soft decoupling for the pseudo mask generation as follows:
\begin{equation}
{M}_{u-dec}={M}_{mix}-\lambda{M}_{l},
\label{soft-dec}
\end{equation}
where $\lambda$ is the parameter from the $Beta(\alpha, \alpha)$ distribution.
Soft decoupling retains the details of the overlapping region by weakening the intensity of ${M}_{l}$ in ${M}_{mix}$.
As shown in Table \ref{ab_each-com}, soft decoupling is better than hard decoupling.

GuidedMix-Net pays attention to the outline of objects in a complex environment by first transferring knowledge from the mixed labeled-unlabeled pairs, and then decoupling their predictions, to understand the complete semantic information of objects. 
As shown in Appendices D, GuidedMix-Net is complements object contours and semantic understanding.

\subsection{Loss Function}
We develop an overall loss function $\mathbb{L}$ for our consistency based semi-supervised learning (SSL) as follows:
\begin{equation}
\mathbb{L}=\mathbb{L}_{sup}+{\omega}_{usup}\mathbb{L}_{usup},
\label{total_loss}
\end{equation}
where ${{\omega}_{usup}}$ is an unsupervised loss weight, such as \cite{laine2016temporal}, that controls the balance between the two losses. On the one hand, $\mathbb{L}_{usup}$ in Eq. \ref{mse_ulmix} is an unsupervised mean squared error (MSE) loss to calculate the difference between the decoupling mask ${M}_{u-dec}$ and the direct prediction ${M}_{u}$:
\begin{equation}
\mathbb{L}_{usup}=\frac{1}{H*W}\sum_{}^{H}\sum_{}^{W}({M}_{u-dec}-{M}_{u})^2.
\label{mse_ulmix}
\end{equation}
On the other hand, for supervised training, the loss $\mathbb{L}_{sup}$ consists of three terms to optimize the model as follows:
\begin{equation}
\mathbb{L}_{sup}={L}_{ce}({M}_{l}, {y}_{l})+{L}_{dec}+{L}_{cla},
\label{sup_loss}
\end{equation}
where ${L}_{ce}({M}_{l}, {y}_{l})$ is the same as in Eq. \ref{ce_mine}, and ${L}_{cla}$ is the classifier loss term for image-level annotations. 
For ${L}_{dec}$, we first select a sample $\hat{{x}_{l}}$ for a seed ${x}_{l}$ according to the matching rules in Sec. \ref{sec:SPS}, where $\lambda$ follows Sec. \ref{sec:LUPI}. 
We then denote $\hat{{y}_{l}}$ and $\hat{{M}_{l}}$ (${{y}_{l}}$ and ${{M}_{l}}$) as the corresponding ground-truth and prediction of $\hat{{x}_{l}}$ (the seed ${x}_{l}$), respectively.
In addition, a mixup operation can be conducted on both labeled sample ${{x}_{l}}$ and $\hat{{x}_{l}}$ following Eq. \ref{mixup} to obtain a mixed sample ${x}_{mix}^{l}$=$\lambda{x}_{l}+(1-\lambda)\hat{{x}_{l}}$.
A consistency loss between decoupled masks ${M}_{dec}={M}_{mix}-{M}_{l} $ and the prediction $\hat{{M}_{l}}$ can be defined as
\begin{equation}
\mathbb{L}_{dec}=\frac{1}{H*W}\sum_{}^{H}\sum_{}^{W}({M}_{dec}-{M}_{l})^2.
\label{sup_con_loss}
\end{equation}
Another analysis of GuidedMix-Net shown in Appendices C.

\section{Experiments}

\subsection{Dataset and Evaluation Metrics}
\textbf{PASCAL VOC 2012.} This dataset is widely used for semantic segmentation and object detection.
It consists of 21 classes including background.
We use 1,464 training images and 1,449 validation images from the original PASCAL dataset, and also leverage the augmented annotation dataset (involving 9,118 images) \cite{Hariharan2011Semantic} like \cite{2018Weakly,zhao2017pyramid}.
\\\textbf{Cityscapes.} We use Cityscapes to further evaluate our model. 
This dataset provides different driving scenes distributed in 19 classes, with 2,975, 500, 1,525 densely annotated images for training, validation and testing.
For semantic segmentation, 59 semantic classes and 1 background class are used in training and validation, respectively.
\\\textbf{Evaluation Metric.} Common data augmentation methods are used during our training procedure, which including random resizing (scale: 0.5$\sim$2.0), cropping (321$\times$321 for PASCAL VOC 2012, 513$\times$513 for Cityscapes, 480$\times$480 for PASCAL Context), horizontal flipping and slight rotation. 
We evaluate different methods by measuring the averaged pixel intersection-over-union (IoU).

\subsection{Network Architecture and Training Details}
\textbf{Encoder.} The encoder is based on ResNet \cite{he2016deep} pretrained on ImageNet \cite{krizhevsky2012imagenet}, and also includes the PSP module \cite{zhao2017pyramid} after the last layer.
\\\textbf{Decoder.} GuidedMix-Net combines labeled and unlabeled data by image linear interpolation, setting the new SOTA, for semi-supervised semantic segmentation. 
To avoid the disintegrates of details for the mixed pairs, we employ a skip connection in the decoder, as done in U-Net \cite{Ronneberger2015UNetCN}.
A pixel shuffle layer \cite{shi2016real} is also utilized to restore the spatial resolution of features.
\\\textbf{Training Details.}
Similar to \cite{chen2017deeplab}, we use a ``poly'' learning rate policy, where the base learning rate is multiplied by $((1-\frac{iter}{max_ier})^{}power)$ and ${power=0.9}$. 
Our segmentation network is optimized using the stochastic gradient descent (SGD) optimizer with a base learning rate of 1e-3, momentum of 0.9 and a weight decay of 1e-4.
The model is trained over 40,000 iterations for all datasets, and the batch-size is set to 12 for PASCAL VOC 2012, and 8 for Cityscapes and PASCAL Context. 
We conduct all our experiments on a Tesla V-100s GPU.

\subsection{Results on Pascal VOC 2012}
\subsubsection{Ablation Studies}
Our ablation studies examine the effect of different values of $\lambda$ and the impact of different components in our framework.

\textbf{Different $\lambda$ Values.}
The results under different values of $\lambda$ are reported in Table \ref{impact_lambda}. 
An can be see, changing $\lambda$ used in Sec. \ref{sec:LUPI} impacts the results, because $\lambda$ controls the intensity of pixels in the mixed input data.
As shown in Table \ref{impact_lambda}, too high or too low a $\lambda$ is not conducive to model optimization.
A high $\lambda$ value leads to the labeled information being discarded, while a low $\lambda$ value results in the unlabeled data being covered.
When $\lambda < 0.5$, GuidedMix-Net provides the best performance on PASCAL VOC 2012.
We thus select $\lambda < 0.5$ for the remaining experiments on PASCAL VOC 2012.
\begin{table}[h]
\small
\centering
\caption{The impact of different lambda values on the experimental results of PASCAL VOC 2012.}
\begin{tabular}{c|c|c|c|c}
\hline 
\multirow{2}{*}{$\lambda$} & \multirow{2}{*}{backbone} & \multicolumn{2}{c|}{Data} & \multirow{2}{*}{mIoU} \\\cline{3-4}
                        &      &  labels   & unlabels        &                       \\\hline
$<$ 0.1 & \multirow{5}{*}{ResNet50} &  \multirow{5}{*}{1464} & \multirow{5}{*}{9118}   & 67.9 \\
$<$ 0.2 &  &  &      & 69.5 \\
$<$ 0.3 &  &  &      & 70.7 \\
$<$ 0.4 &  &  &      & 71.4 \\
$<$ 0.5 &  &  &      & \textbf{73.7} \\\hline
\end{tabular}
\label{impact_lambda}
\end{table}
\vspace{-1mm}
\begin{table}[h]
\small
\centering
\caption{Ablation studies of using similar image pairs, MITrans, and hard $\&$ soft decoupling modules in GuidedMix-Net. 
We train the models on ResNet50 and test them on the validation set of PASCAL VOC 2012.}
\begin{tabular}{c|c|c|c|c}
\hline 
\multirow{2}{*}{Method} & \multirow{2}{*}{used} & \multicolumn{2}{c|}{Data} & \multirow{2}{*}{mIoU} \\\cline{3-4}
						&      &  labels   & unlabels        &                       \\\hline
\multirow{2}{*}{Similar Pair} & $\times$& \multirow{7}{*}{1464} & \multirow{7}{*}{9118} & 71.9 \\
								& $\surd$ &      &      & \textbf{73.7} \\\cline{1-2}\cline{5-5}
\multirow{2}{*}{MITrans}      & $\times$&      &      & 72.7 \\
								& $\surd$ &      &      & \textbf{73.7} \\\cline{1-2}\cline{5-5}
Hard Decoupling               & $\surd$ &      &      & 71.4 \\
Soft Decoupling               & $\surd$ &      &      & \textbf{73.7} \\\hline\hline
Suponly $w/o$ MITrans          & $\surd$ & \multirow{2}{*}{1464} & \multirow{2}{*}{9118} & 70.2 \\
Suponly $w/$ MITrans         & $\surd$ &      &      & 70.5 \\\hline
\end{tabular}
\label{ab_each-com}
\end{table}
\vspace{-1mm}

\textbf{Different Components.}
As shown in Table \ref{ab_each-com}, we evaluate the influence of different components of GuidedMix-Net. 
For fair comparison, we evaluate one component per experiment and freeze the others.
Firstly, we investigate different strategies for constructing image pairs, ${i.e.}$, random selection of similar pairs.
i) The first strategy randomly selects a labeled image for each unlabeled image in a mini-batch.
ii) To seek similar pairs of labeled and unlabeled images, we add a classifier after the encoder model. 
We match the most similar unlabeled images for each labeled sample according to the Euclidean distance between the features.
As shown in the Table \ref{ab_each-com}, the similar pairs bring a 2.5$\%$ mIoU gain over the plain random selection (71.9$\%$ vs. 73.7$\%$). 
The construction of similar pairs provides context for the target objects in the subsequent segmentation task, and assists GuidedMix-Net in transferring knowledge from the labeled images to the unlabeled samples, with little increasing complexity.
Secondly, we explore whether MITrans is useful for knowledge transfer.
The results provided in the Table \ref{ab_each-com} clearly show that MITrans achieves a significant mIoU gain of 1.4$\%$ (72.7$\%$ vs. 73.7$\%$) by explicitly referencing similar and high-confidence non-local feature patches to refine the coarse features of the unlabeled samples.
Thirdly, as shown in the Table \ref{ab_each-com}, the performance of soft decoupling is 3.2$\%$ better than hard decoupling (71.4$\%$ vs. 73.7$\%$), since soft decoupling considers the overlapping that occurs in the mixed data and tends to preserve local details.
The various components used in our GuidedMix-Net are beneficial alone, and therefore combining them leads to significantly improved optimization.

The above experimental results indicated that all of our designed components were conducive to semi-supervisedd semantic segmentation learning.
However, we still puzzle with was the performance gain of the MITrans module attributed to the knowledge transfer of using labeled samples as a reference paradigm when learning unlabeled data representations?
Or to the feature expression of labeled samples enhanced by nonlocal blocks?
We further examined this question by additional experiments, as shown in Table \ref{ab_each-com}.
``Suponly $w/o$ MITrans'' was the experiment training on the labeled samples only and the nonlocal blocks in MITrans were removed to obtain 70.2 mIoU.
IN contrast, ``Suponly $w/$ MITrans'' was the experiment training on the labeled samples only, and MITrans was fully used to obtain 70.5 mIoU.
This set of experiments clearly showed that there was little performance gain for nonlocal labeled samples.
The huge performance gains described above are mainly due to MITrans' ability to effectively transfer knowledge from labeled samples to unlabeled data.

\subsubsection{Semi-Supervised Semantic Segmentation} 
The experimental setting of CCT \cite{ouali2020semi} is different from other semi-supervised semantic segmentation methods AdvSSL \cite{2018Adversarial}, S4L \cite{zhai2019s4l}, GCT \cite{ke2020guided}, CutMix \cite{french2020semi}, Reco \cite{liu2021bootstrapping}, ClassMix \cite{olsson2021classmix} and SSContrast \cite{alonso2021semi}.
Here, we provide the comparison experiment in Table \ref{comparison with ssl}.
We can see that GuidedMix-Net outperforms CCT with a performance increase of over 5.7$\%$.

\textbf{Comparing with Other State of the Arts.}
We explore the performance using the deeper backbone of ResNet101 for the semi-supervised semantic segmentation task.
The results show in Table \ref{comparison with ssl}.
GuidedMix-Net outperforms the current methods for semi-supervised image segmentation by $3.4\%$, $3.7\%$, and $3.4\%$ on 1/8 labels, 1/4 labels, 1/2 labels, respectively.
The significant performance gains on different ratios of labeled data demonstrate that GuidedMix-Net is a generally efficient and effective semi-supervised semantic segmentation method.  
Quality visualization results provided in Appendices D.

\begin{table}[h]
\small
\centering
\caption{Comparison with other state-of-the-art semi-supervised semantic segmentation methods under different ratios of labeled data on PASCAL VOC 2012.}
\begin{tabular}{c|c|c|c|c}
\hline
    SSL & 1/8 & 1/4 & 1/2 &-\\\hline
    AdvSSL & 68.4 & 70.8 & 73.3 &-\\
    S4L & 67.2 & 68.4 & 72.0 &-\\
    GCT & 70.7 & 72.8 & 74.0 &-\\
    ReCo & 71.0 & - & - &-\\
    CutMix & 70.8 & 71.7 & 73.9 &-\\
    Ours & \textbf{73.4} & \textbf{75.5} & \textbf{76.5}&-\\\hline\hline

    SSL & \begin{tabular}[c]{@{}c@{}}500 \\ labels\end{tabular} & \begin{tabular}[c]{@{}c@{}}1000 \\ labels\end{tabular} & \begin{tabular}[c]{@{}c@{}}1464 \\ labels\end{tabular} & \begin{tabular}[c]{@{}c@{}}Memory \\ Size\end{tabular} \\\hline
    CCT & 58.6 & 64.4 & 69.4 & 24kM \\
    Ours & \textbf{65.4} & \textbf{68.1} & \textbf{73.7} & \textbf{15kM} \\\hline
\end{tabular}
\label{comparison with ssl}
\end{table}

\begin{table}[h]
\small
\centering
\caption{Influence of different lambda values on the experimental results of Cityscapes.}
\begin{tabular}{c|c|c|c|c}
\hline 
\multirow{2}{*}{$\lambda$} & \multirow{2}{*}{backbone} & \multicolumn{2}{c|}{Data} & \multirow{2}{*}{mIoU} \\\cline{3-4}
                        &      &  labels   & unlabels        &                       \\\hline
$<$ 0.1 & \multirow{5}{*}{ResNet101} & \multirow{5}{*}{1/8} & \multirow{5}{*}{7/8} & 64.3\\
$<$ 0.2 &  &  &      & 64.0 \\
$<$ 0.3 &  &  &      & \textbf{65.8} \\
$<$ 0.4 &  &  &      & 65.5 \\
$<$ 0.5 &  &  &      & 65.7 \\\hline
\end{tabular}
\label{impact_lambda_city}
\end{table}
\begin{table}[h]
\small
\centering
\caption{Comparison with other semi-supervised semantic segmentation methods under different ratios of labeled data on Cityscapes.}
\begin{tabular}{c|c|c|c|c}
\hline
    SSL & 100 & 1/8 & 1/4 & 1/2 \\
    Methods & labels & labels & labels & labels \\\hline
    AdvSSL & - & 57.1 & 60.5 & - \\\hline
    s4GAN & - & 59.3 & 61.9 & - \\\hline
    CutMix & 51.2 & 60.3 & 63.9 & - \\\hline
    ClassMix & 54.1 & 61.4 & 63.6 & 66.3 \\\hline
    ReCo & 56.5 & 64.9 & 67.5 & 68.7\\\hline
    Ours & \textbf{56.9} & \textbf{65.8} & \textbf{67.5} & \textbf{69.8}\\\hline
\end{tabular}
\label{the experiments of city}
\end{table}
\subsection{Results on Cityscapes}
Cityscapes has 2,975 training images.
In our experiments, we divide them into 1/8 labels and 1/4 labels, while the remaining data are treated as unlabeled.
We use ResNet101 as the backbone to train the models.
Since the optimal value of $\lambda$ varies with the training dataset, we conduct experiments on Cityscapes leveraging 1/8 labeled images as the training data to explore the impact of $\lambda$ on this dataset, and show the results in Table \ref{impact_lambda_city}.
For Cityscapes, when $\lambda < 0.3$, GuidedMix-Net achieves 65.8 mIoU on the validation dataset, which is better than other selected value ranges.
We thus fix the value of $\lambda$ to be less than 0.3, and verify the gap between GuidedMix-Net and other approaches.
Relevant results are presented in Table \ref{the experiments of city}.
GuidedMix-Net yields considerable improvements on Cityscapes over other semi-supervised semantic segmentation methods, \textit{i.e.}, mIoU increases of $0.7\%$, $1.4\%$, and $1.6\%$ for the 100, 1/8, and 1/2 labels, respectively.
The distribution of different classes on Cityscapes is highly imbalanced.
The vast majority of classes are present in almost every image, and the few remaining classes occur scarcely.
As such, inserting a classifier after the encoder to semantically enhance features and assist in matching similar images is unhelpful.
Thus, we use the mixture of randomly selected image pairs in GuidedMix-Net.
We also provide the visul results on Cityscapes in Appendices E using only 1/8 labeled images.

In addition, the experiments of GuidedMix-Net, which compares with other SOTA on PASCAL-Context dataset had introduced in Appendices F.

\section{Conclusion}
This paper has presented a novel semi-supervised learning method for semantic segmentation, called GuidedMix-Net, and achieves SOTA performance.
In the future, we will investigate the use of unlabeled data in other related areas, such as medical imaging. We will continue improving the learning mechanism of the unlabeled samples guided by labeled data.
\\\textbf{Acknowledgements.} This work is supported by the National Natural Science Foundation of China under Grant No. 61972188 and No. 62122035.

\section{Appendix}
\subsection{Appendices A: About the Similar Image Pair Selection}
\begin{figure}[t!]
    \centering
    \includegraphics[height=7cm]{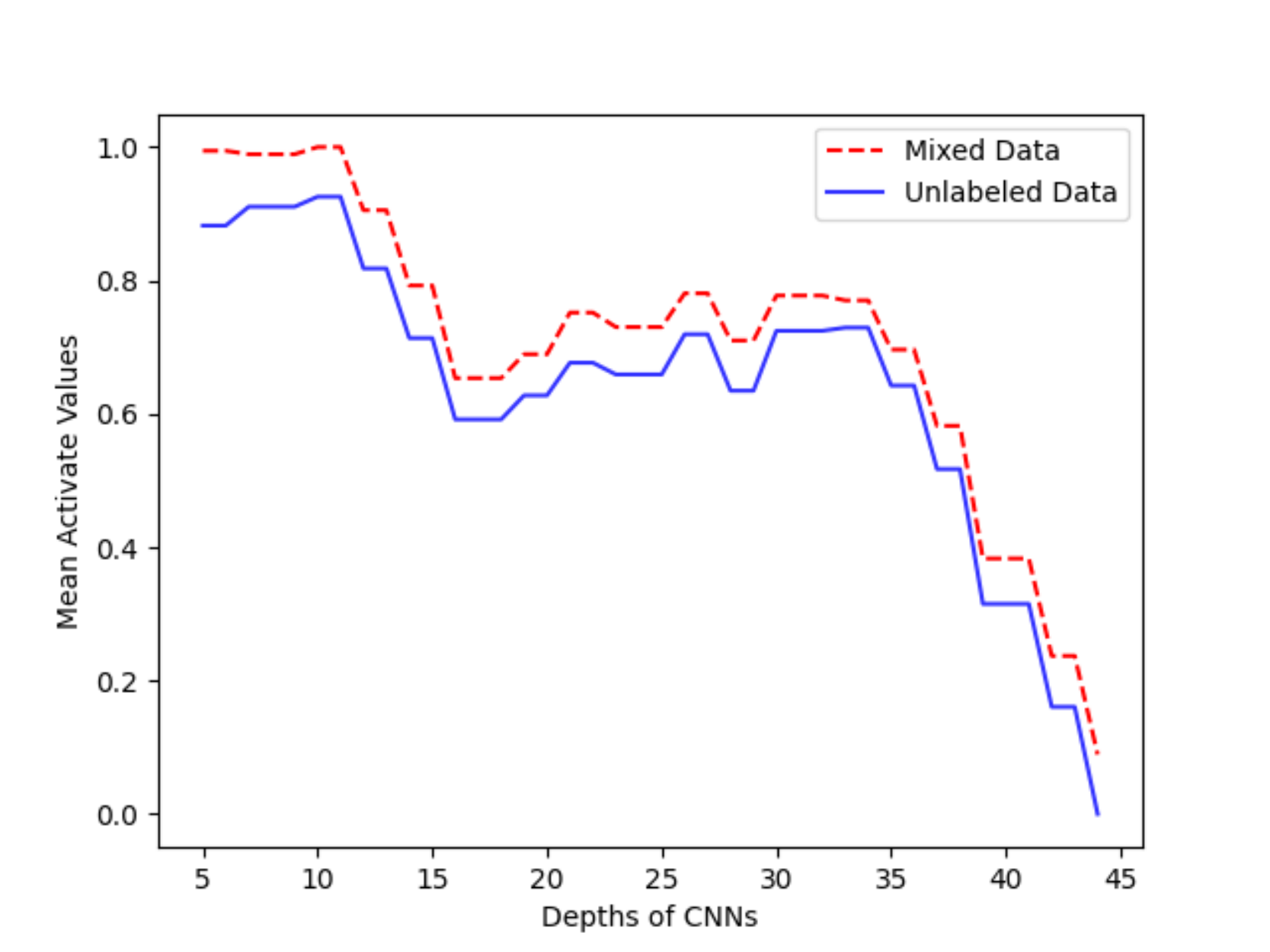}
    \caption{Similar to \protect\cite{Faramarzi2020PatchUpAR}, we provide the comparison of mean activation over all convolutional layers between an unlabeled instance and a mixed image (mixing the labeled-unlabeled image pairs).
    The horizontal axis represents the different levels of convolutional layers in ResNet101 \protect\cite{he2016deep}.
    Filter information in the CNN is represented by the activation value. 
    The higher the value, the greater the effect of activation \protect\cite{Leshno1993MultilayerFN,Bouvrie2006NotesOC}.
    We can see that the mixed data input has higher values of mean activation, which means more filters or neurons in the networks are activated.
    Therefore, with a high probability, more potential structures or correlations can be tapped to improve the segmentation decision.}
    \label{mean_act}
\end{figure}

It is worth noting that Labeled-Unlabeled Image Pair Interpolation (LUPI) is quite different from the typical mixup for image classification, in which both images are associated with one-hot labels. Our method instead attempts to mix a pair of samples by fixing one of them as the labeled one, which brings two advantages:
Firstly, the unlabeled image can be mixed with different labeled samples to obtain more diverse perturbations.
The subsequent optimization of the pseudo mask supervision signal can encourage the model to focus on the object contour.
Secondly, ${x}_{mix}$ is a mixture of ${x}_{l}^{i}$ and ${x}_{u}^{k}$, containing complete contour and texture information of all similar objects in each pair.
To recognize objects, CNNs commonly learn complex representations of object shapes. 
They combine low-level features (${e.g.}$ edges) to increase complex shapes like wheels and car windows until the object can be identified readily \cite{Kriegeskorte2015DeepNN,Goodfellow2015DeepL}.
As neurons interact through local connections, CNNs combine information in a growing perception field, where information is transmitted through successive filters, resulting in purified outputs \cite{chollet2018deep,yosinski2015understanding,Shrikumar2016NotJA}.
GuidedMix-Net unifies the dimensional space of labeled-unlabeled image pairs through a specially developed LUPI.
Further , it leverages the property of CNNs to refer the complete contour and texture information within a short-range enabling it to implicitly refine and diversify the features of unlabeled data, as shown in Fig. \ref{mean_act}.
The proposed method also prepares for the further mutual information transfer.

\subsection{Appendices B: The Processes of Pseudo Mask Generation}
\begin{figure}[t!]
\centering
\subfloat[Labeled data]{\includegraphics[width=3.5cm, height=3cm]{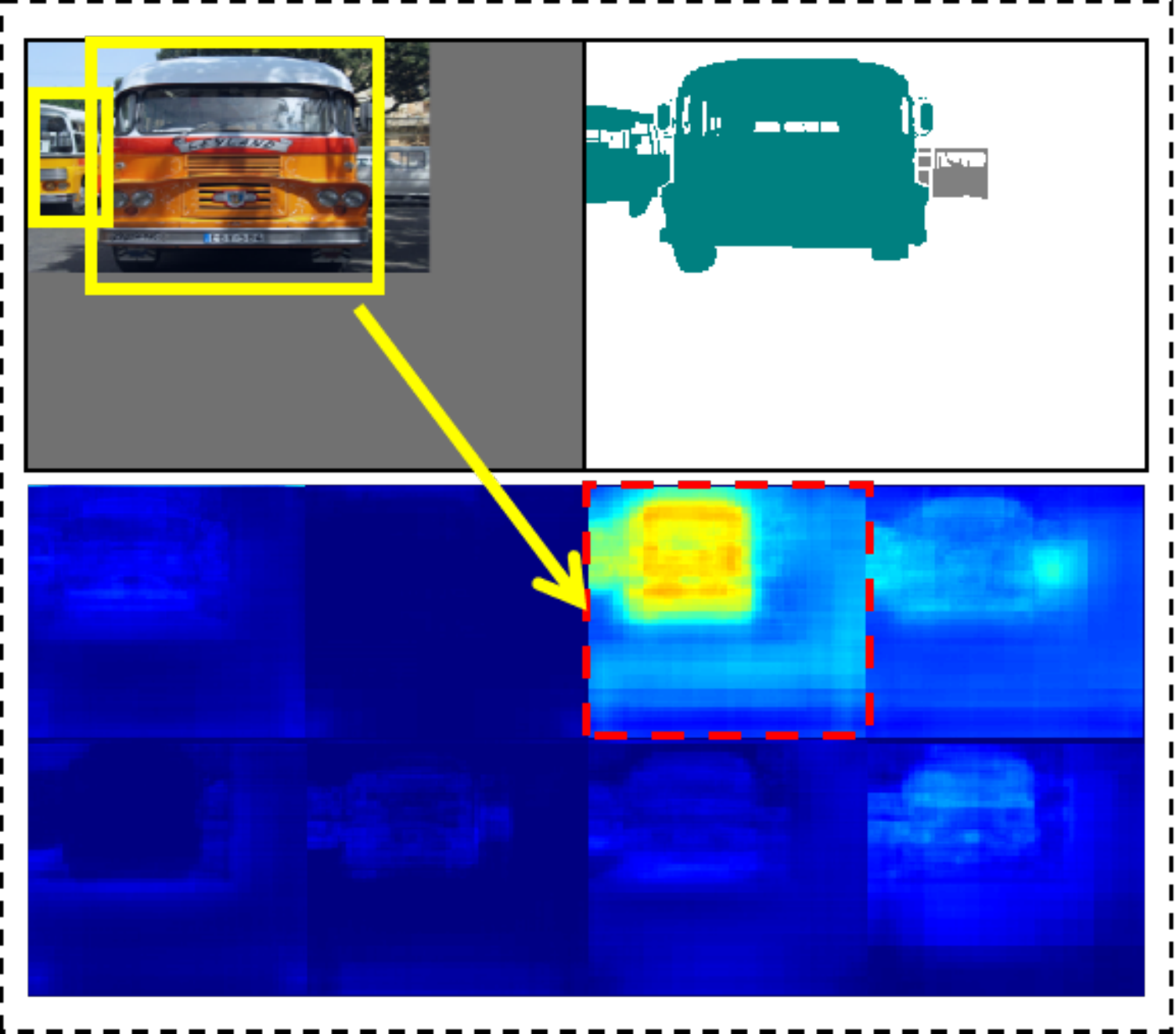}}\hspace{0.1mm}
\subfloat[Unlabeled data]{\includegraphics[width=3.5cm, height=3cm]{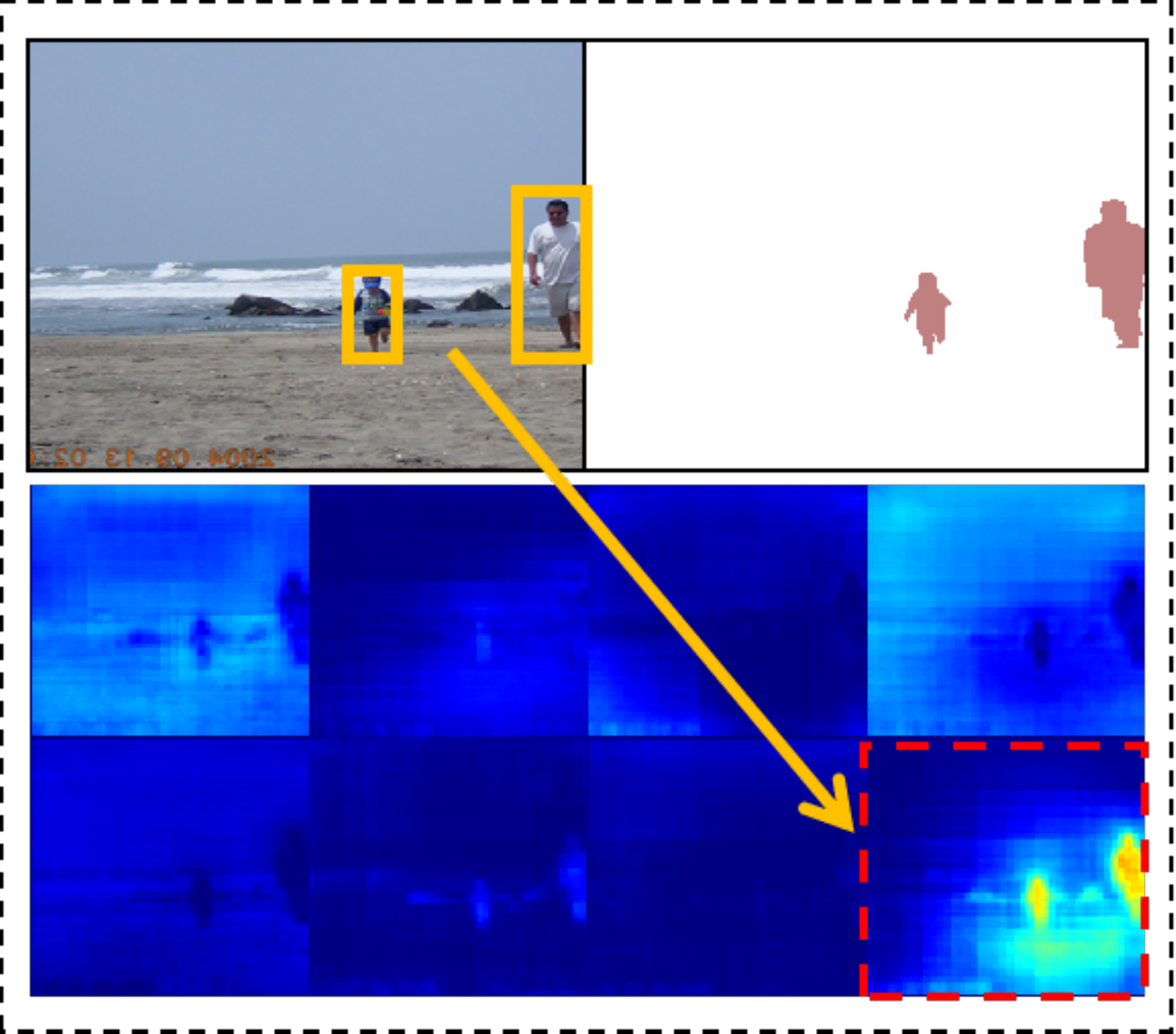}}\hspace{0.1mm}
\subfloat[Mixed data]{\includegraphics[width=3.5cm, height=3cm]{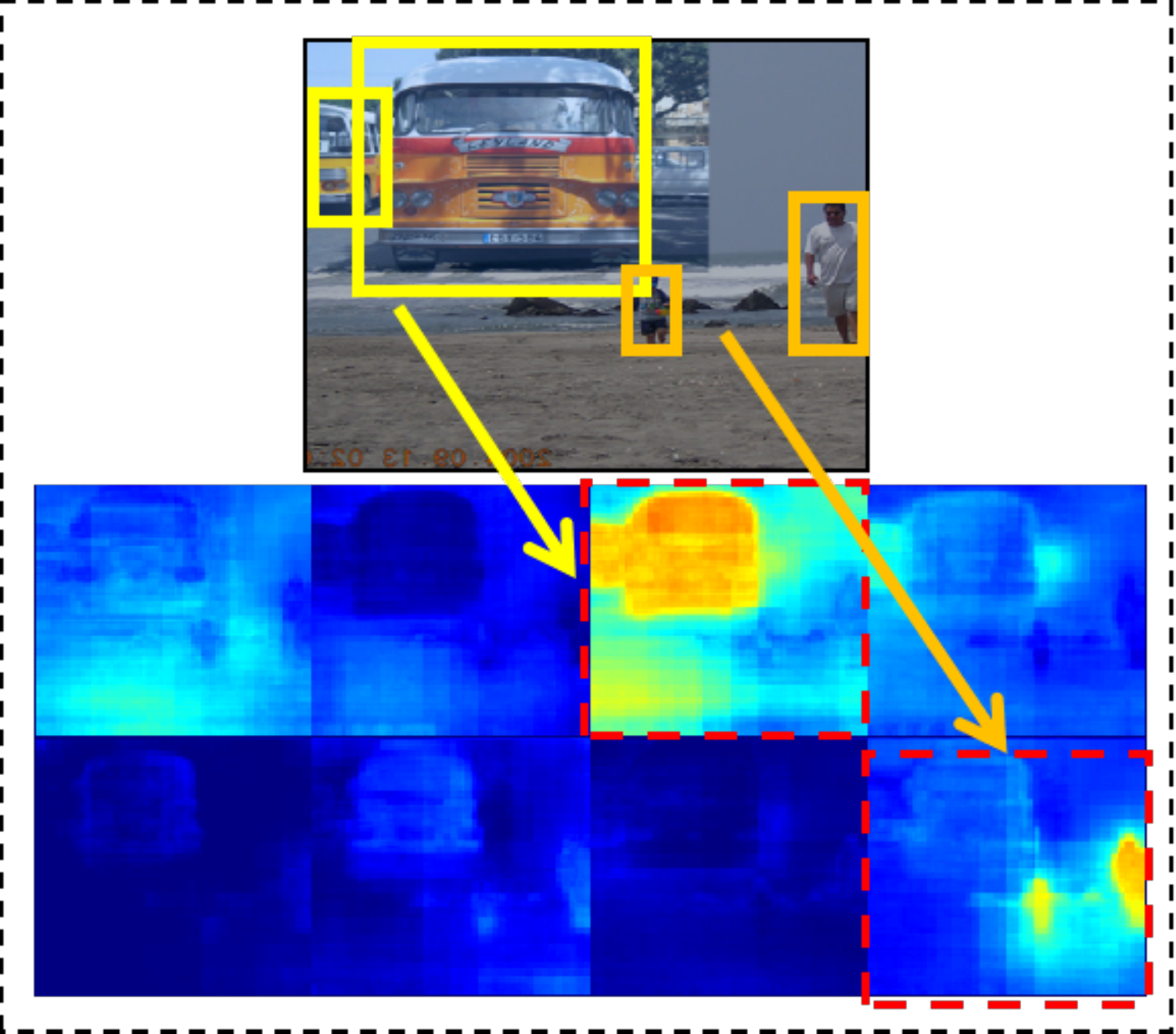}}\hspace{0.1mm}
\subfloat[Decoupled data]{\includegraphics[width=3.5cm, height=3cm]{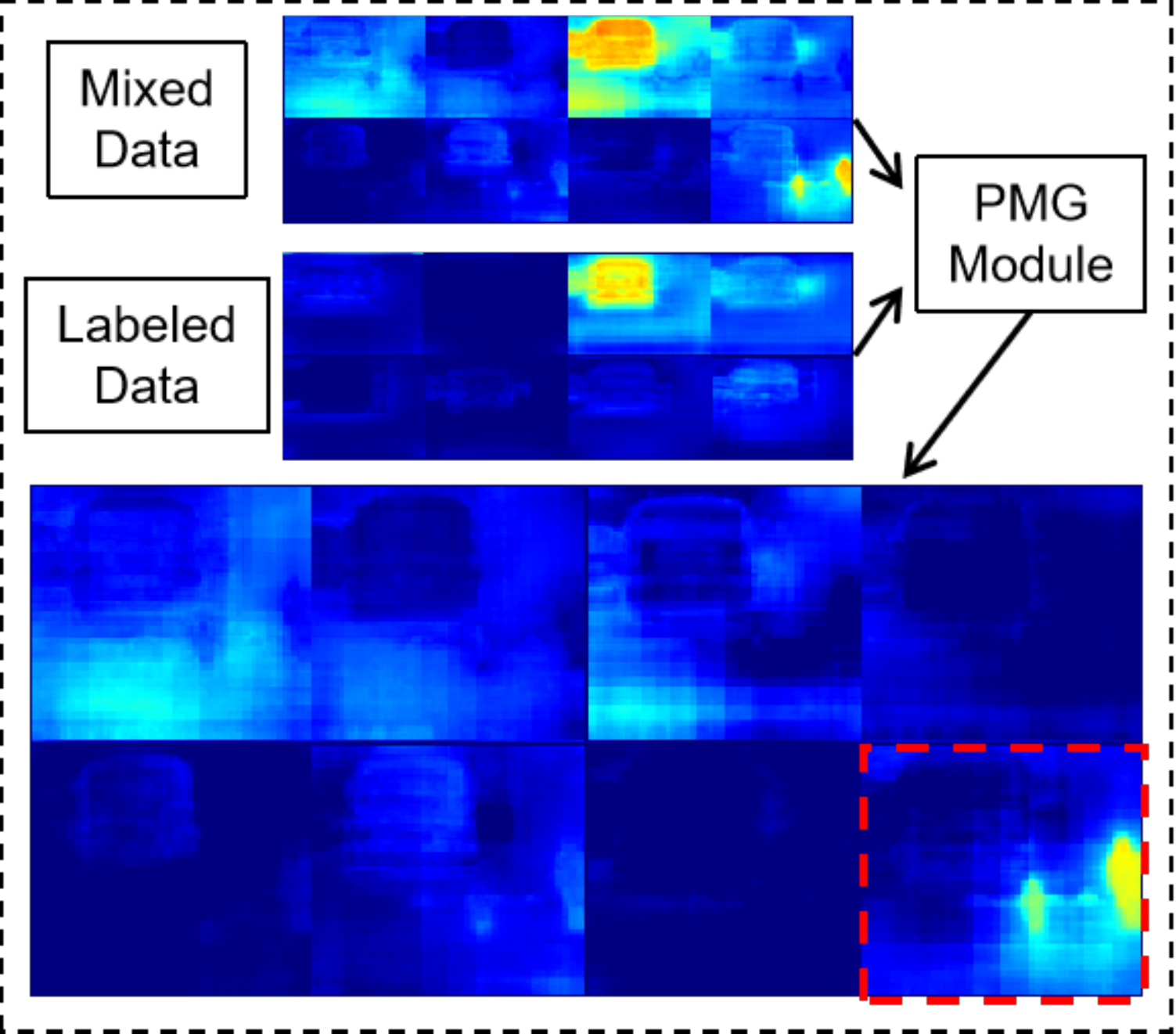}}
\caption{\textbf{Visualization of PMG processes.} Blue images are network features of the input, and the bright area represent the foreground objects. (a) \textbf{Labeled data:} the first row is the image with the corresponding ground-truth, both of which are used for training. In the labeled image, the interested object is a bus, where the channel represents different categories of buses, and the most highlighted area is the location of the bus. (b) \textbf{Unlabeled data:} the first row shows an image and its ground-truth, where only the image itself participates in the training. (c) \textbf{Mixed data:} the first row is the mixed data and its prediction, where the interested objects from (a) and (b) are highlighted. (d) \textbf{Decoupled data:} objects will be activated in their corresponding category channels, which are shown in (a), (b) and (c). Decoupling the pseudo masks by subtraction using PGM is effective for separate and obtaining the required valuable information.}
\label{features_visualization}
\vspace{-1mm}
\end{figure}

Hard decoupling is reasonable since the neural network has the ability to separate the corresponding category channels, an example is shown in Fig. \ref{features_visualization}.
Directly subtracting between final predictions can separate and obtain more refined results for the unlabeled samples.

\subsection{Appendices C: Analysis of GuidedMix-Net}
Edges and textures of images can be interactively combined through layer-by-layer convolutions, and various combinations can be obtained at the higher level of a neural network \cite{2017Feature}.
Inspired by this, we propose LUPI for map labeled-unlabeled image pairs in the same dimension, enabling similar cues to be interacted in the hidden states.
To refine the unlabeled features by learning relevant features from labeled data, we propose MITrans to correct the local errors generated from unlabeled data by capturing the similar long-range cues from the labeled instances.
Fig. 1 and Table 3 of main body show the results of CCT (direct prediction) and our GuidedMix-Net for an unlabeled image.
We can see that the proposed method encourages knowledge transfer from the labeled instances to the unlabeled samples, while other approach cannot achieve. 
We also observe that, although ${x}_{mix}$ seems to overlap between objects, the neural network can easily separate objects in the hidden layer (shown in Fig. \ref{features_visualization}).
In addition, PMG can be used to generate the pseudo masks for the unlabeled samples by performing subtraction after the prediction layer.
In summary, the proposed approach greatly encourages the segmentation model to mine the rich underlying semantic structures provided by the unlabeled samples.

\subsection{Appendices D: Quality Visualization Results of PASCAL VOC 2012}
\begin{figure}[t!]
    \centering
    \subfloat{\includegraphics[width=2cm, height=1.6cm]{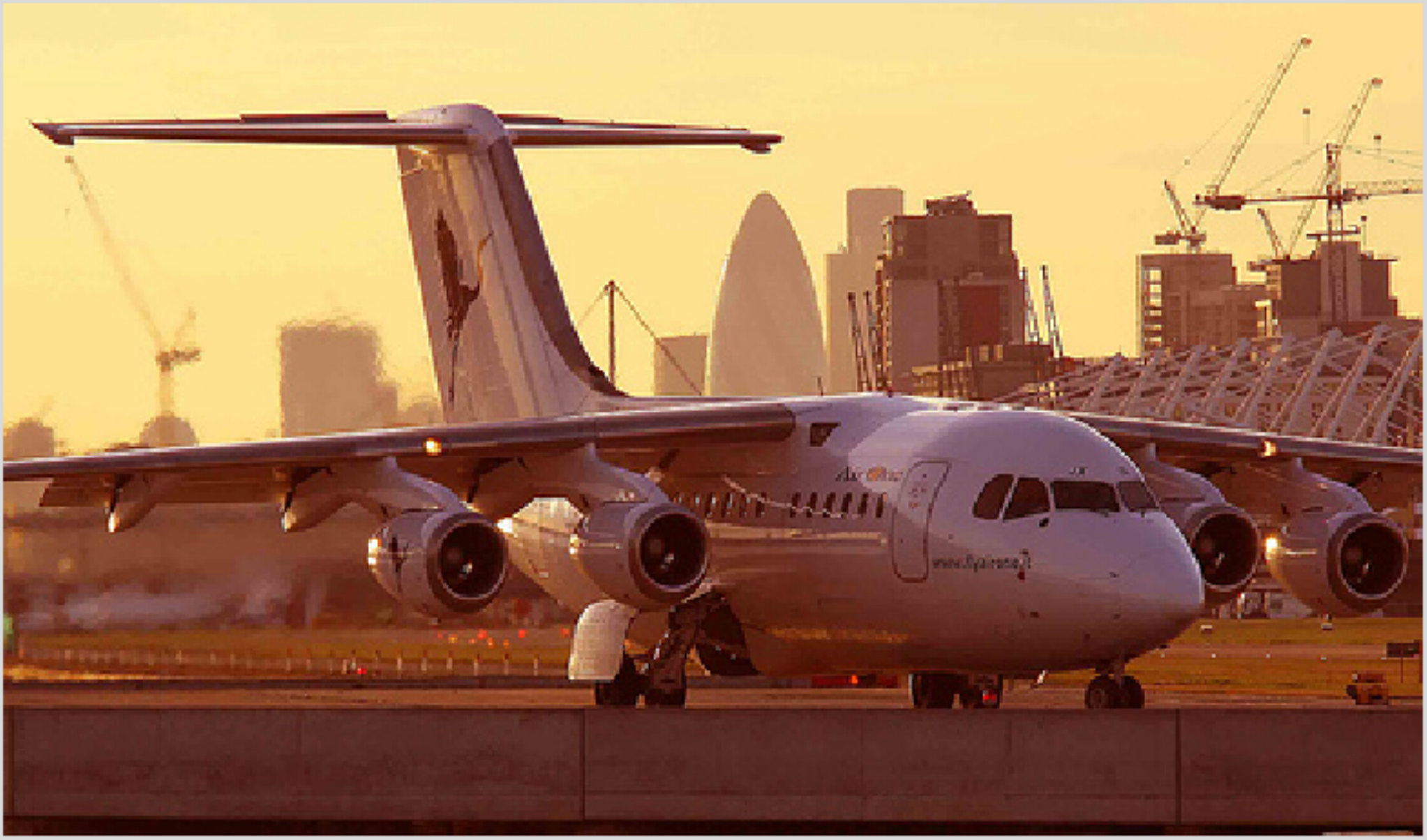}}\hspace{0.1mm}
    \subfloat{\includegraphics[width=2cm, height=1.6cm]{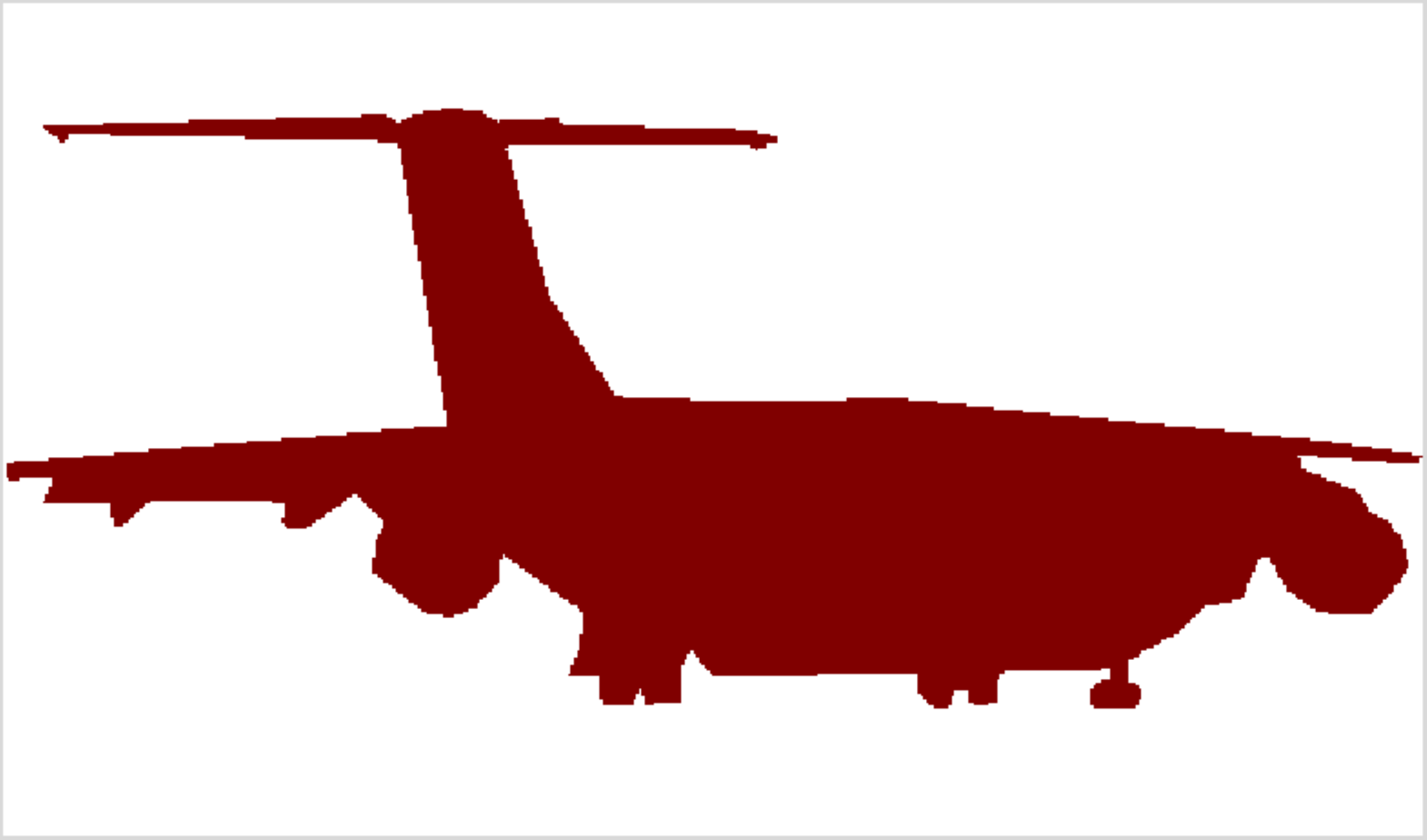}}\hspace{0.1mm}
    \subfloat{\includegraphics[width=2cm, height=1.6cm]{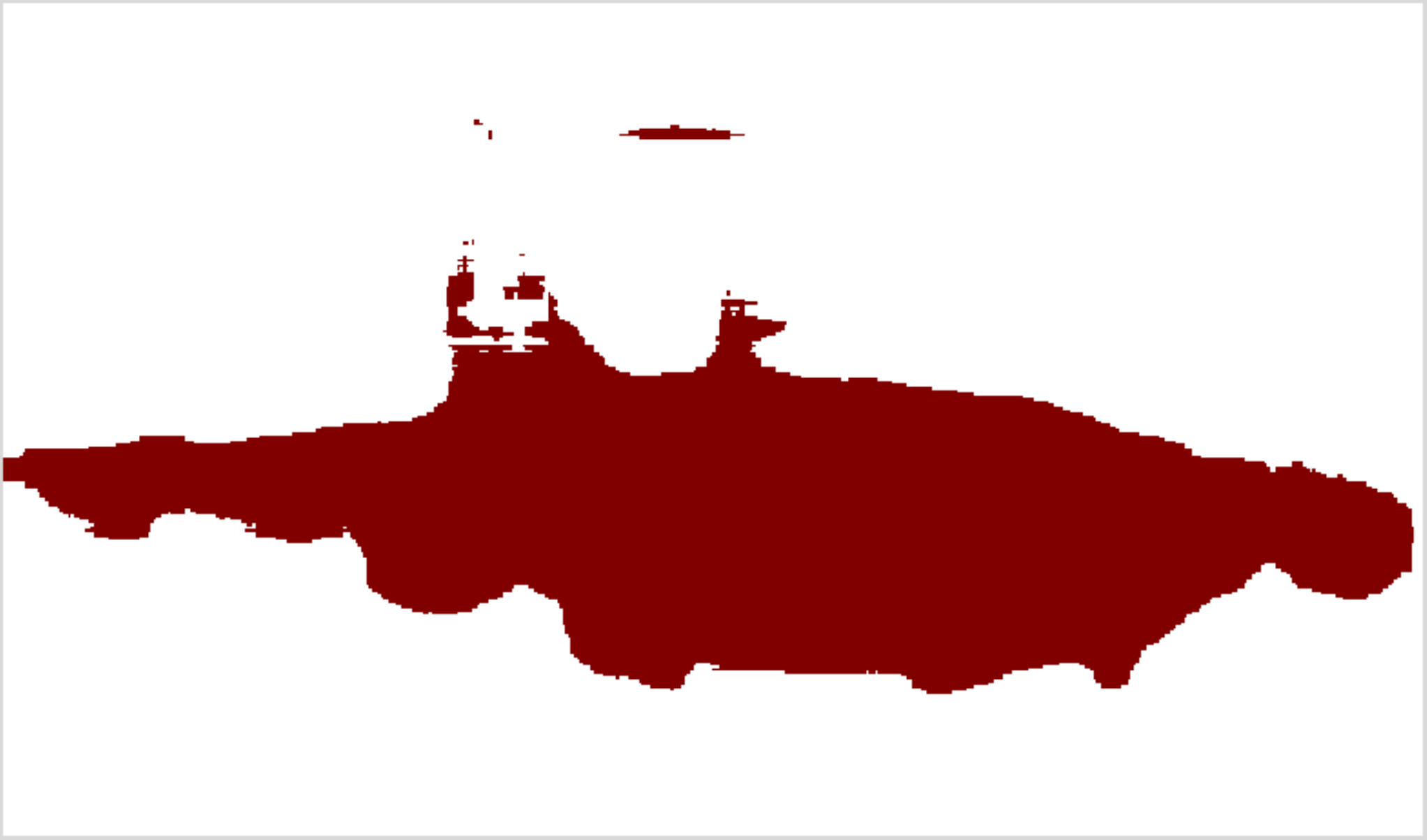}}\hspace{0.1mm}
    \subfloat{\includegraphics[width=2cm, height=1.6cm]{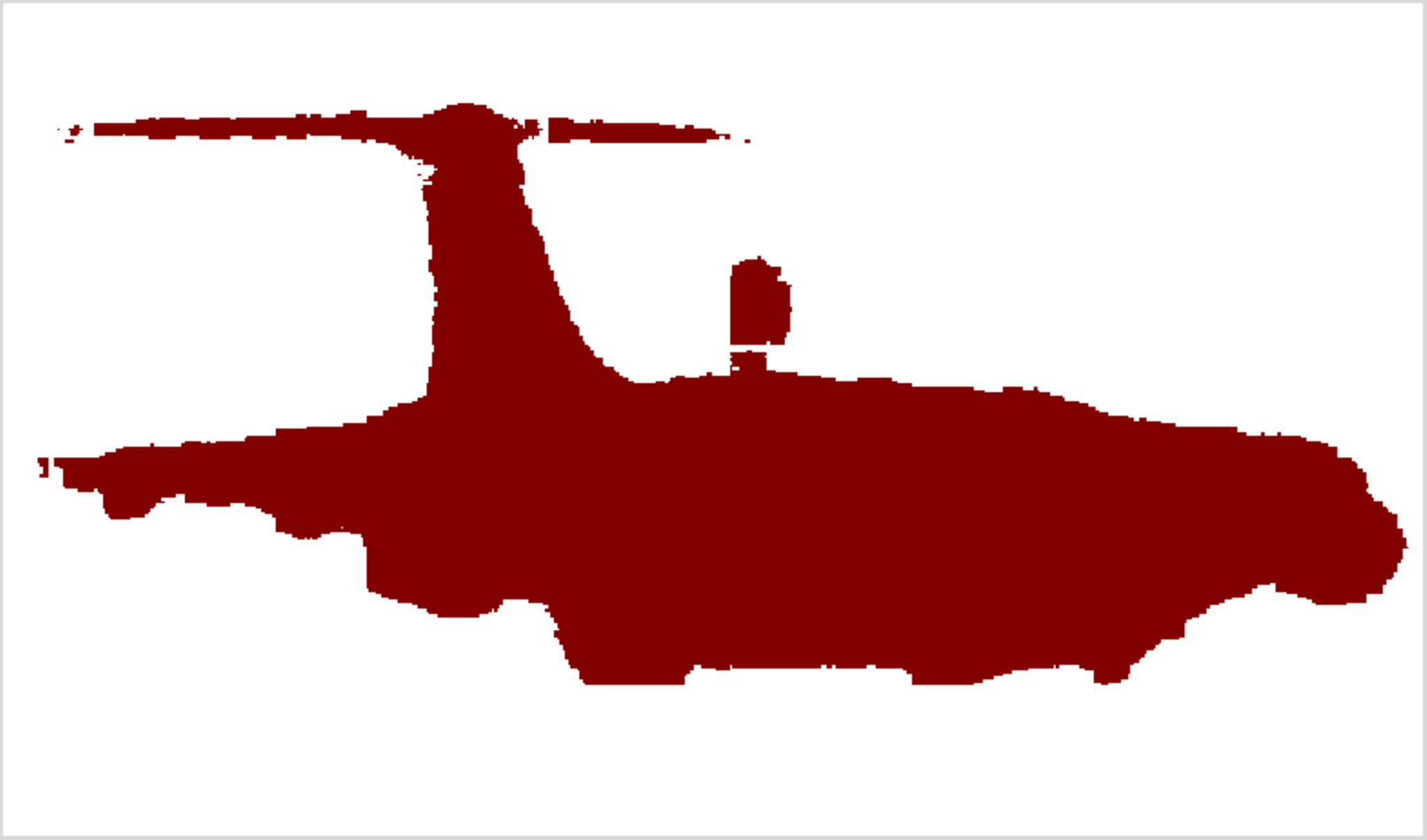}}\\\vspace{0.1mm}
    \subfloat{\includegraphics[width=2cm, height=1.6cm]{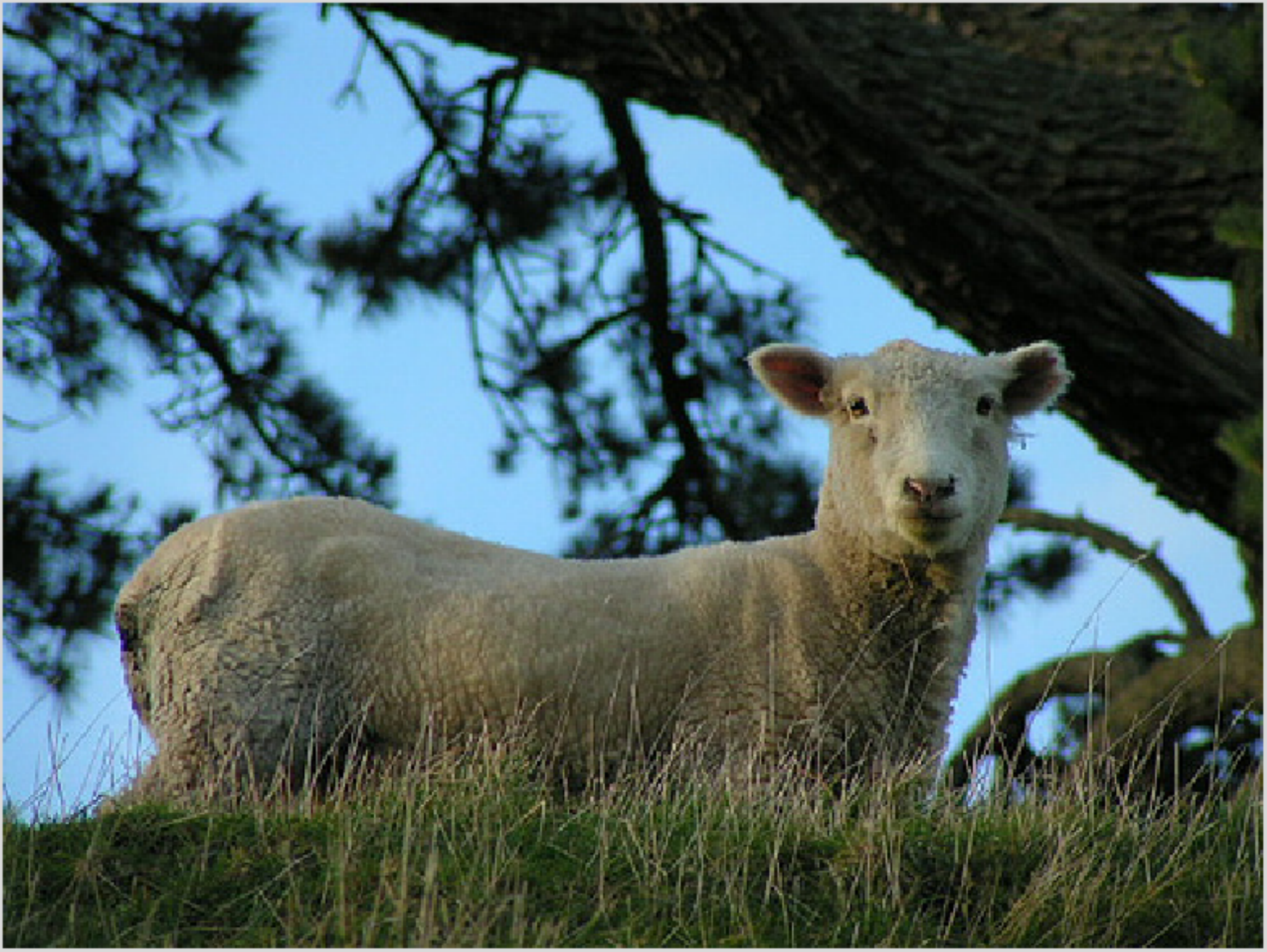}}\hspace{0.1mm}
    \subfloat{\includegraphics[width=2cm, height=1.6cm]{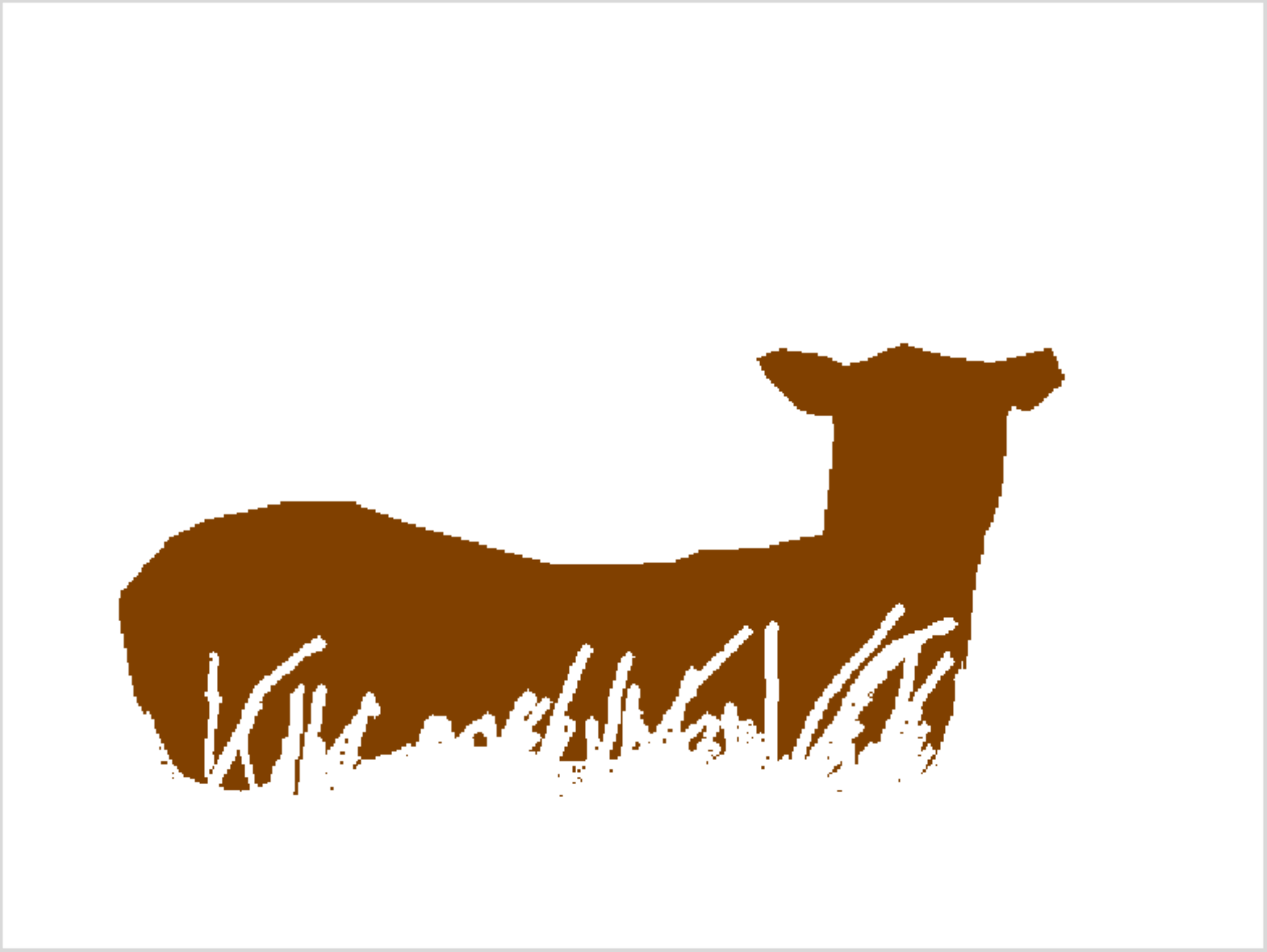}}\hspace{0.1mm}
    \subfloat{\includegraphics[width=2cm, height=1.6cm]{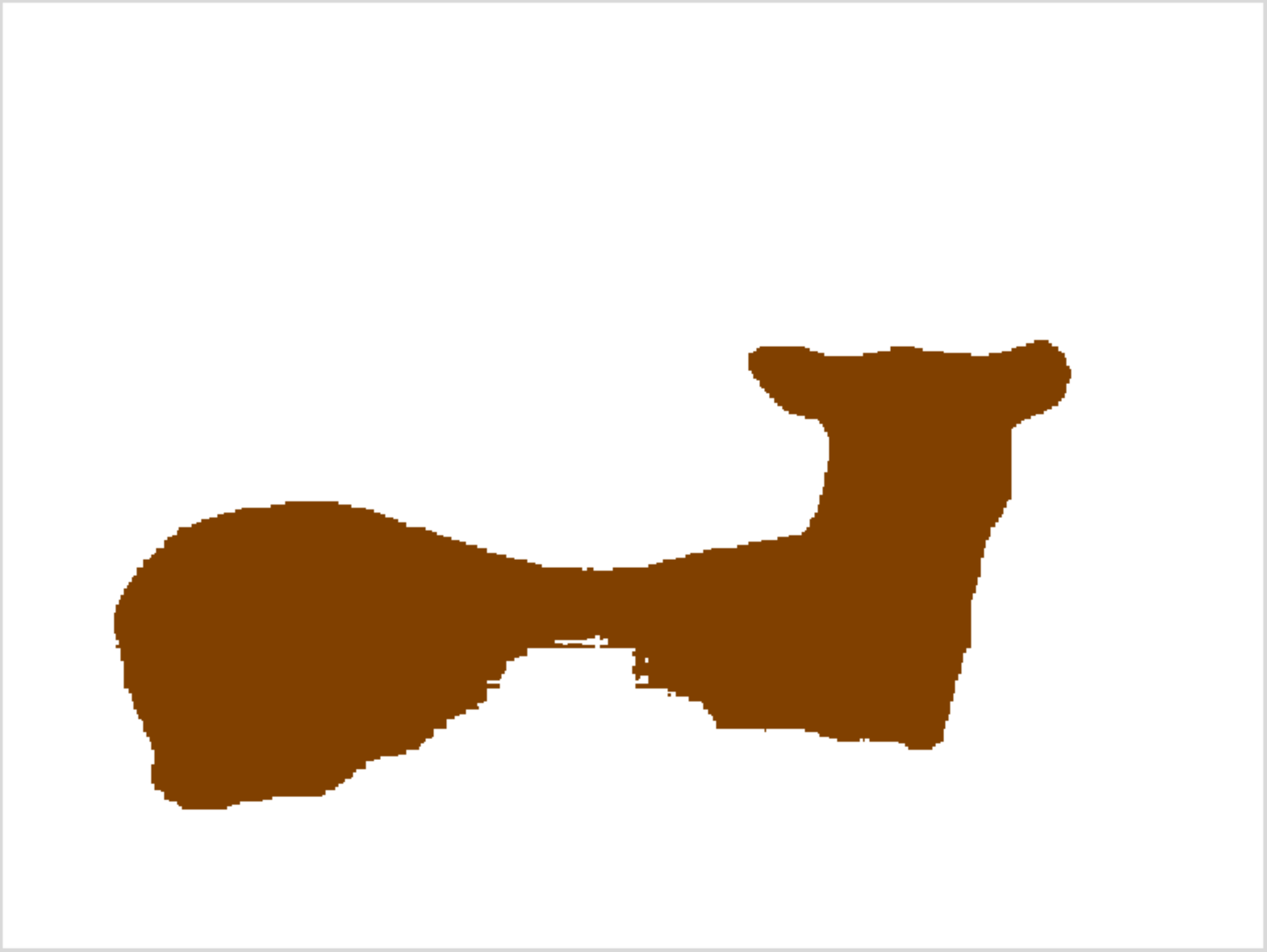}}\hspace{0.1mm}
    \subfloat{\includegraphics[width=2cm, height=1.6cm]{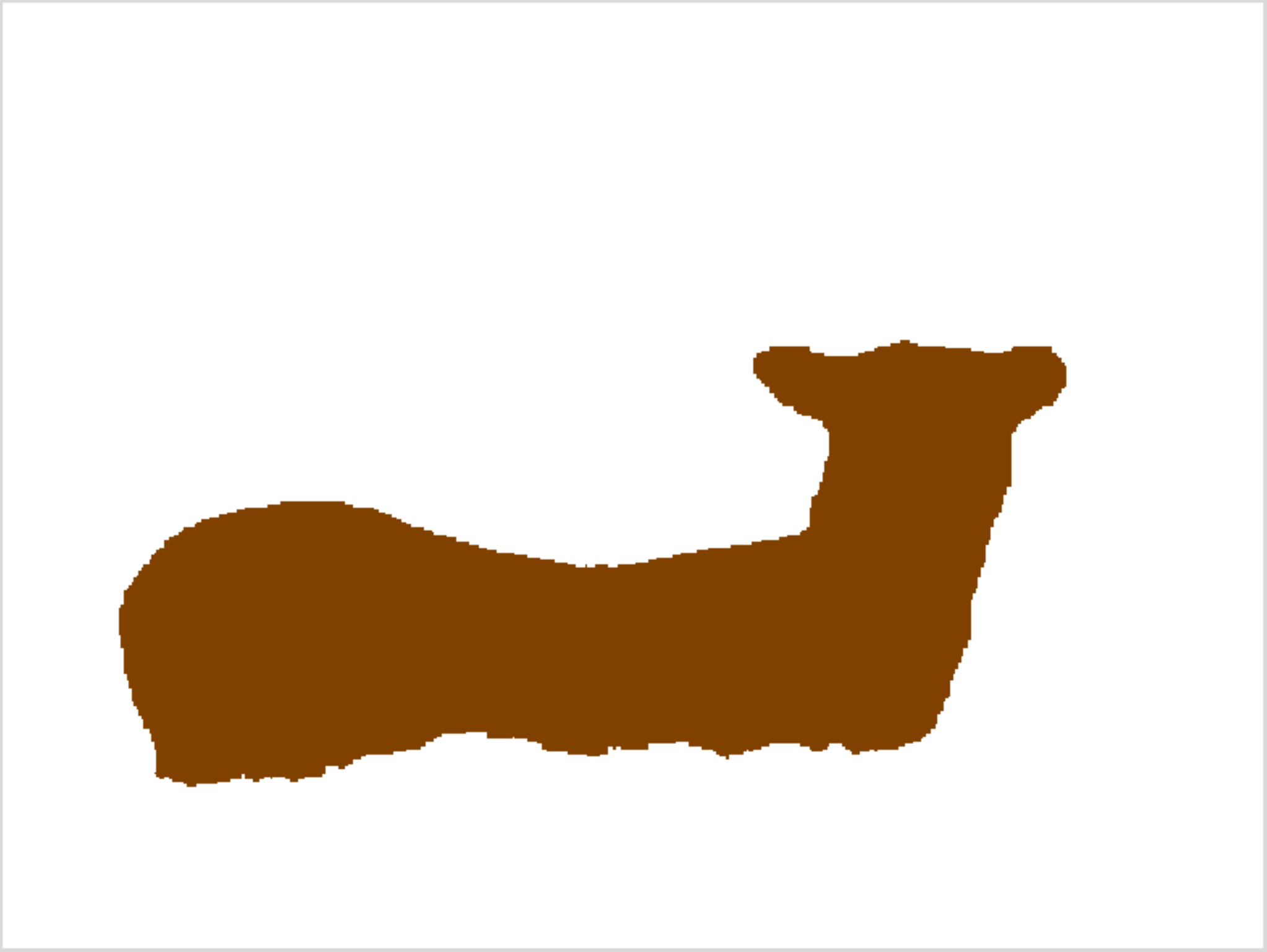}}\\\vspace{0.1mm}
    \subfloat{\includegraphics[width=2cm, height=1.6cm]{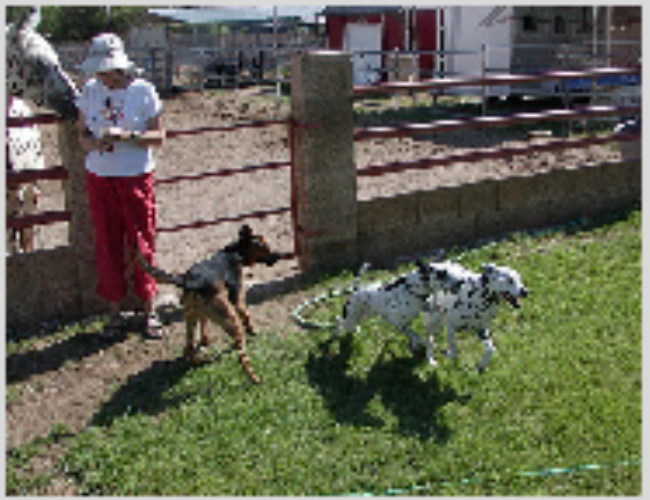}}\hspace{0.1mm}
    \subfloat{\includegraphics[width=2cm, height=1.6cm]{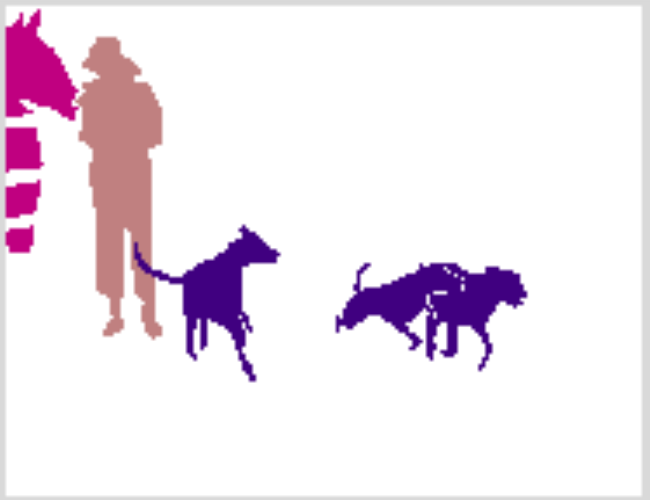}}\hspace{0.1mm}
    \subfloat{\includegraphics[width=2cm, height=1.6cm]{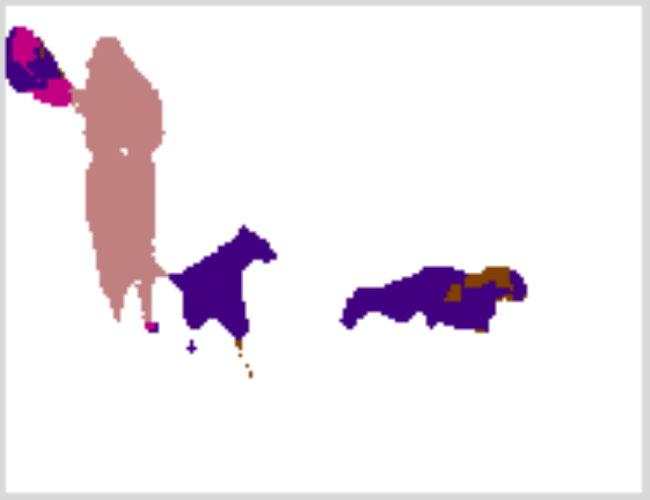}}\hspace{0.1mm}
    \subfloat{\includegraphics[width=2cm, height=1.6cm]{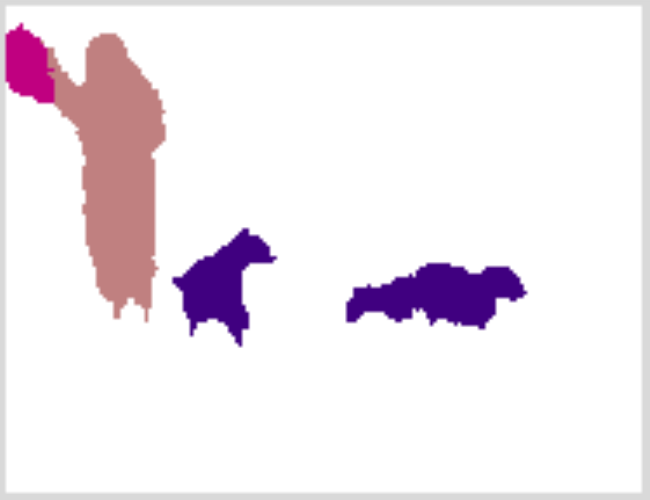}}\\\vspace{0.1mm}
    \subfloat{\includegraphics[width=2cm, height=1.6cm]{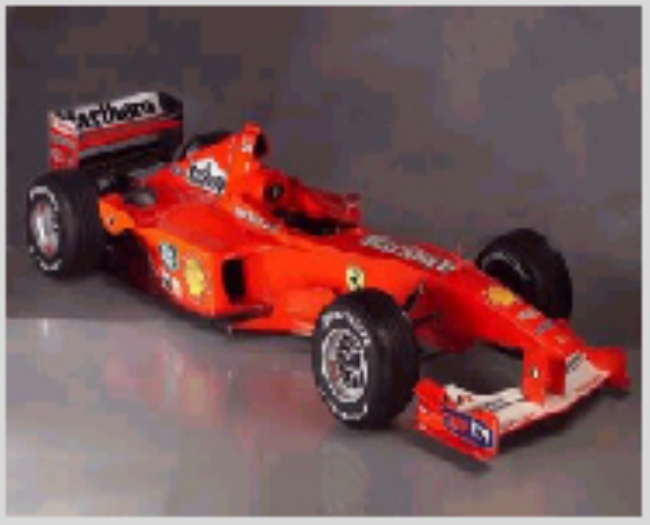}}\hspace{0.1mm}
    \subfloat{\includegraphics[width=2cm, height=1.6cm]{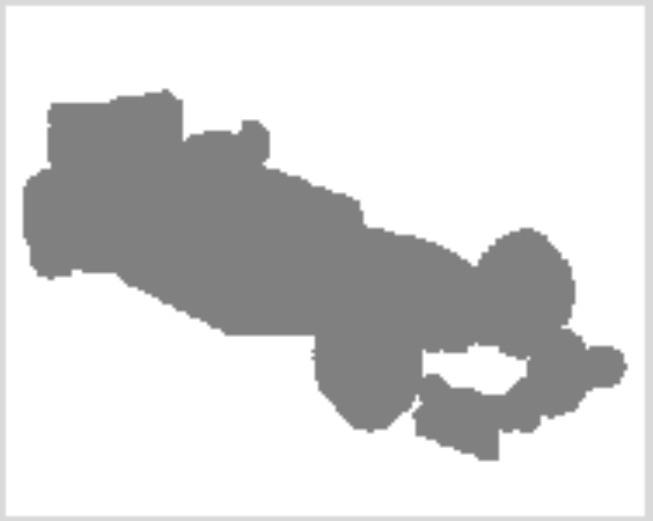}}\hspace{0.1mm}
    \subfloat{\includegraphics[width=2cm, height=1.6cm]{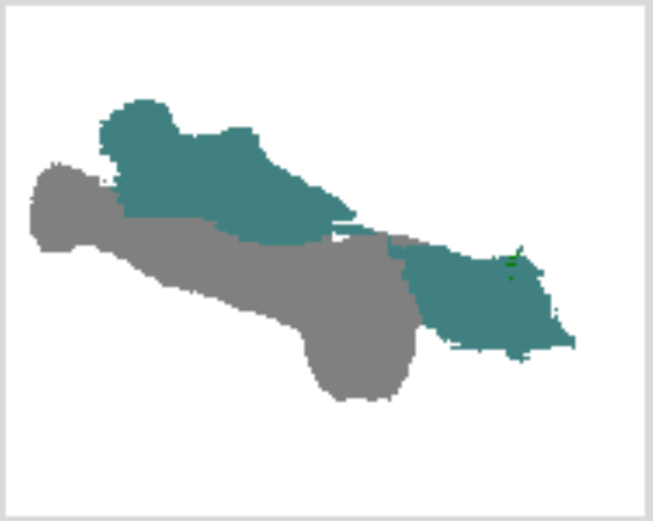}}\hspace{0.1mm}
    \subfloat{\includegraphics[width=2cm, height=1.6cm]{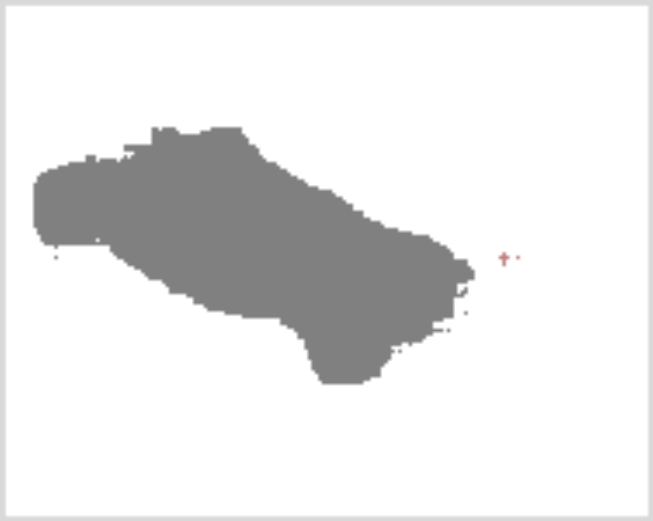}}\\\vspace{0.1mm}
    \subfloat{\includegraphics[width=2cm, height=1.6cm]{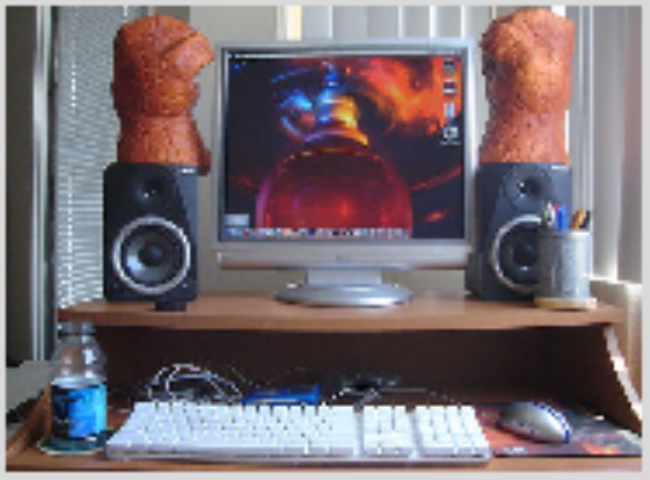}}\hspace{0.1mm}
    \subfloat{\includegraphics[width=2cm, height=1.6cm]{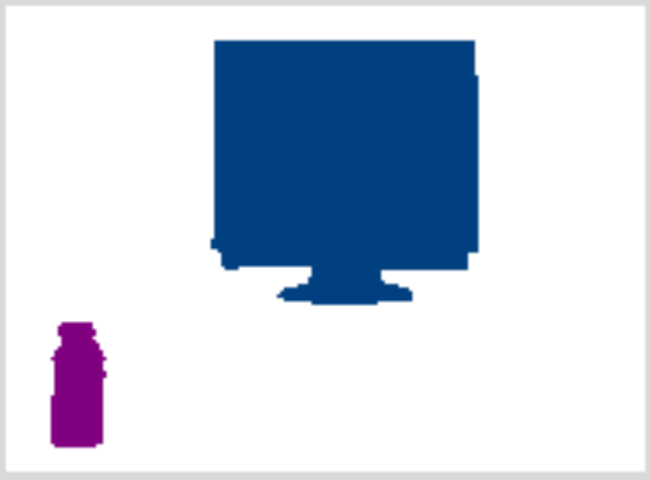}}\hspace{0.1mm}
    \subfloat{\includegraphics[width=2cm, height=1.6cm]{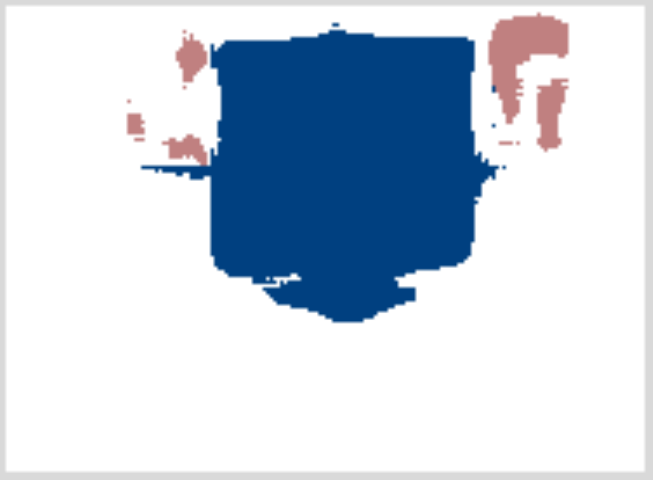}}\hspace{0.1mm}
    \subfloat{\includegraphics[width=2cm, height=1.6cm]{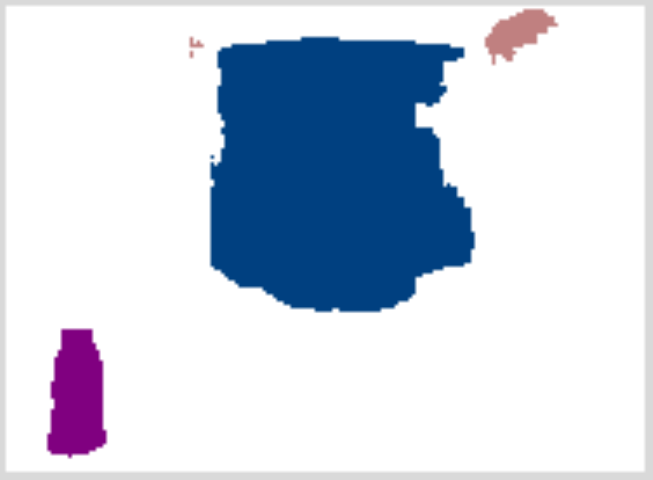}}\\\vspace{0.1mm}
    \subfloat{\includegraphics[width=2cm, height=1.6cm]{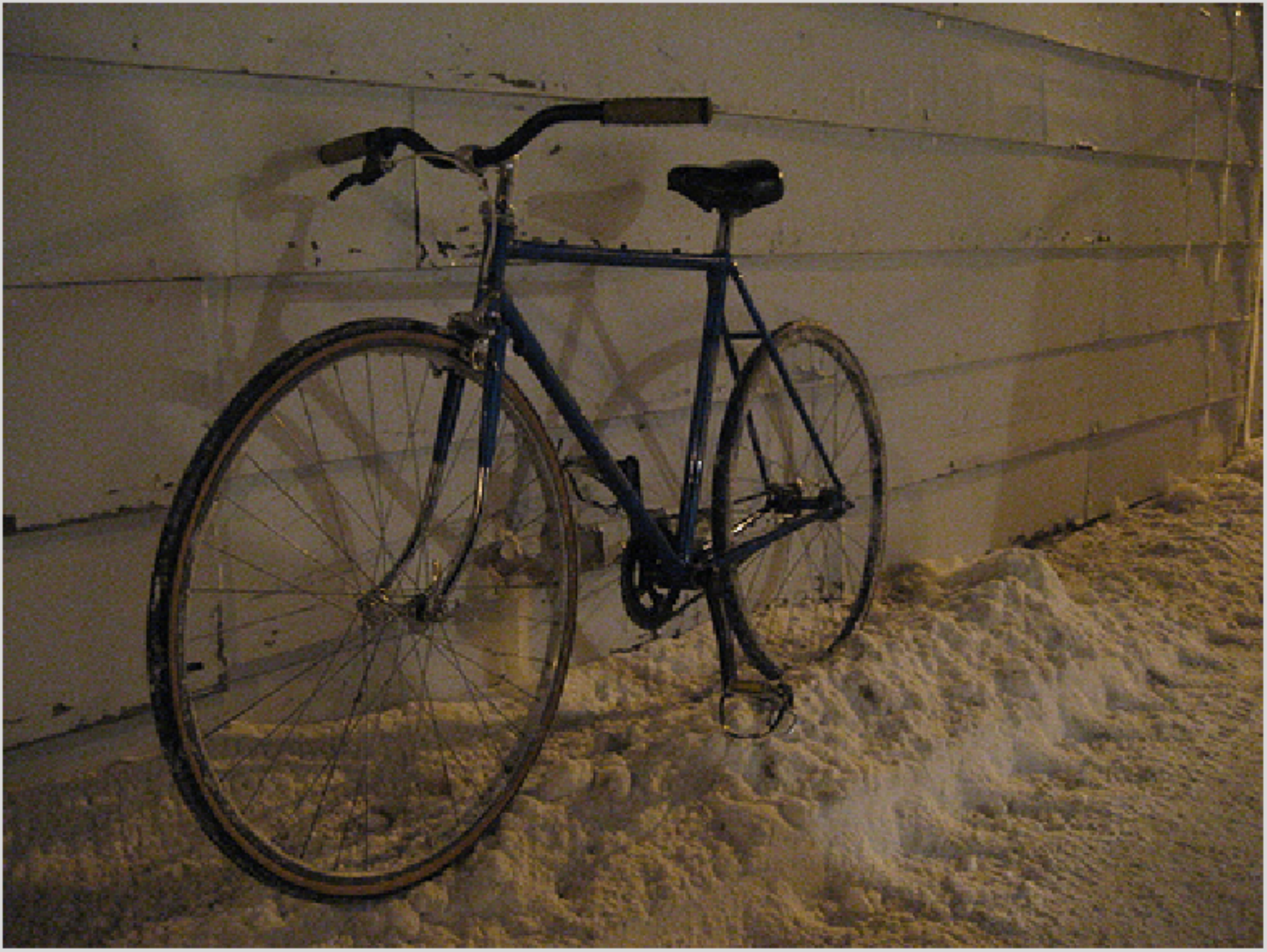}}\hspace{0.1mm}
    \subfloat{\includegraphics[width=2cm, height=1.6cm]{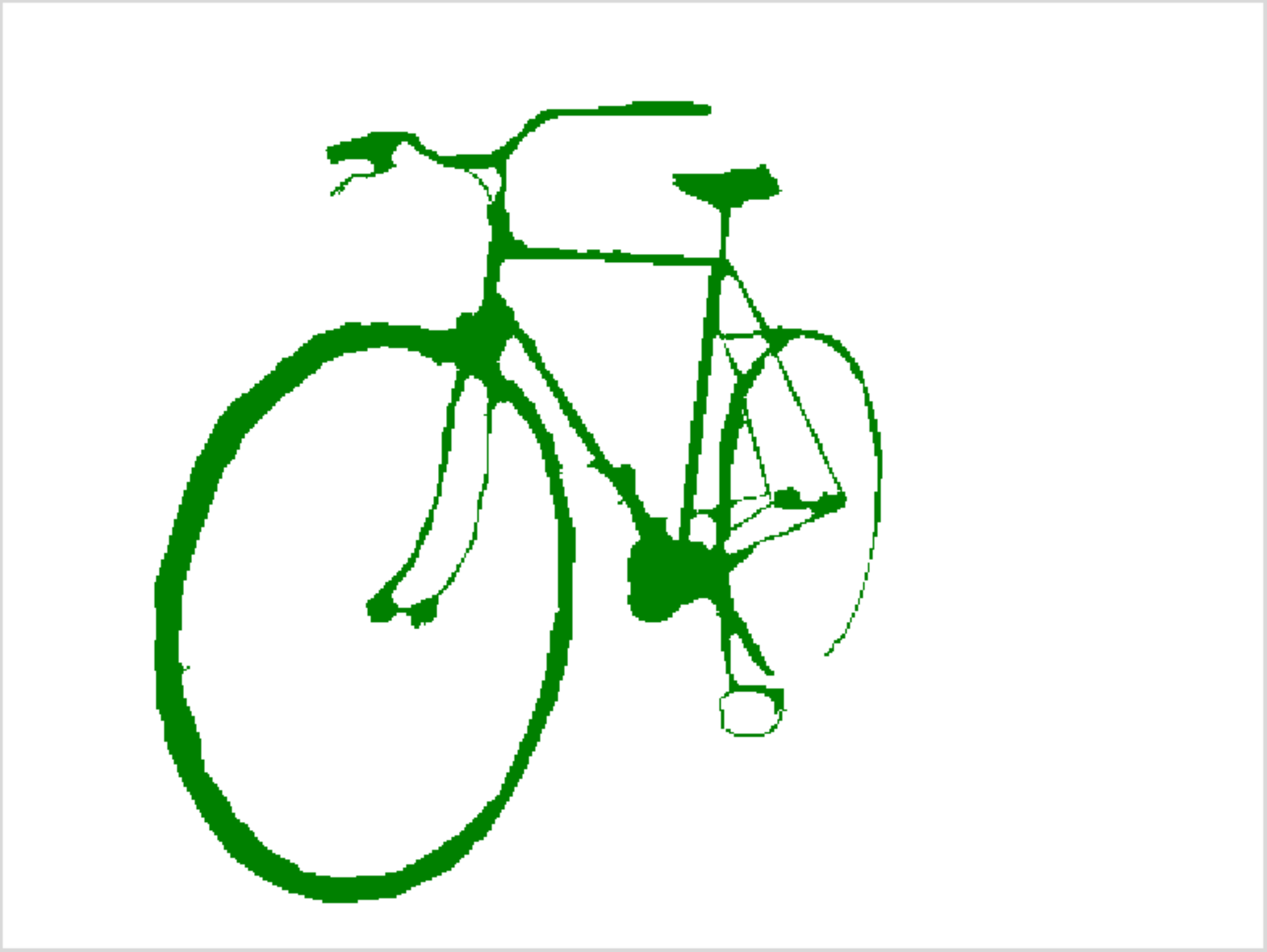}}\hspace{0.1mm}
    \subfloat{\includegraphics[width=2cm, height=1.6cm]{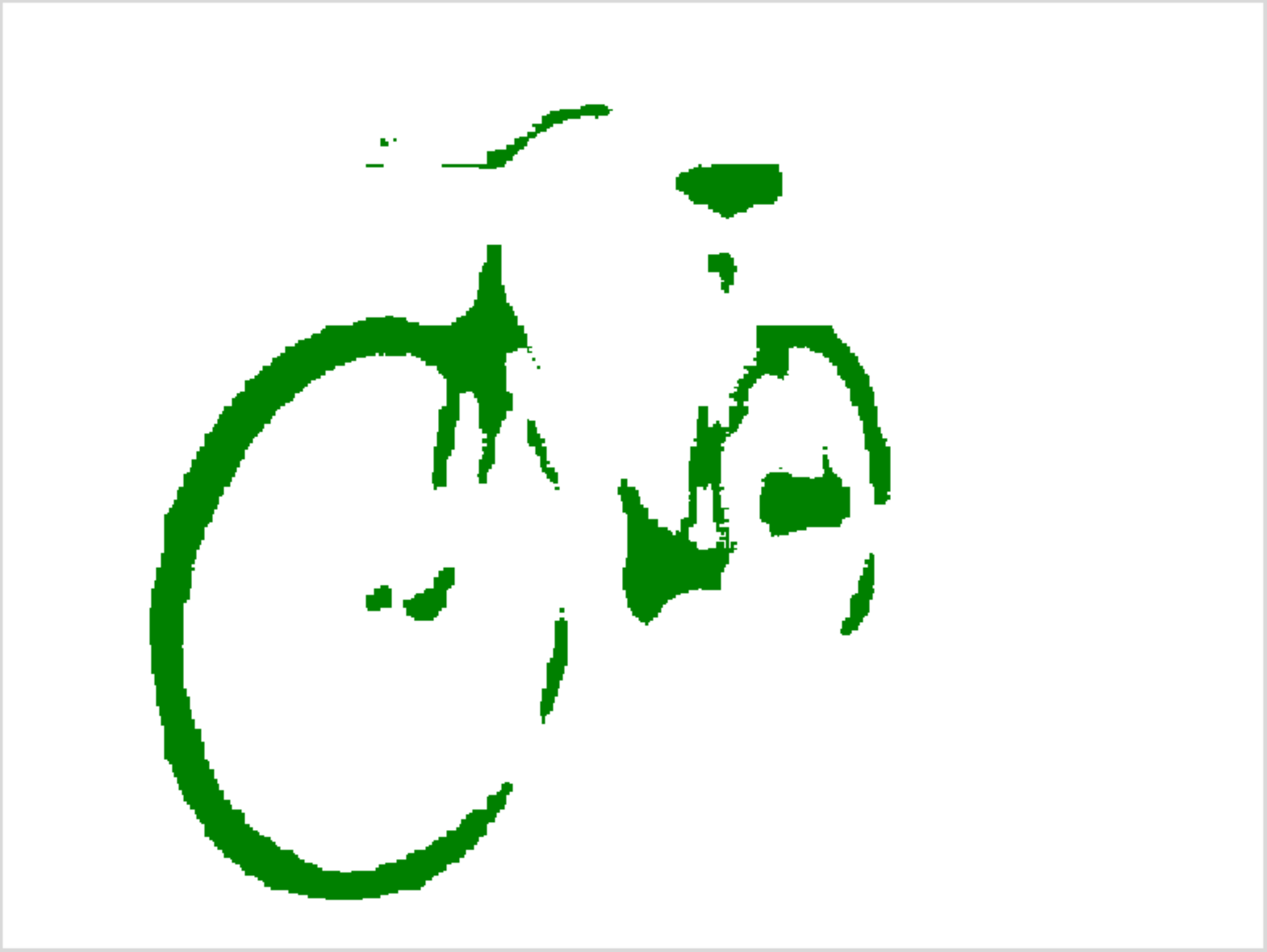}}\hspace{0.1mm}
    \subfloat{\includegraphics[width=2cm, height=1.6cm]{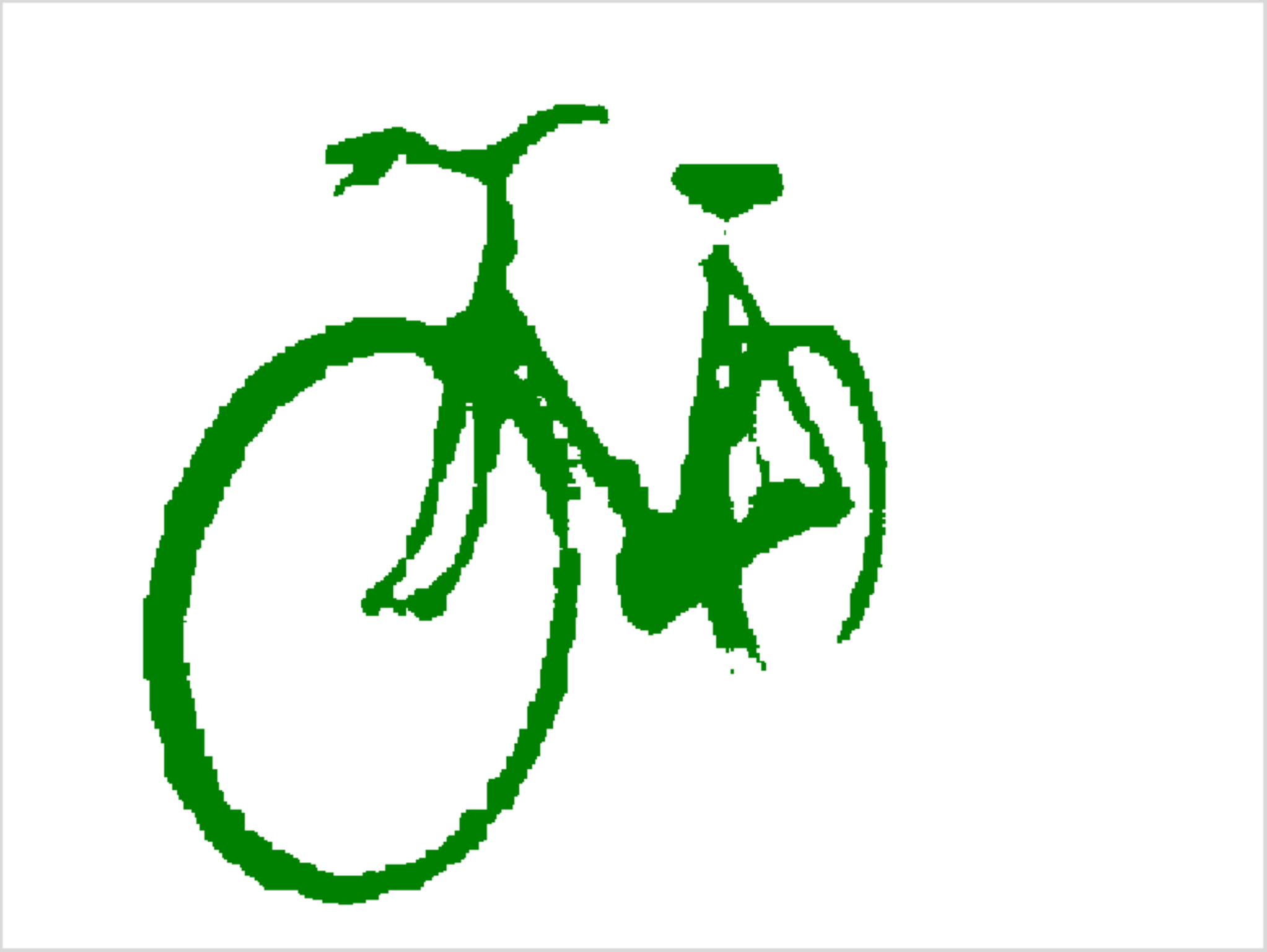}}\\\vspace{0.1mm}
    \subfloat{\includegraphics[width=2cm, height=1.6cm]{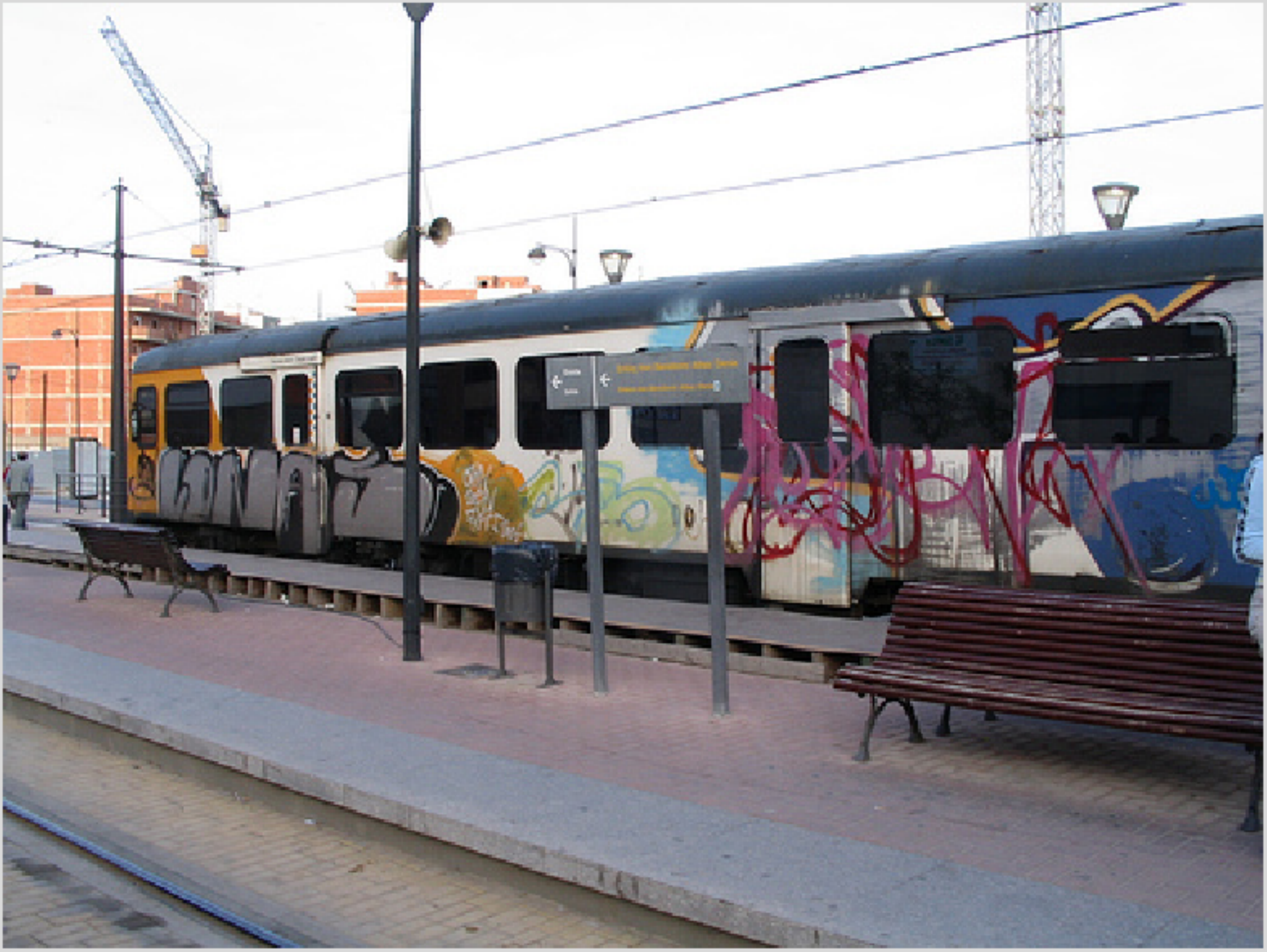}}\hspace{0.1mm}
    \subfloat{\includegraphics[width=2cm, height=1.6cm]{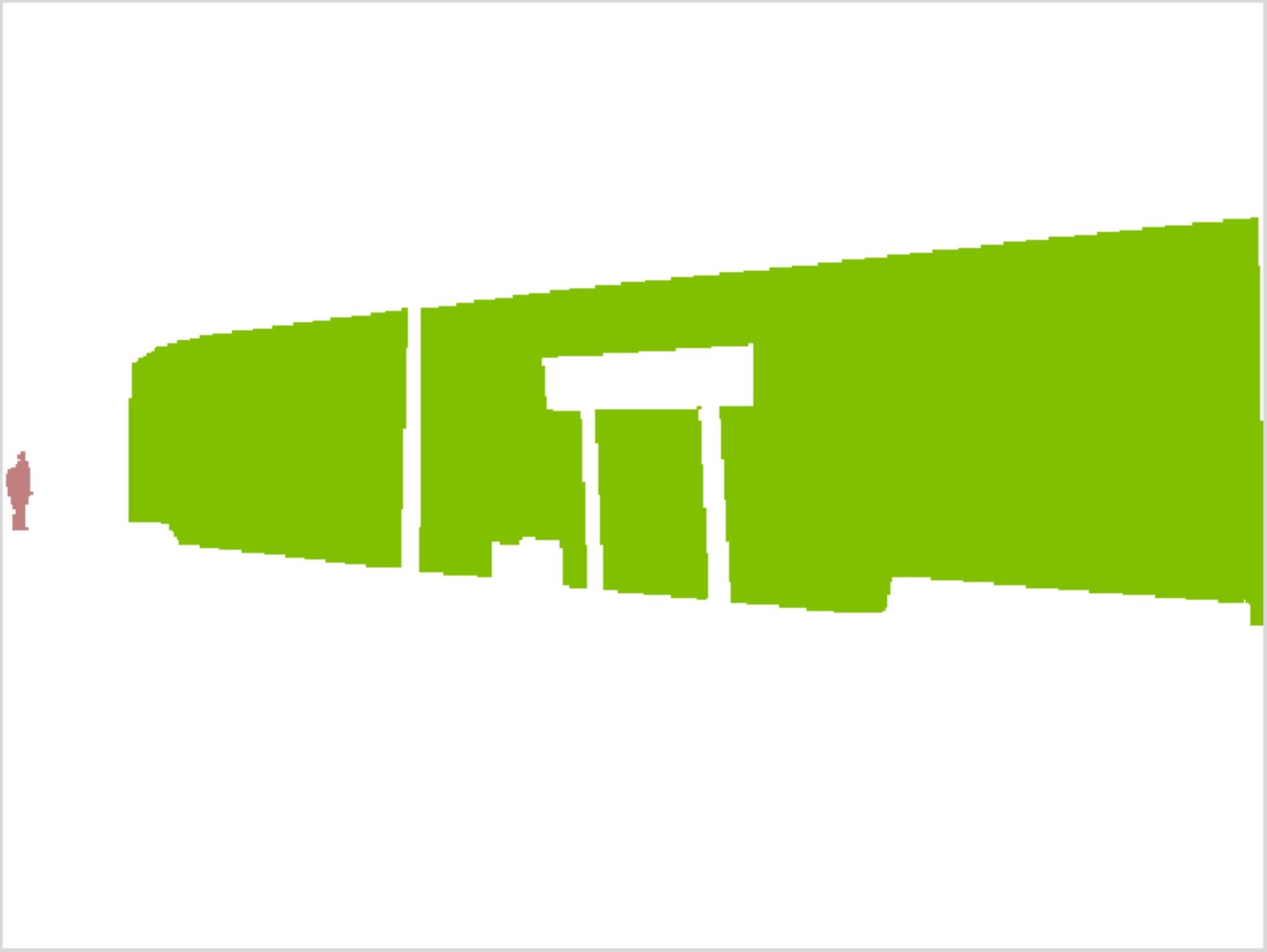}}\hspace{0.1mm}
    \subfloat{\includegraphics[width=2cm, height=1.6cm]{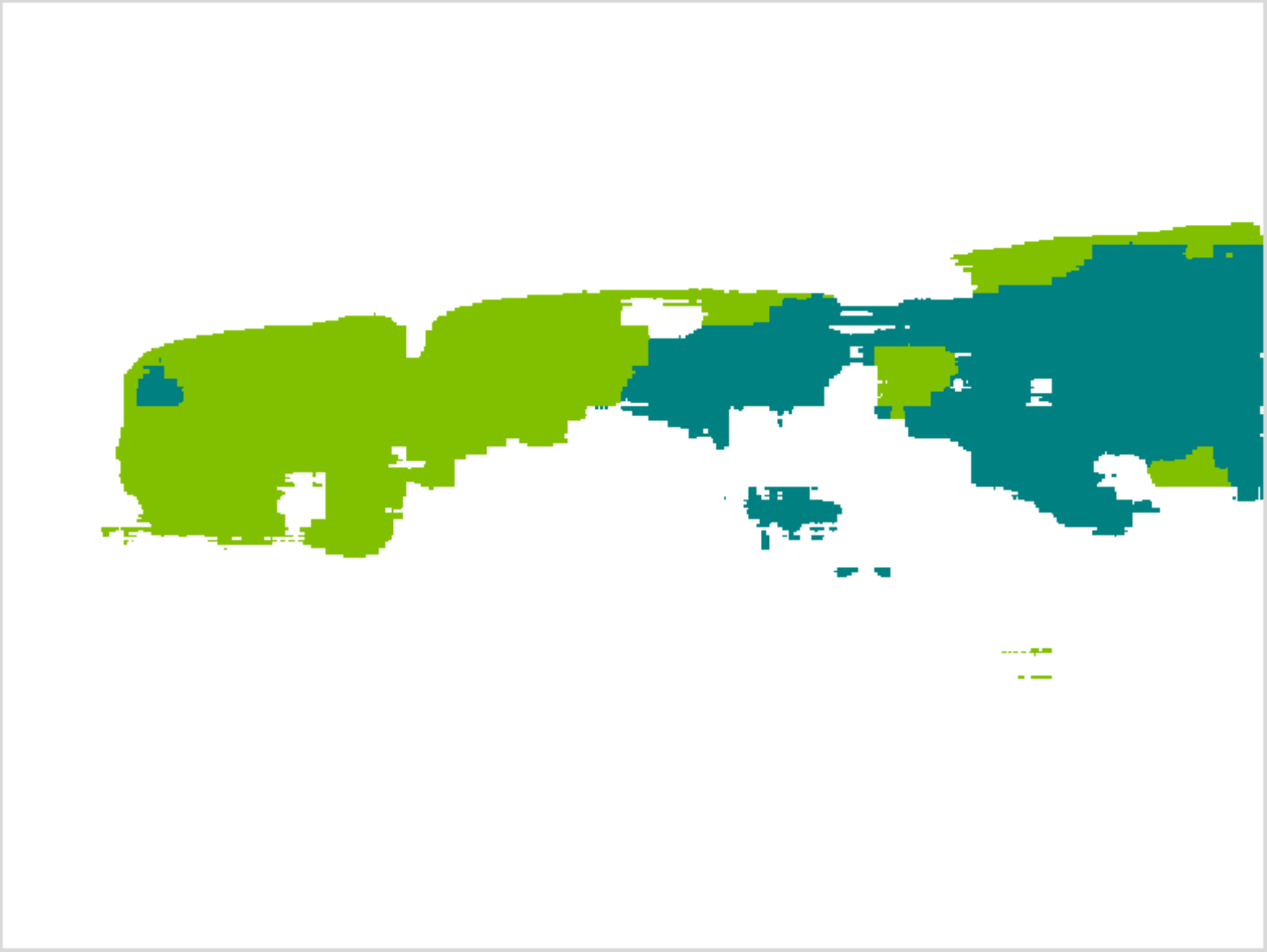}}\hspace{0.1mm}
    \subfloat{\includegraphics[width=2cm, height=1.6cm]{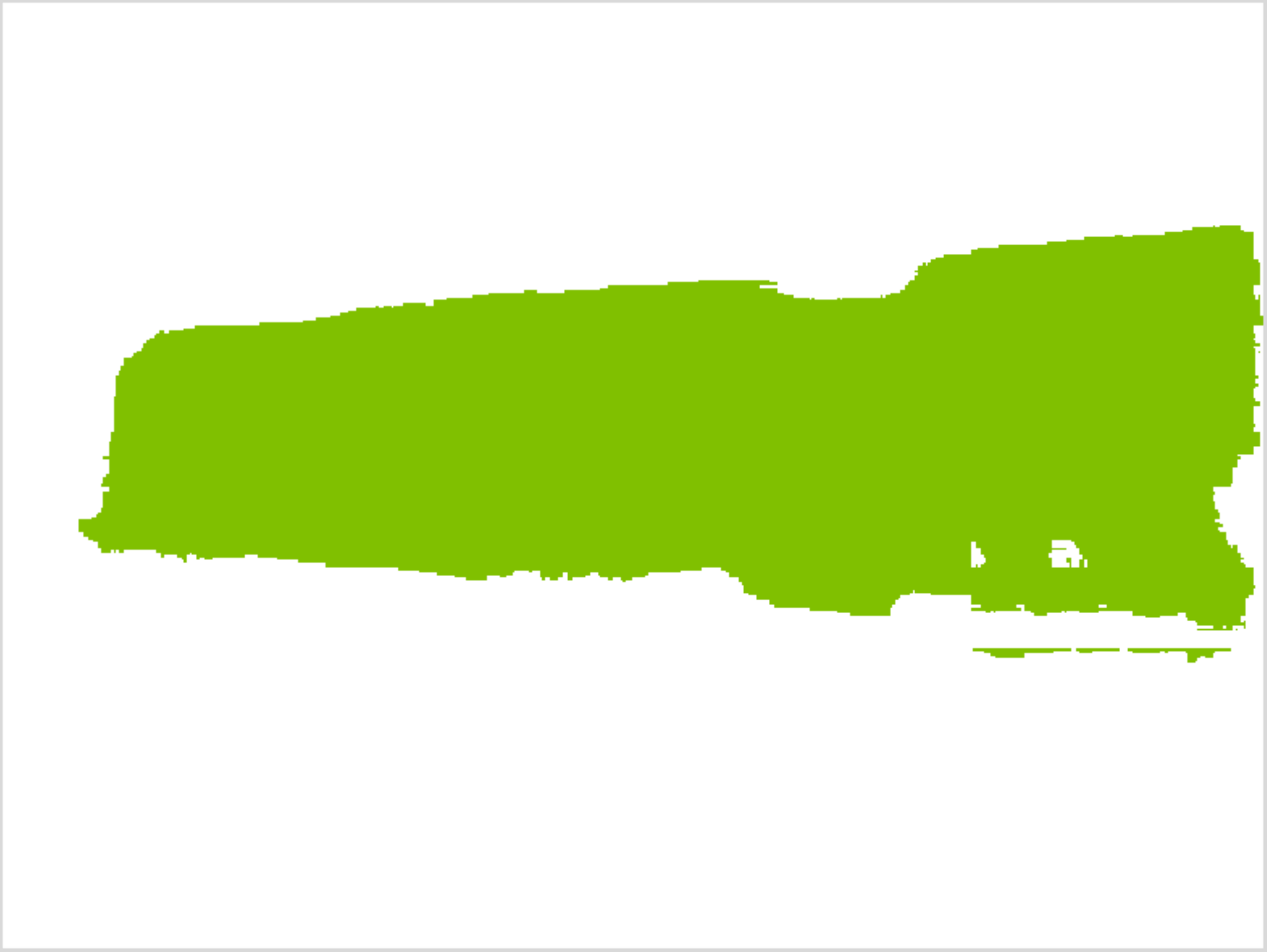}}\\\vspace{0.1mm}
    \subfloat{\includegraphics[width=2cm, height=1.6cm]{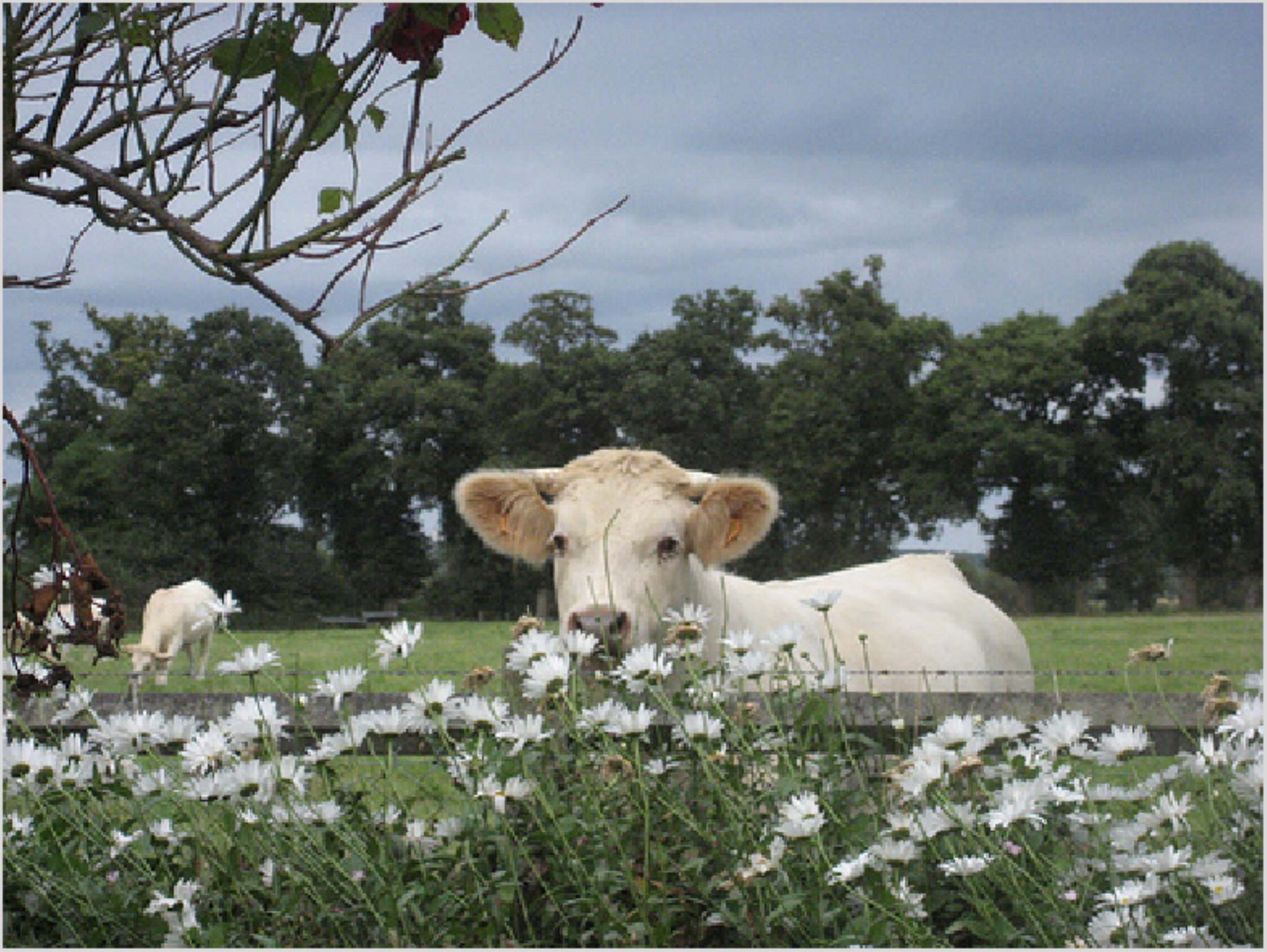}}\hspace{0.1mm}
    \subfloat{\includegraphics[width=2cm, height=1.6cm]{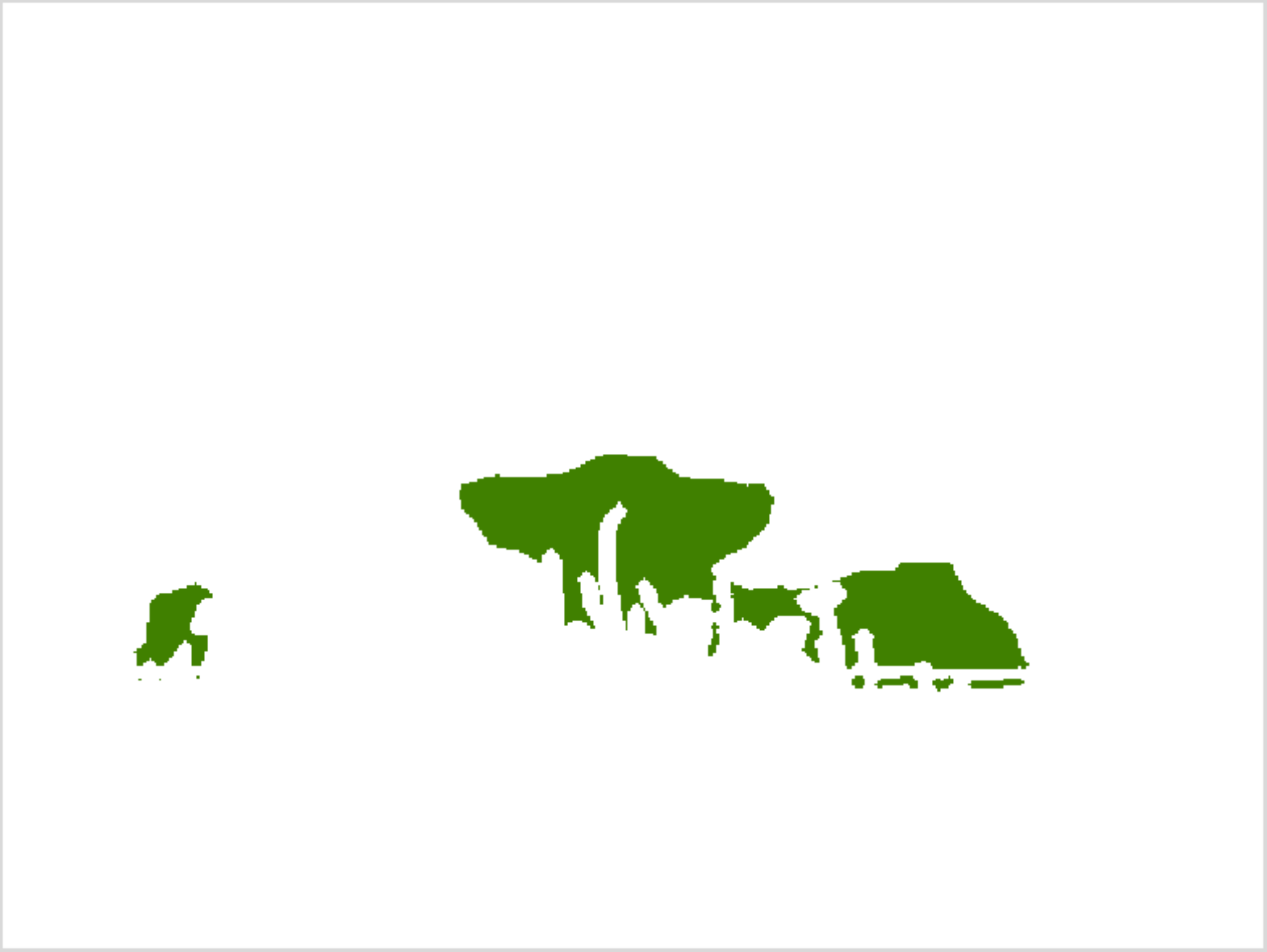}}\hspace{0.1mm}
    \subfloat{\includegraphics[width=2cm, height=1.6cm]{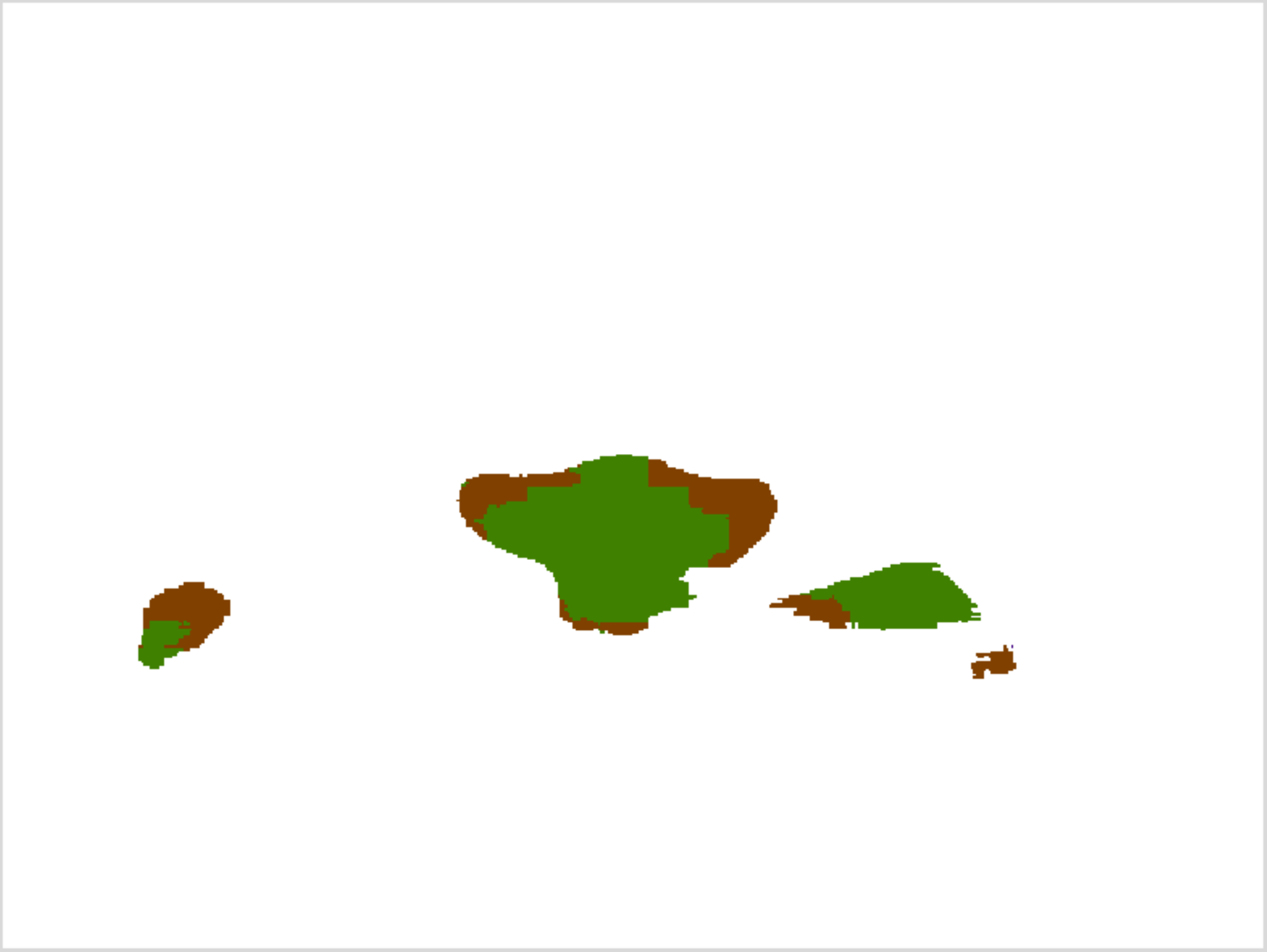}}\hspace{0.1mm}
    \subfloat{\includegraphics[width=2cm, height=1.6cm]{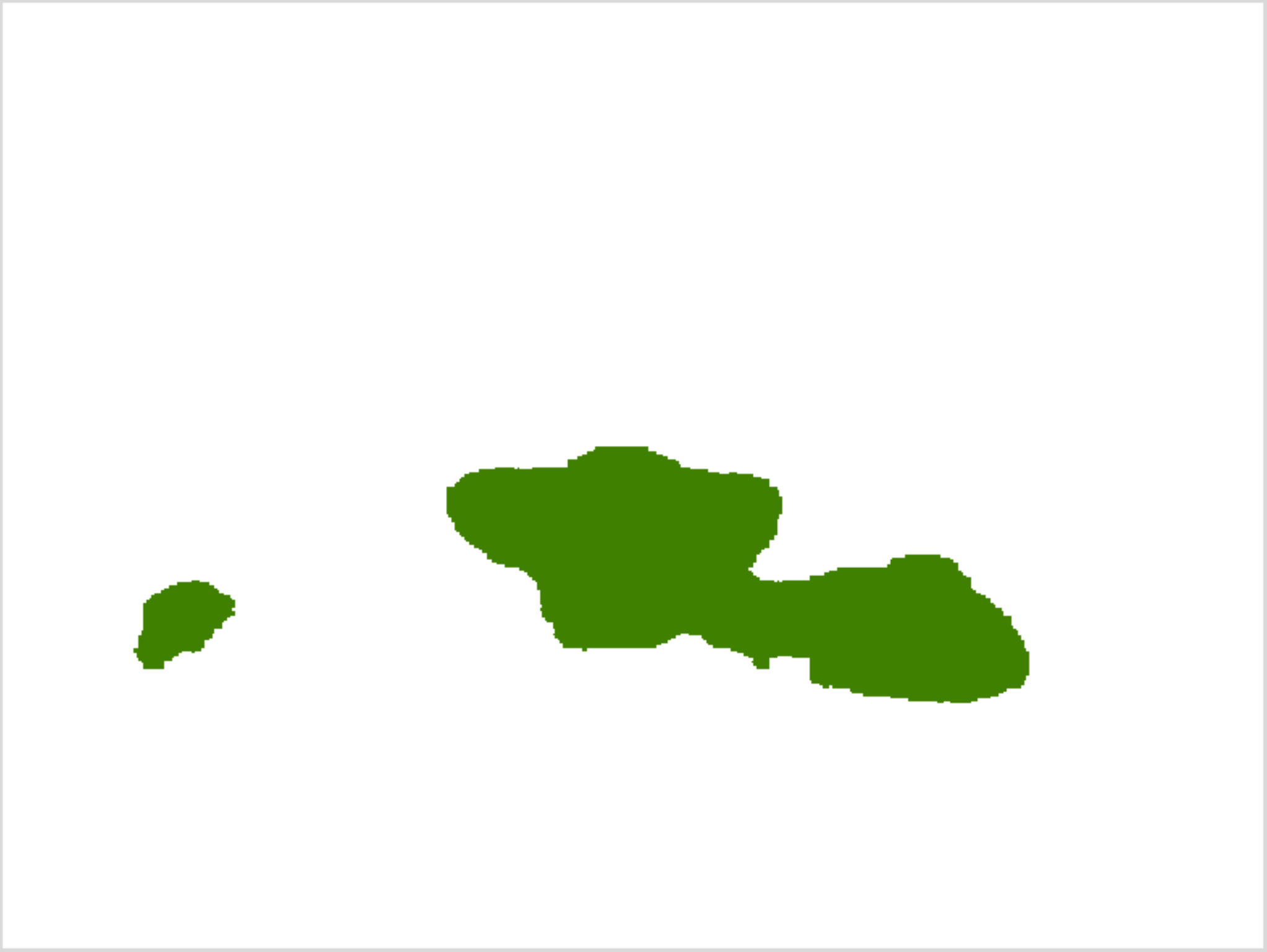}}\\\vspace{0.1mm}
    \setcounter{subfigure}{0}
    \subfloat[Input]{\includegraphics[width=2cm, height=1.6cm]{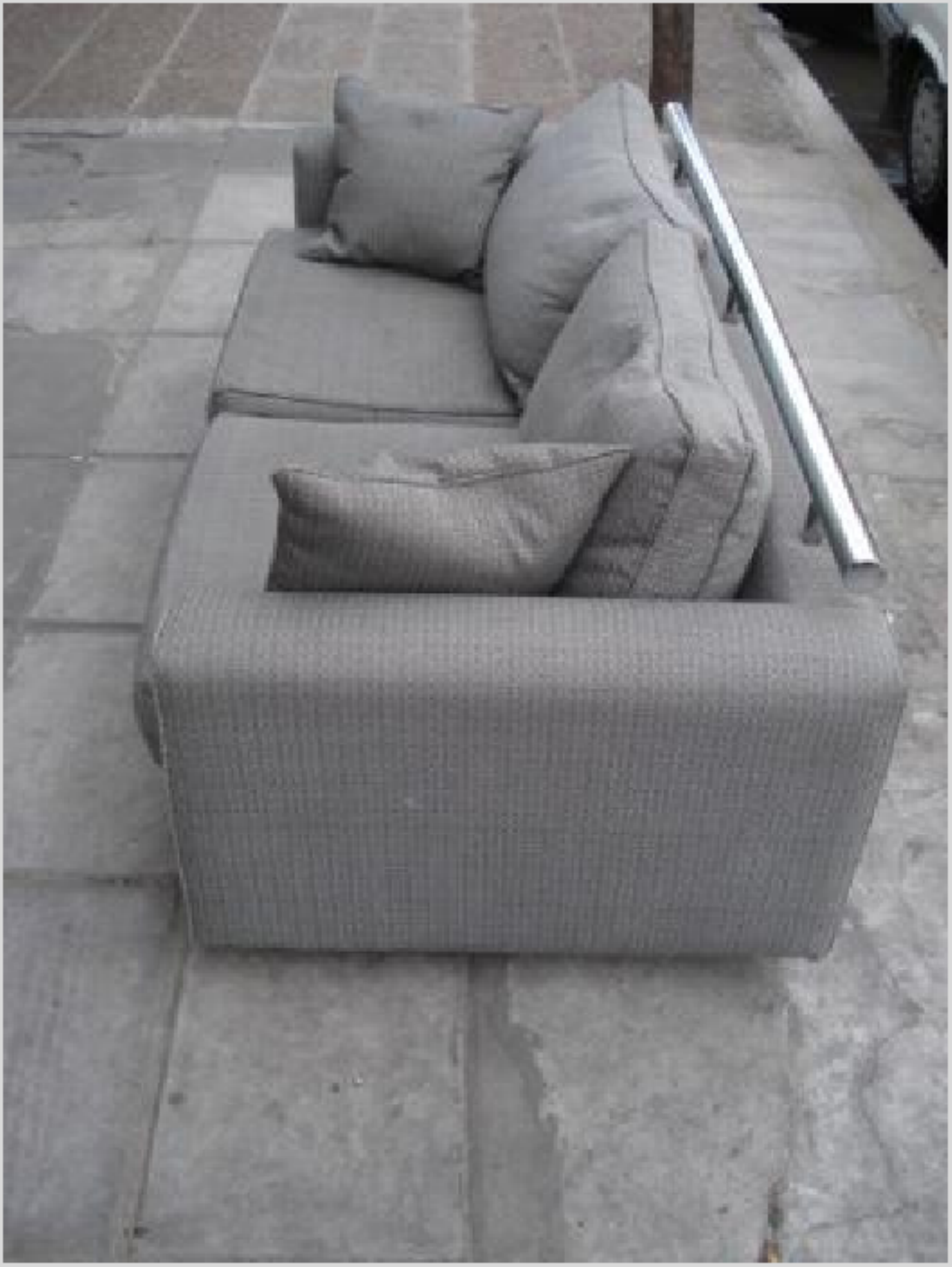}}\hspace{0.1mm}
    \subfloat[GT]{\includegraphics[width=2cm, height=1.6cm]{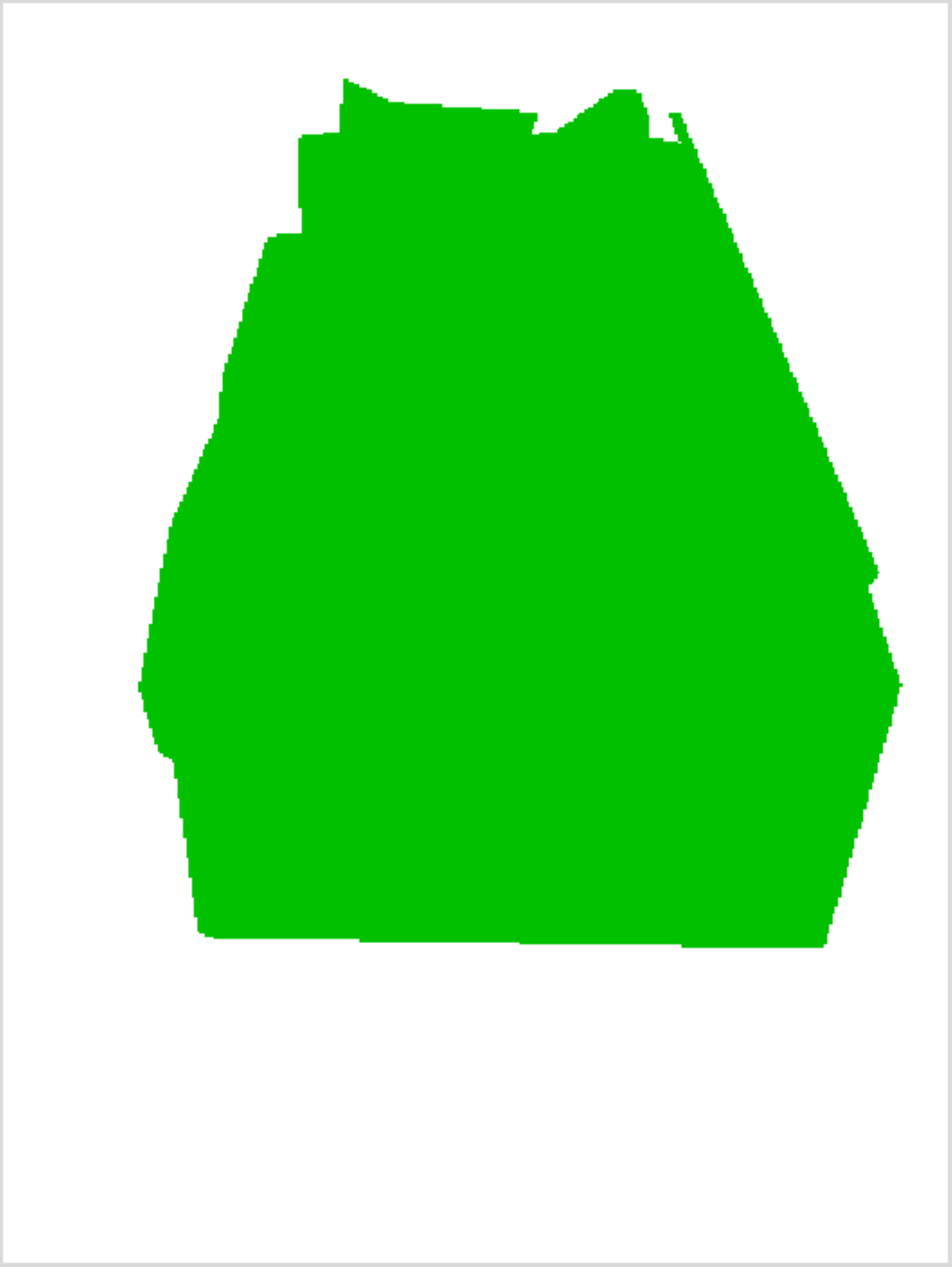}}\hspace{0.1mm}
    \subfloat[CCT]{\includegraphics[width=2cm, height=1.6cm]{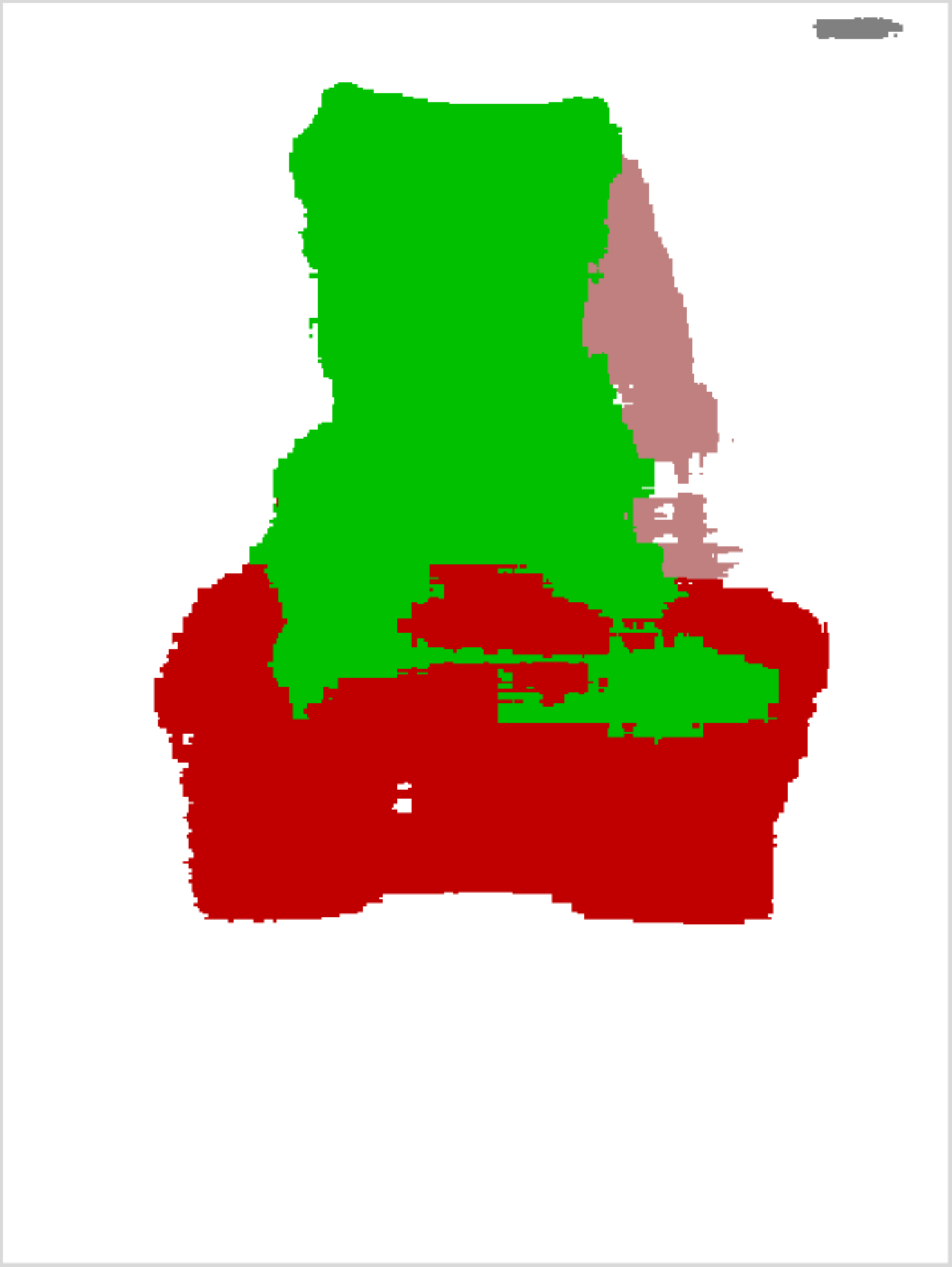}}\hspace{0.1mm}
    \subfloat[Ours]{\includegraphics[width=2cm, height=1.6cm]{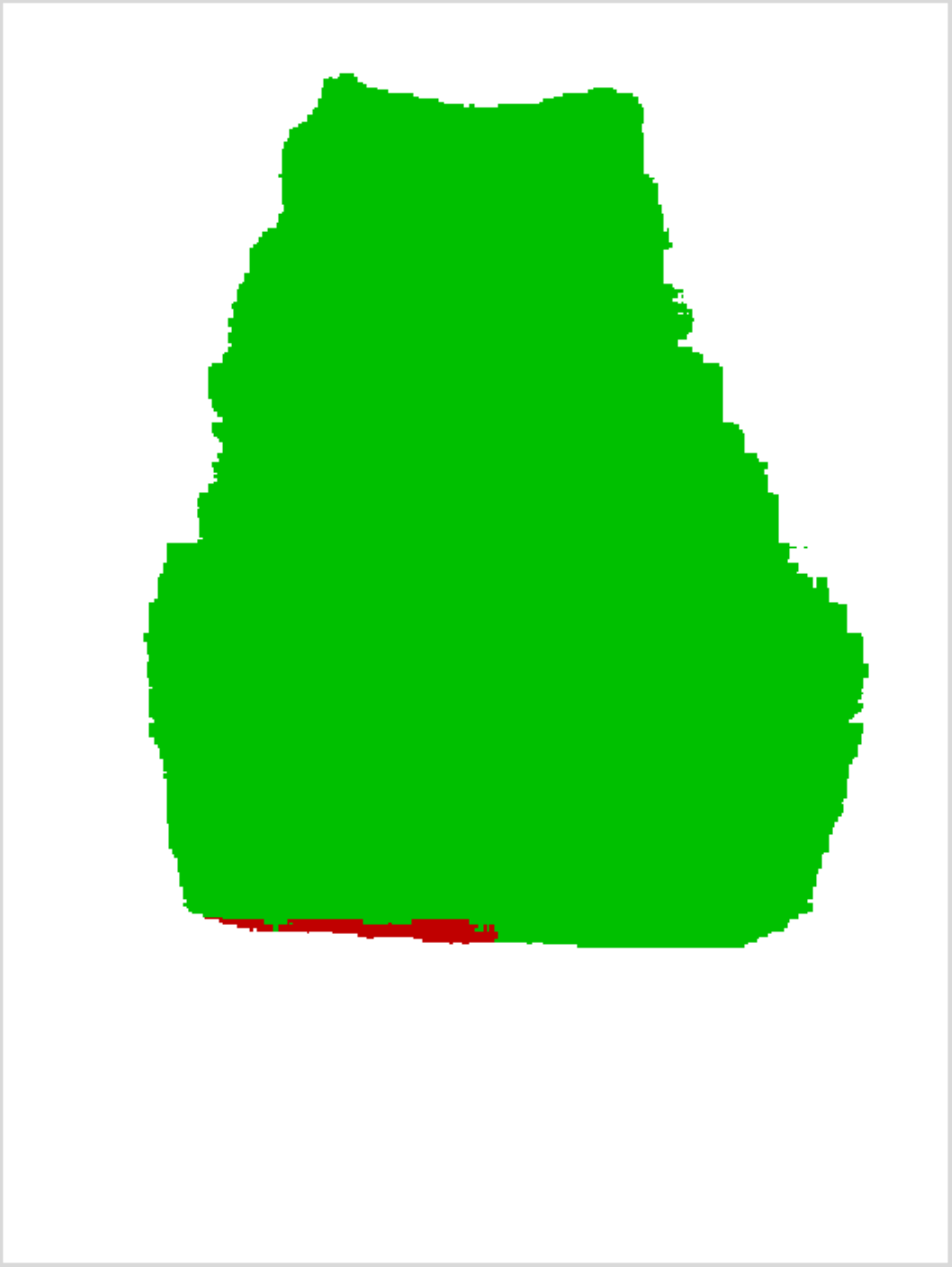}}\\
    \caption{Experimental eesults on Pascal VOC 2012. 
    Compared to CCT, our method is able to predict more complete contours and correct semantics.
    The visual results represent for tiny object parts (the predictions of airplane tails and bicycle handlebars provide more complete contours) and objects occluded by the foreground (shown in the second row).
    In terms of category semantics, results show that CCT easing misinterprets the semantic information of objects under complex environments, while our method achieves better results (an example is shown in the last row).}
    \label{OCCExam}
\end{figure}

The visualization is shown in Fig. \ref{OCCExam}.

\subsection{Appendices E: Quality Visualization Results of Cityscapes}
\begin{figure*}[t!]
    \centering
    \subfloat{\includegraphics[width=4.2cm, height=2.5cm]{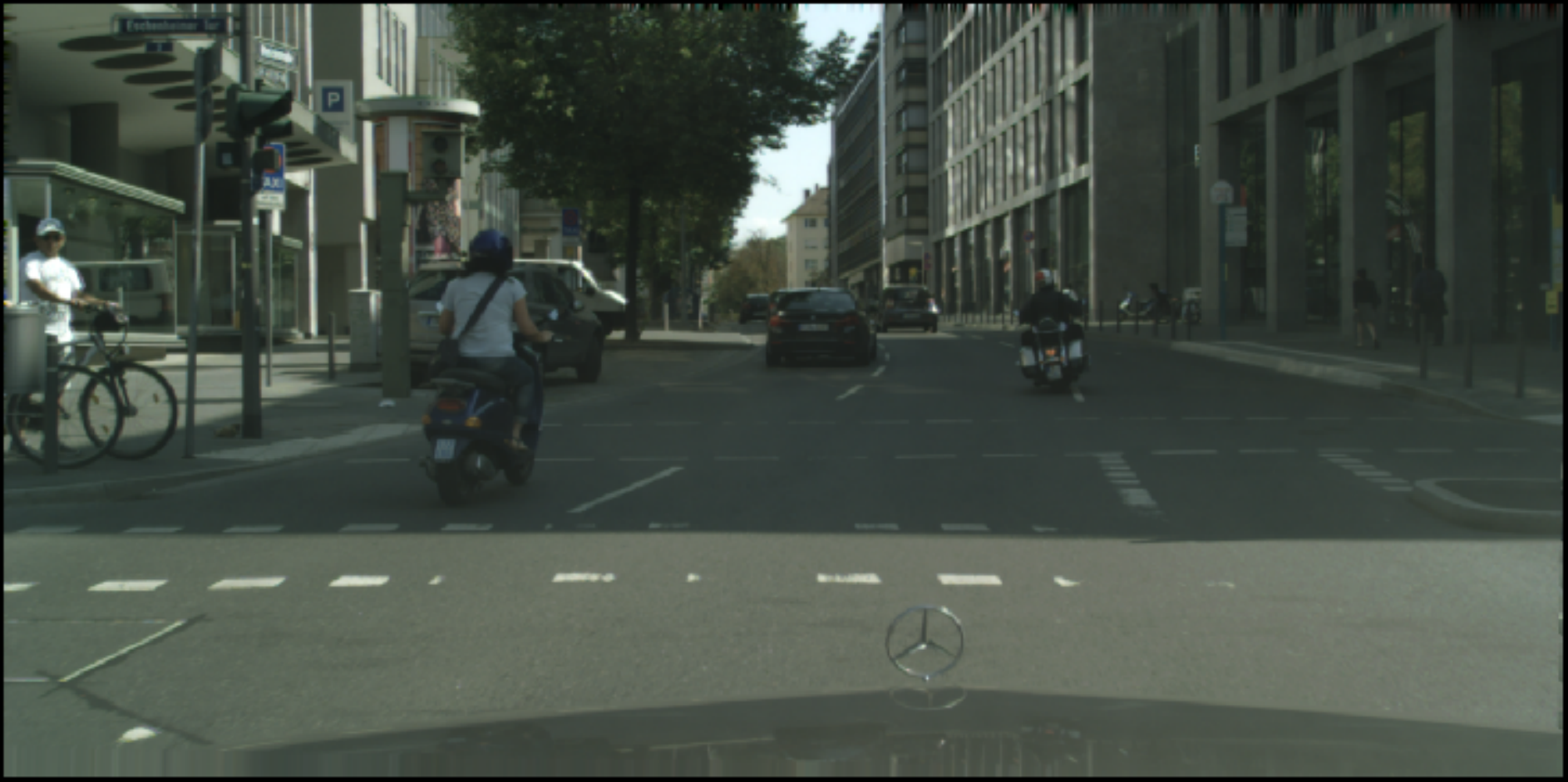}}\hspace{0.1mm}
    \subfloat{\includegraphics[width=4.2cm, height=2.5cm]{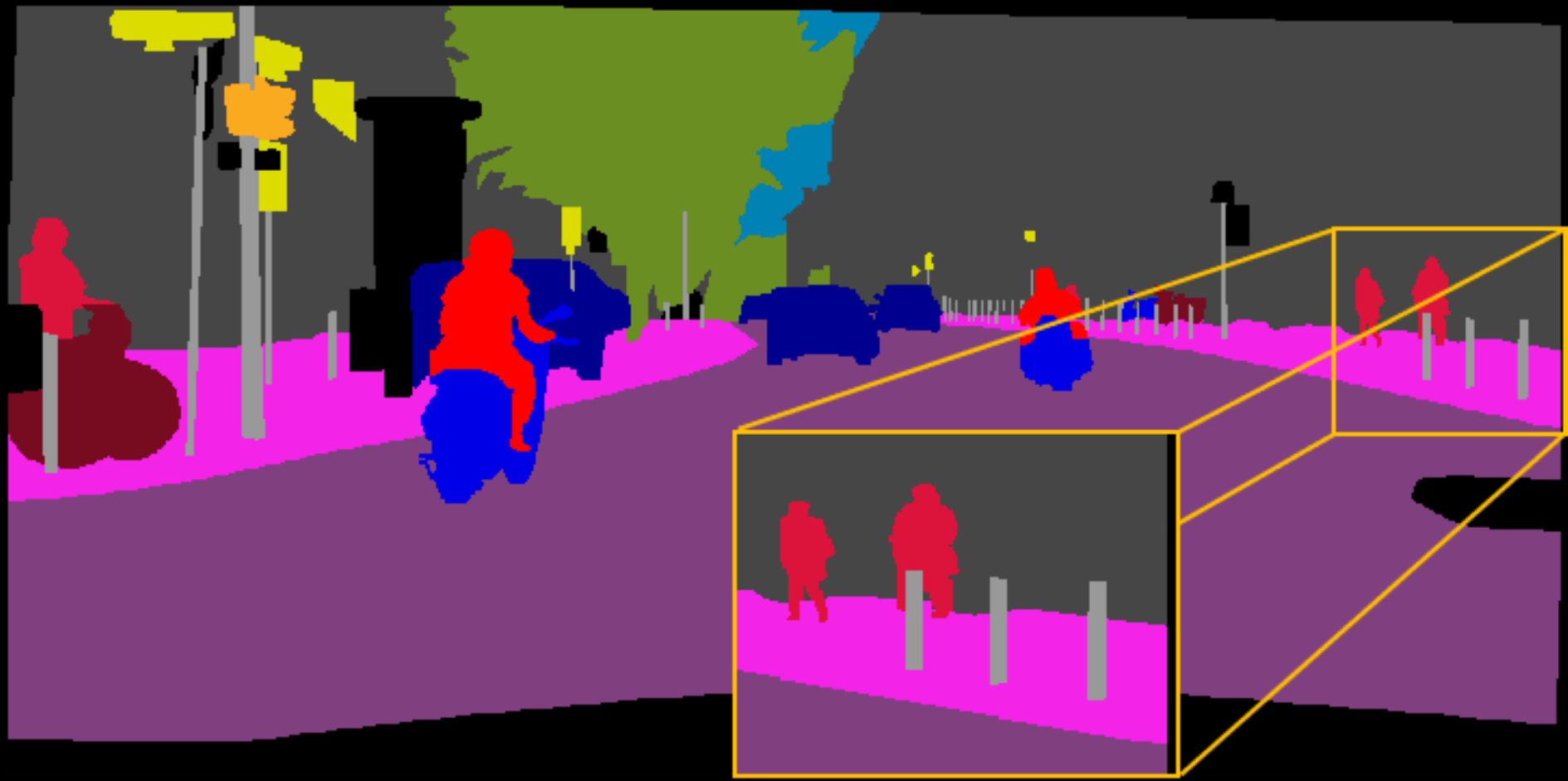}}\hspace{0.1mm}
    \subfloat{\includegraphics[width=4.2cm, height=2.5cm]{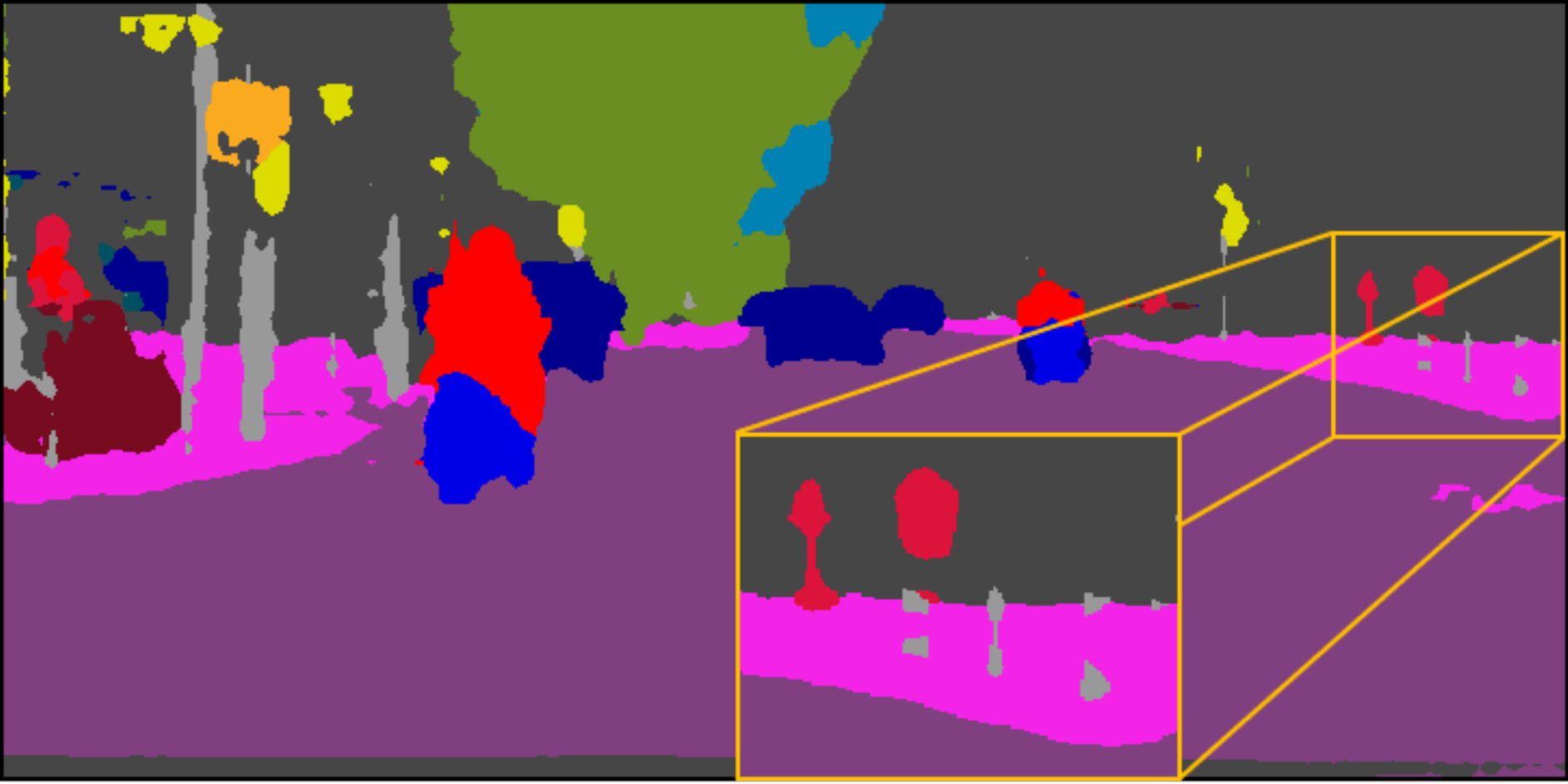}}\hspace{0.1mm}
    \subfloat{\includegraphics[width=4.2cm, height=2.5cm]{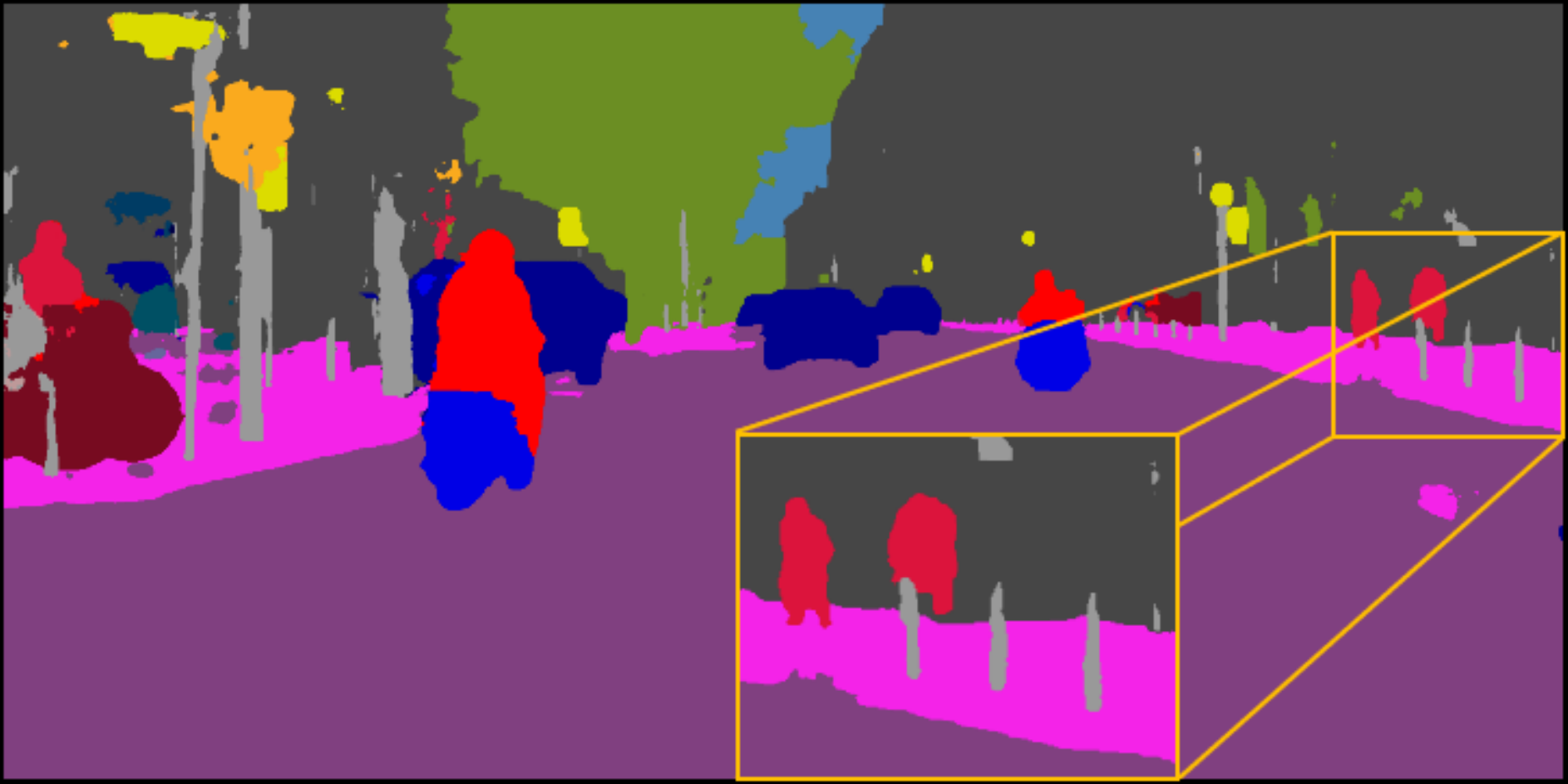}}\\\vspace{0.3mm}
    \subfloat{\includegraphics[width=4.2cm, height=2.5cm]{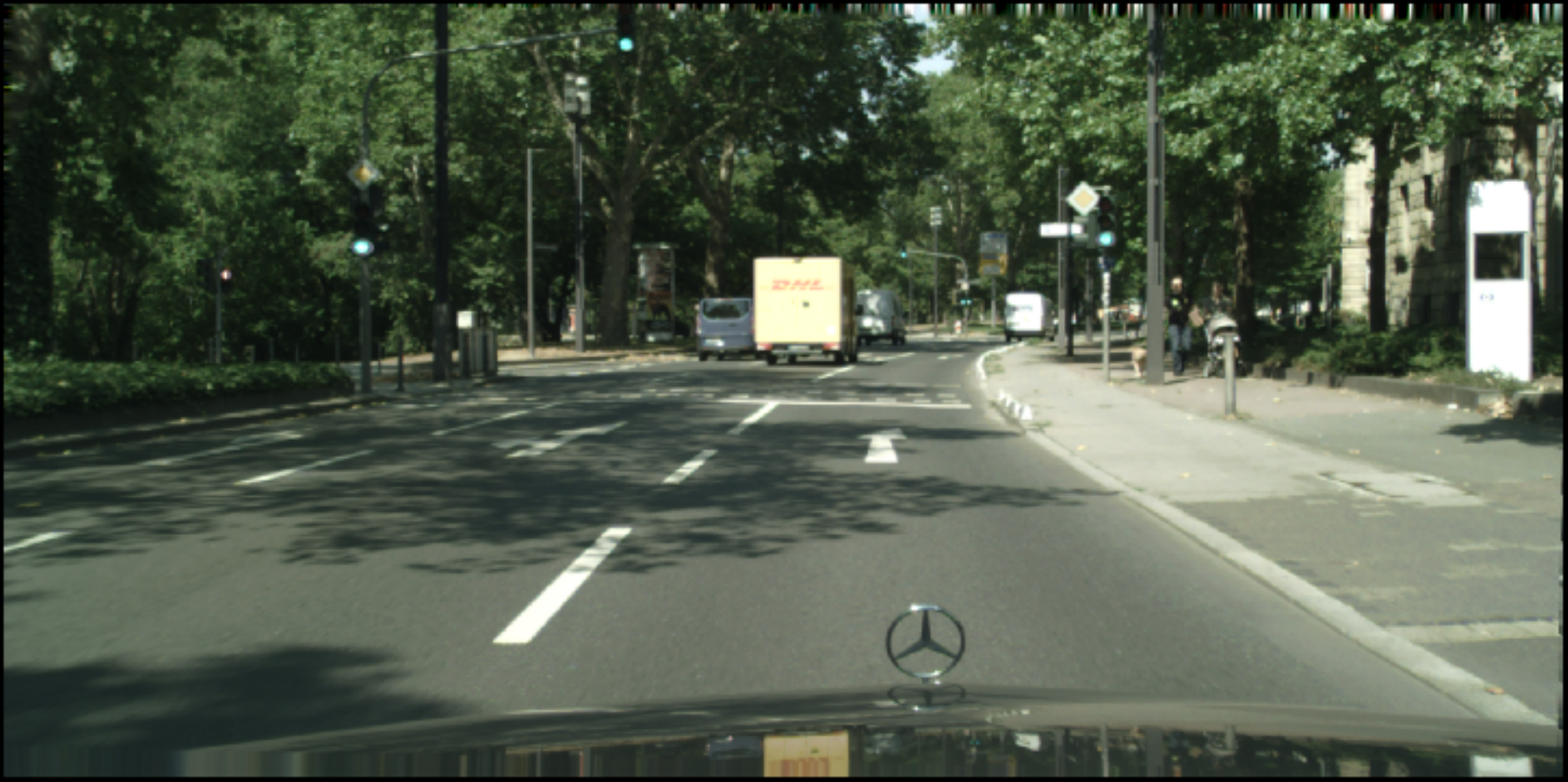}}\hspace{0.1mm}
    \subfloat{\includegraphics[width=4.2cm, height=2.5cm]{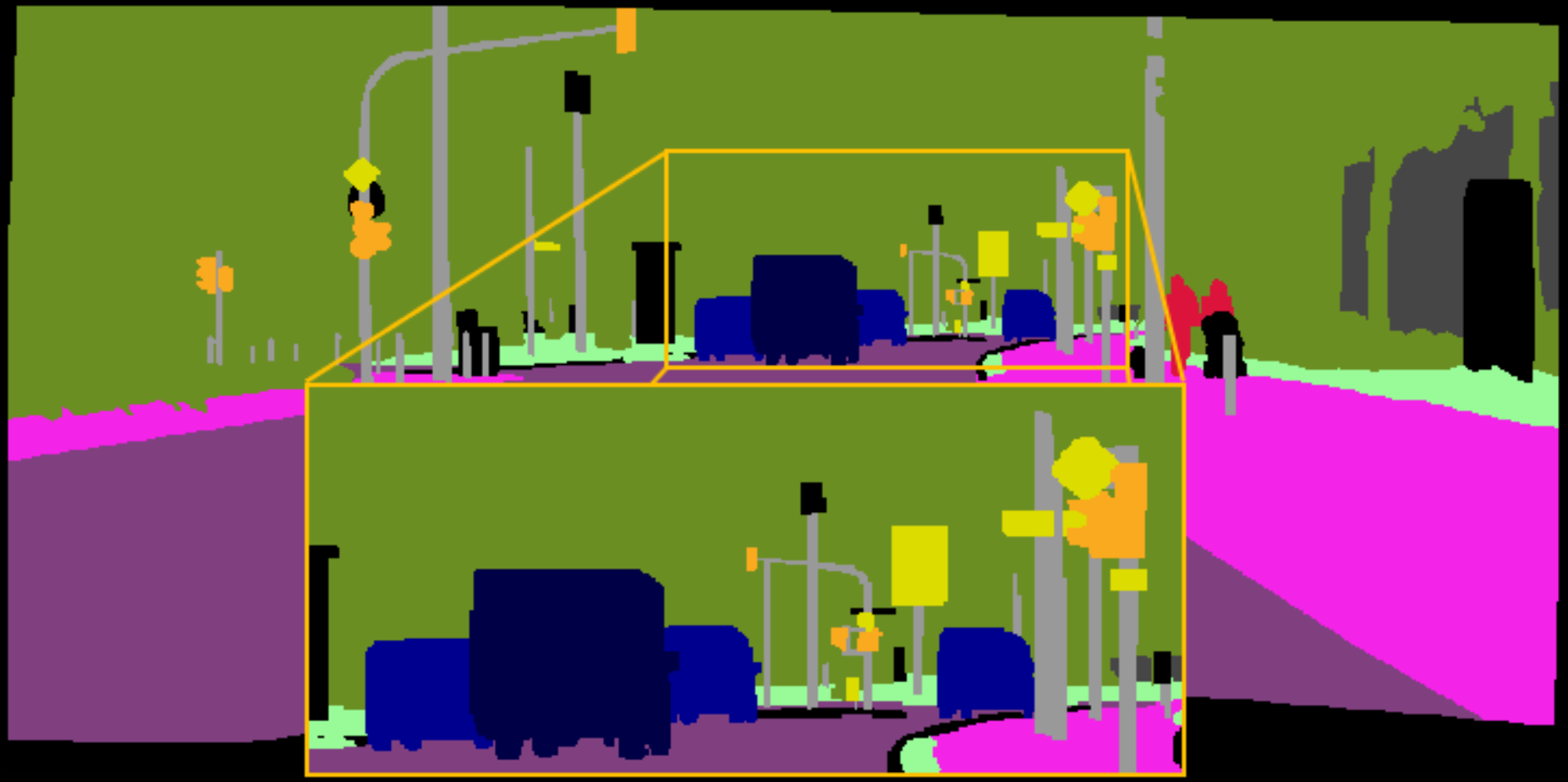}}\hspace{0.1mm}
    \subfloat{\includegraphics[width=4.2cm, height=2.5cm]{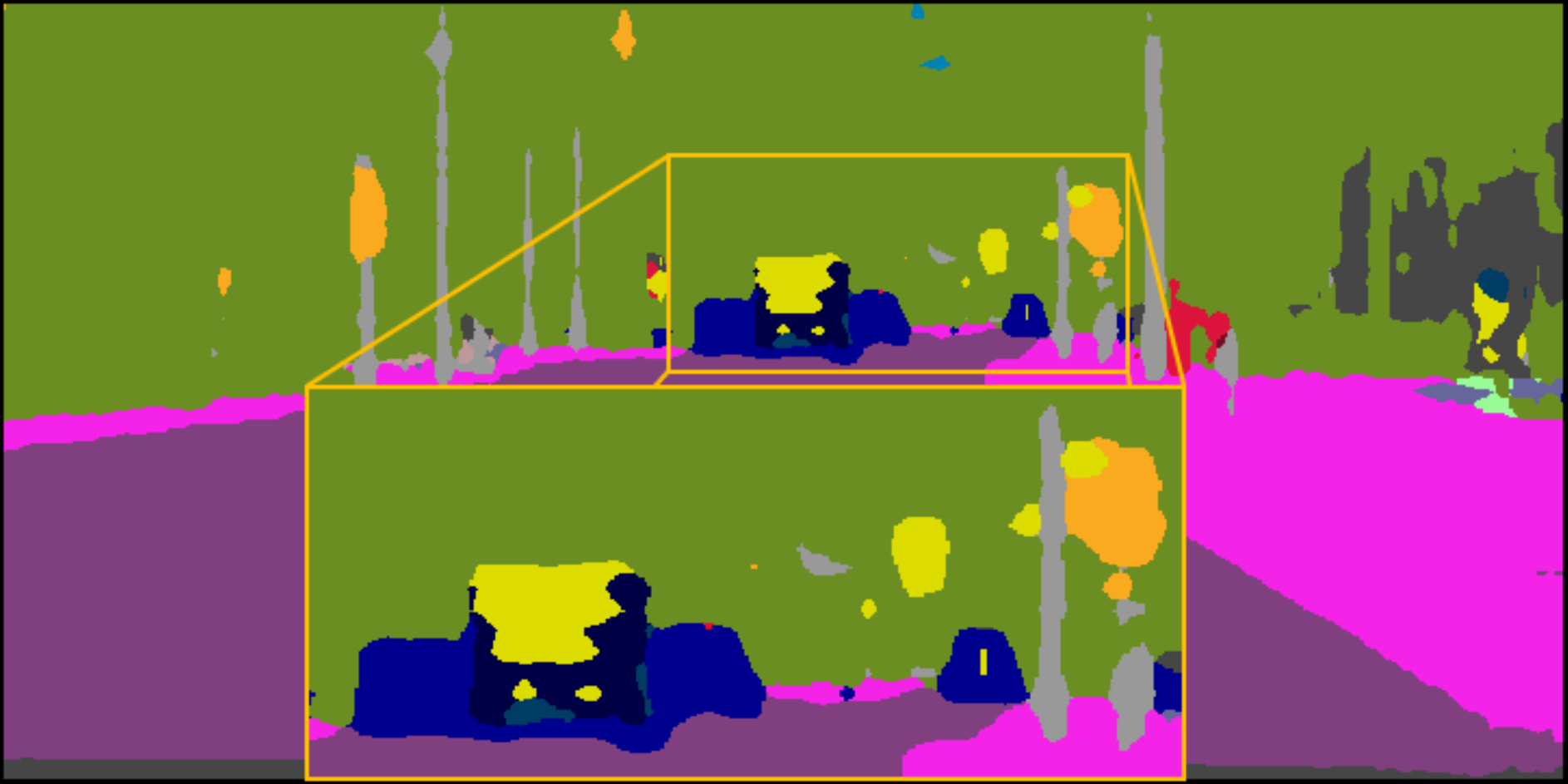}}\hspace{0.1mm}
    \subfloat{\includegraphics[width=4.2cm, height=2.5cm]{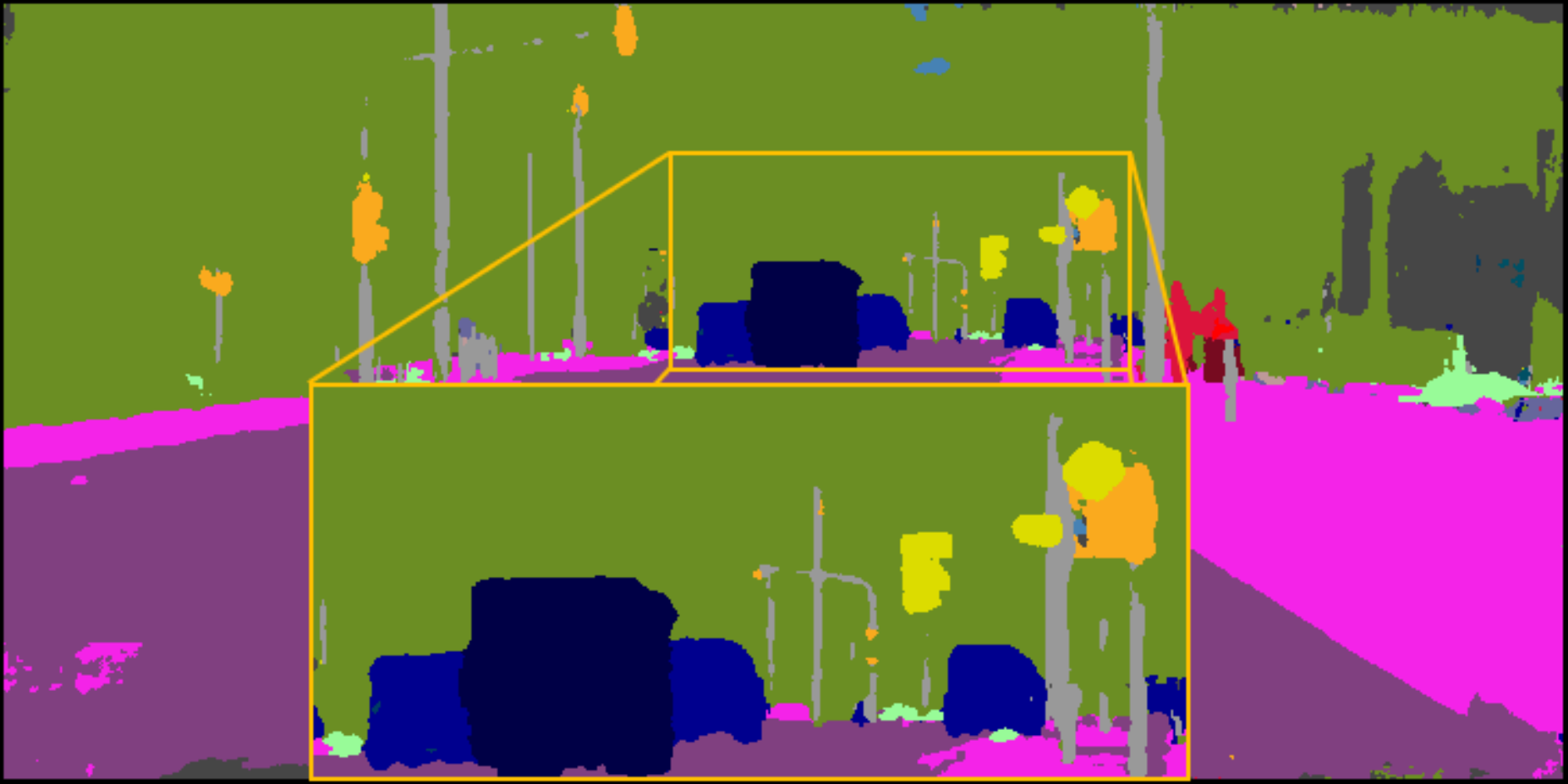}}\\\vspace{0.3mm}
    \subfloat{\includegraphics[width=4.2cm, height=2.5cm]{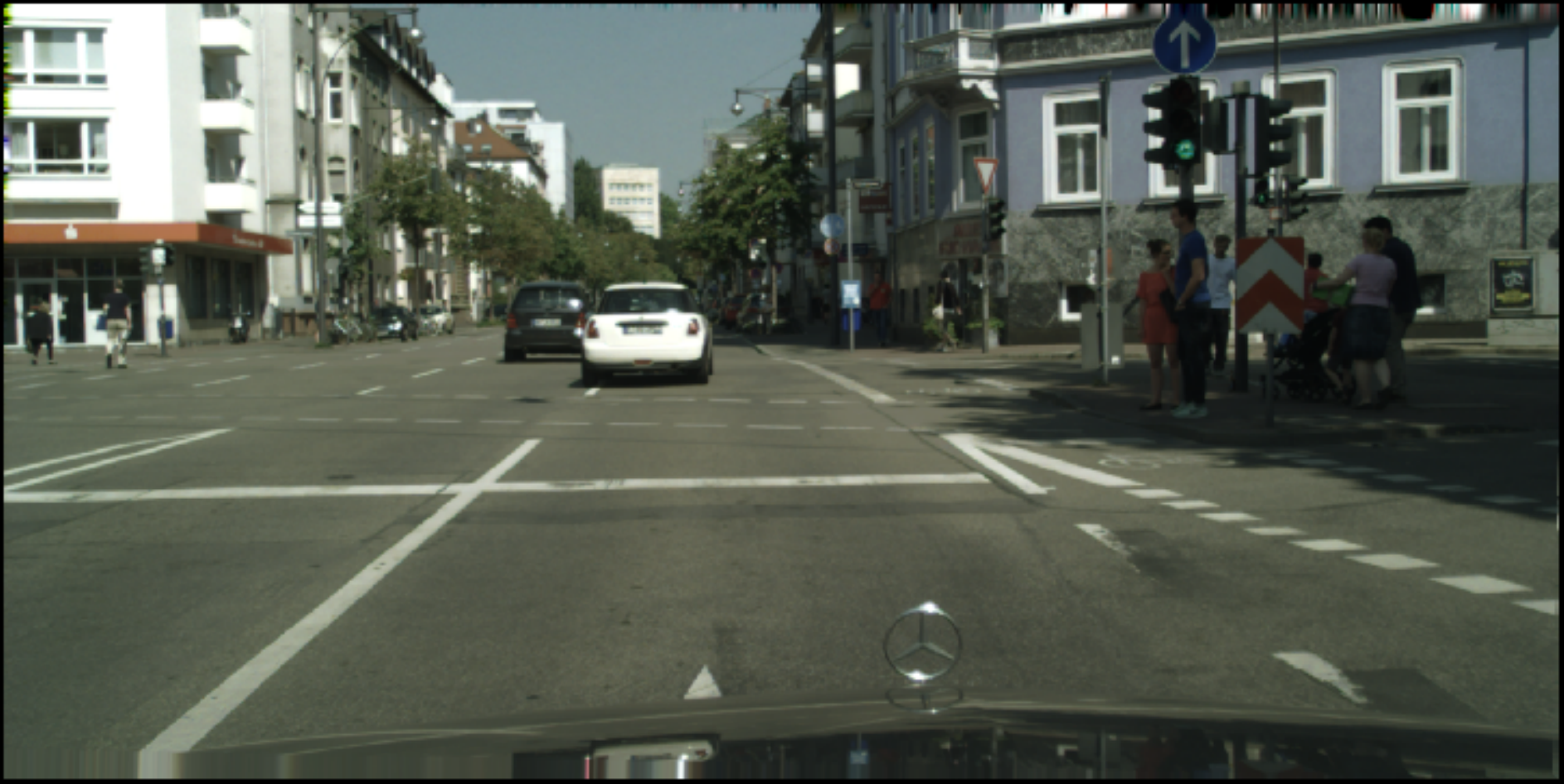}}\hspace{0.1mm}
    \subfloat{\includegraphics[width=4.2cm, height=2.5cm]{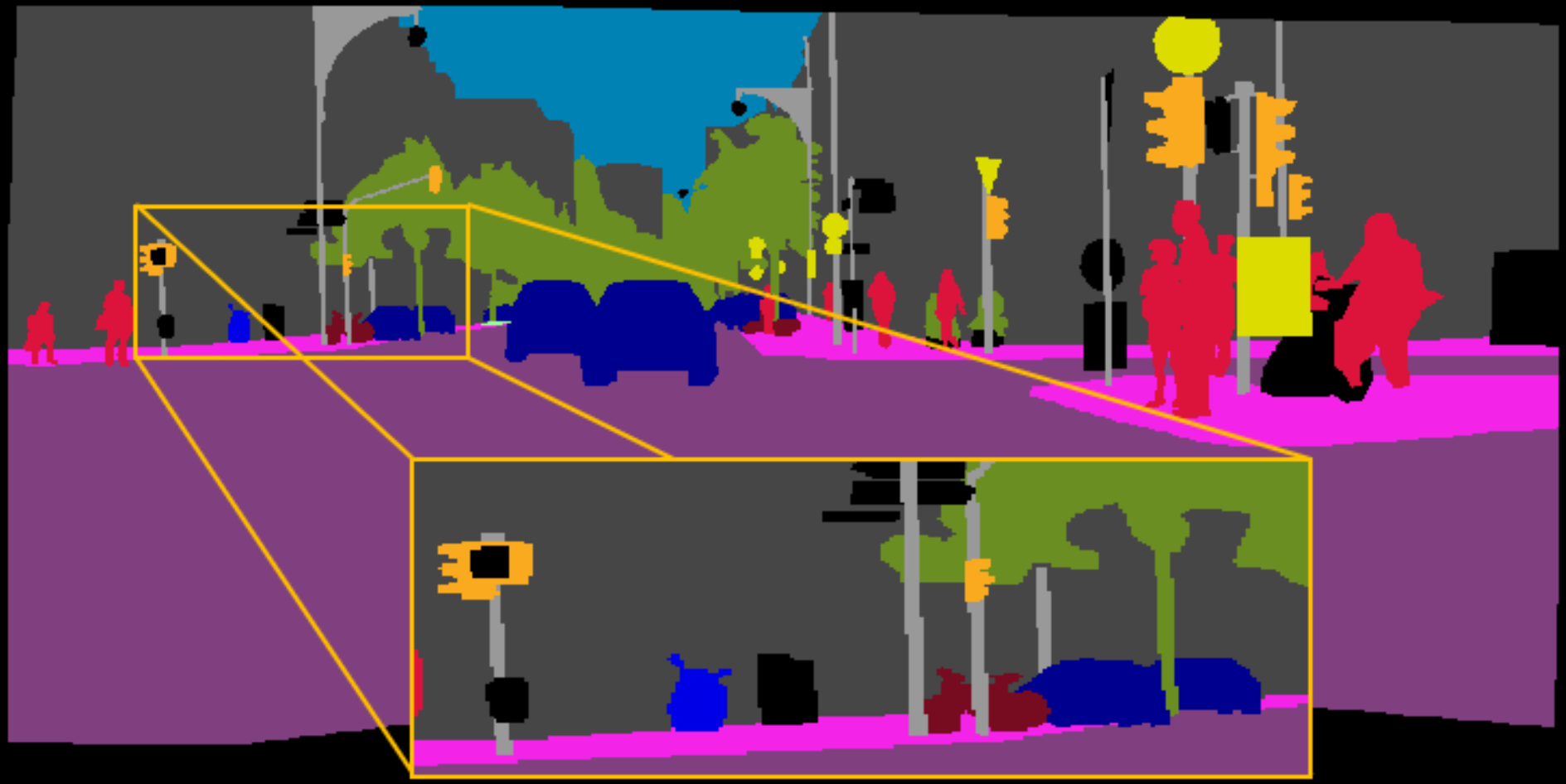}}\hspace{0.1mm}
    \subfloat{\includegraphics[width=4.2cm, height=2.5cm]{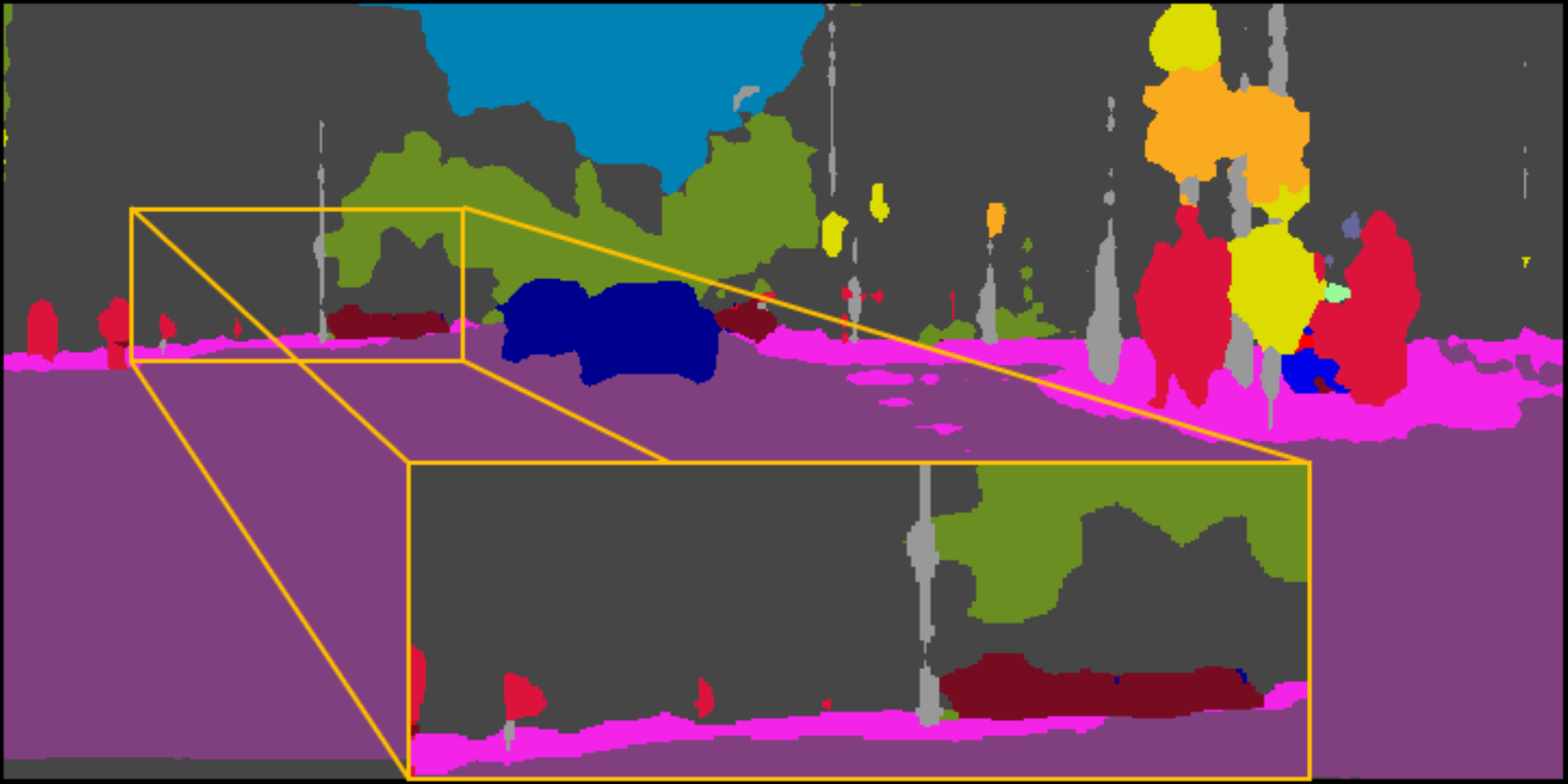}}\hspace{0.1mm}
    \subfloat{\includegraphics[width=4.2cm, height=2.5cm]{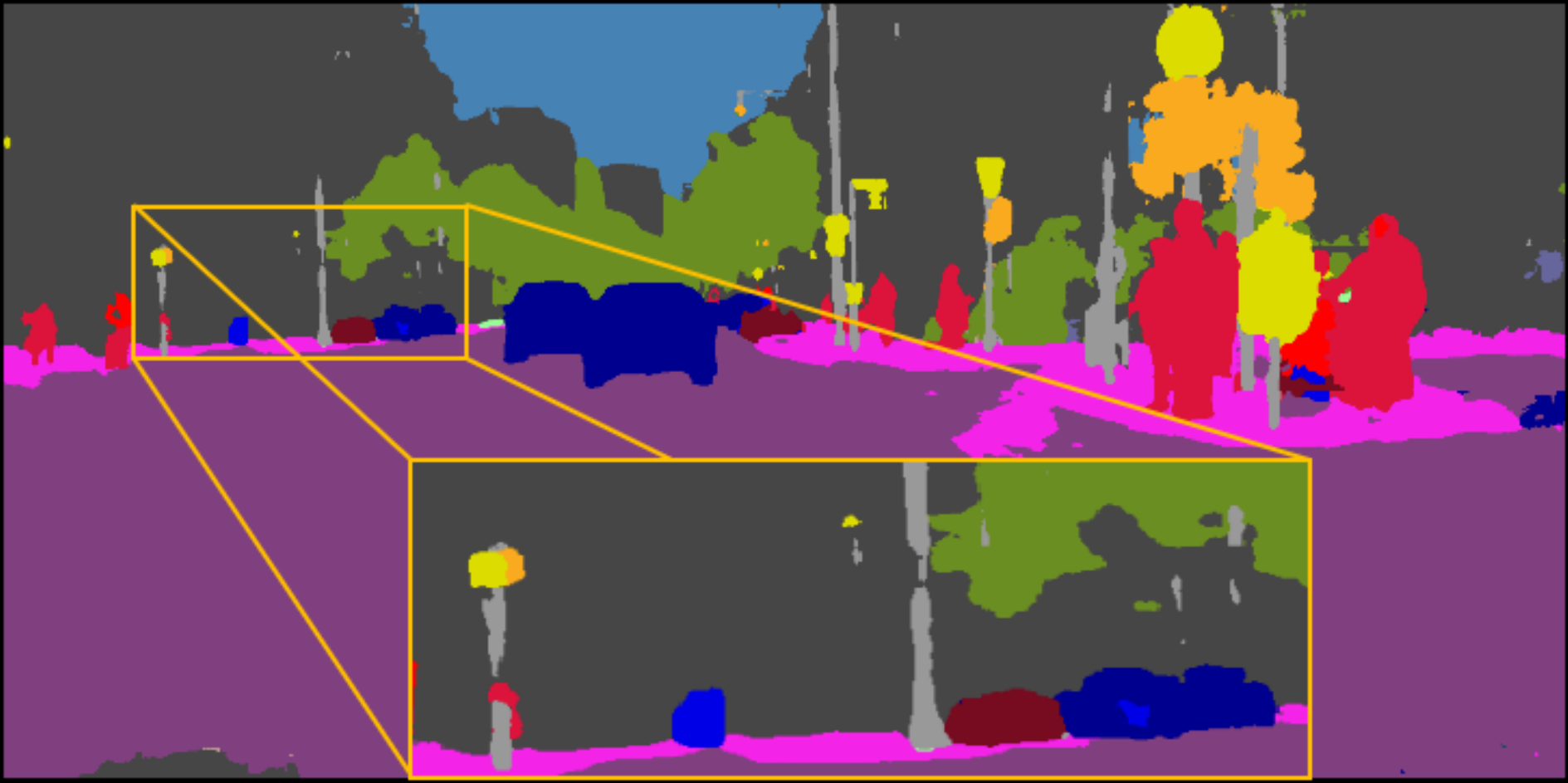}}\\\vspace{0.3mm}
    \subfloat{\includegraphics[width=4.2cm, height=2.5cm]{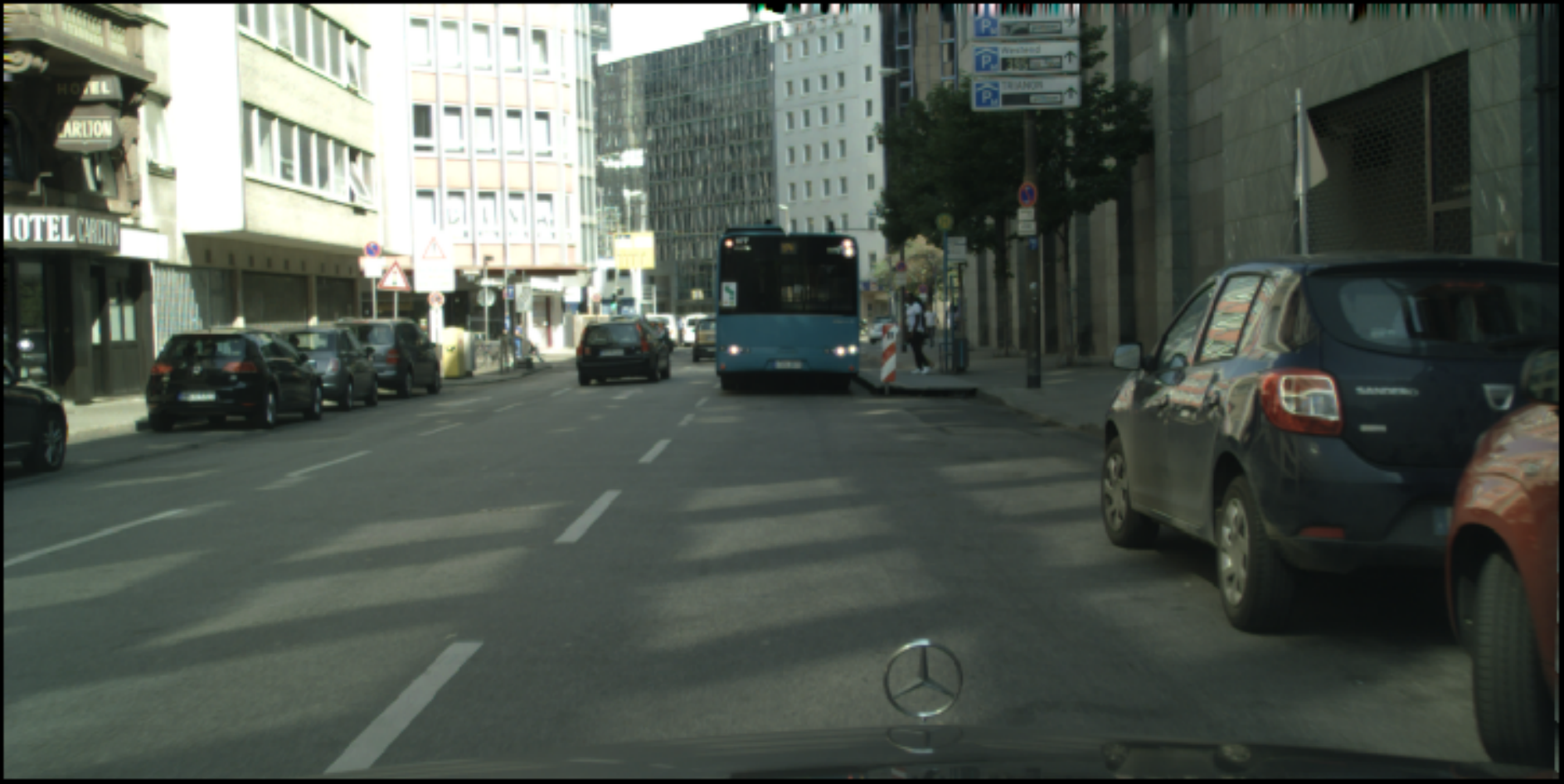}}\hspace{0.1mm}
    \subfloat{\includegraphics[width=4.2cm, height=2.5cm]{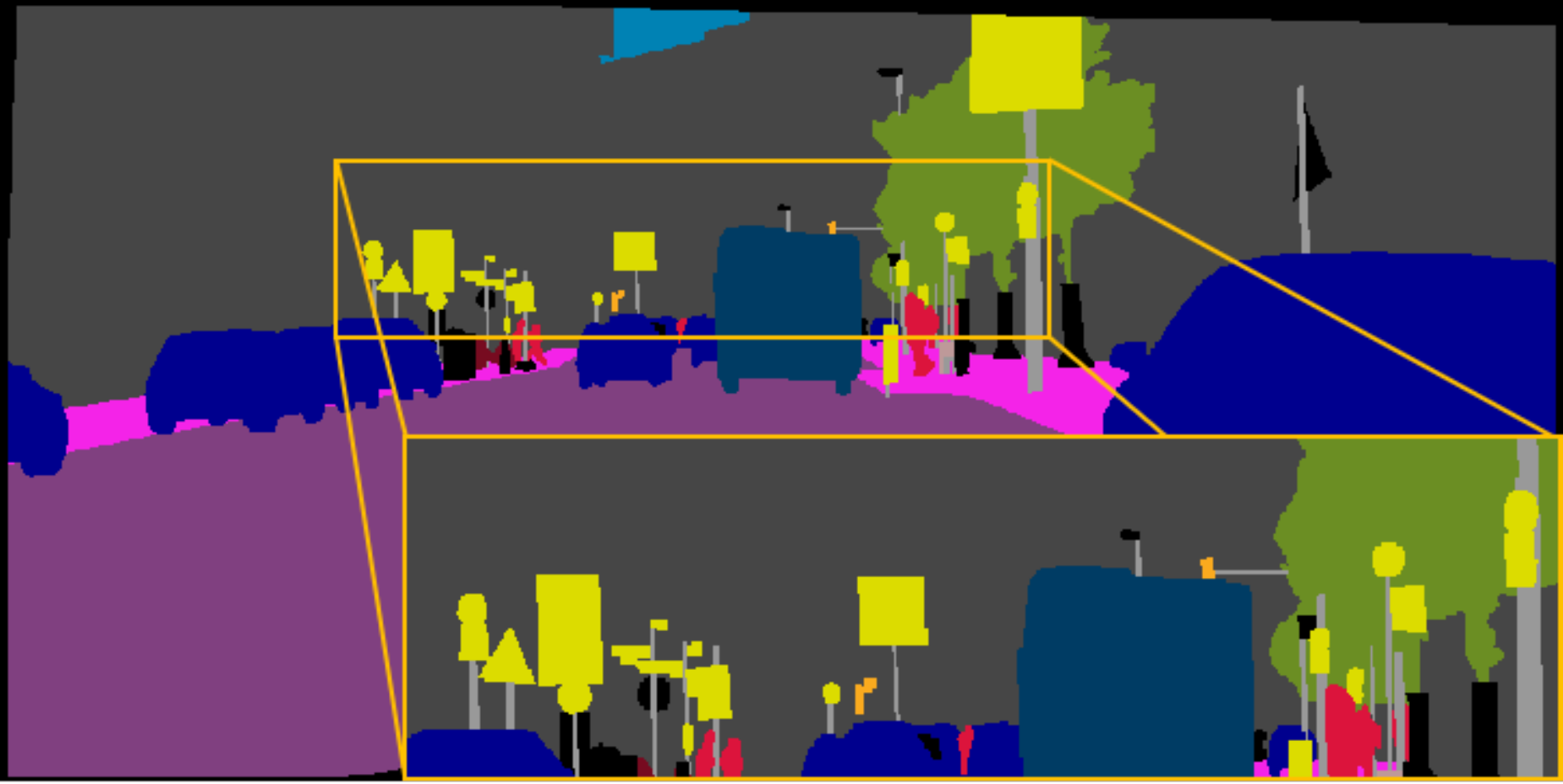}}\hspace{0.1mm}
    \subfloat{\includegraphics[width=4.2cm, height=2.5cm]{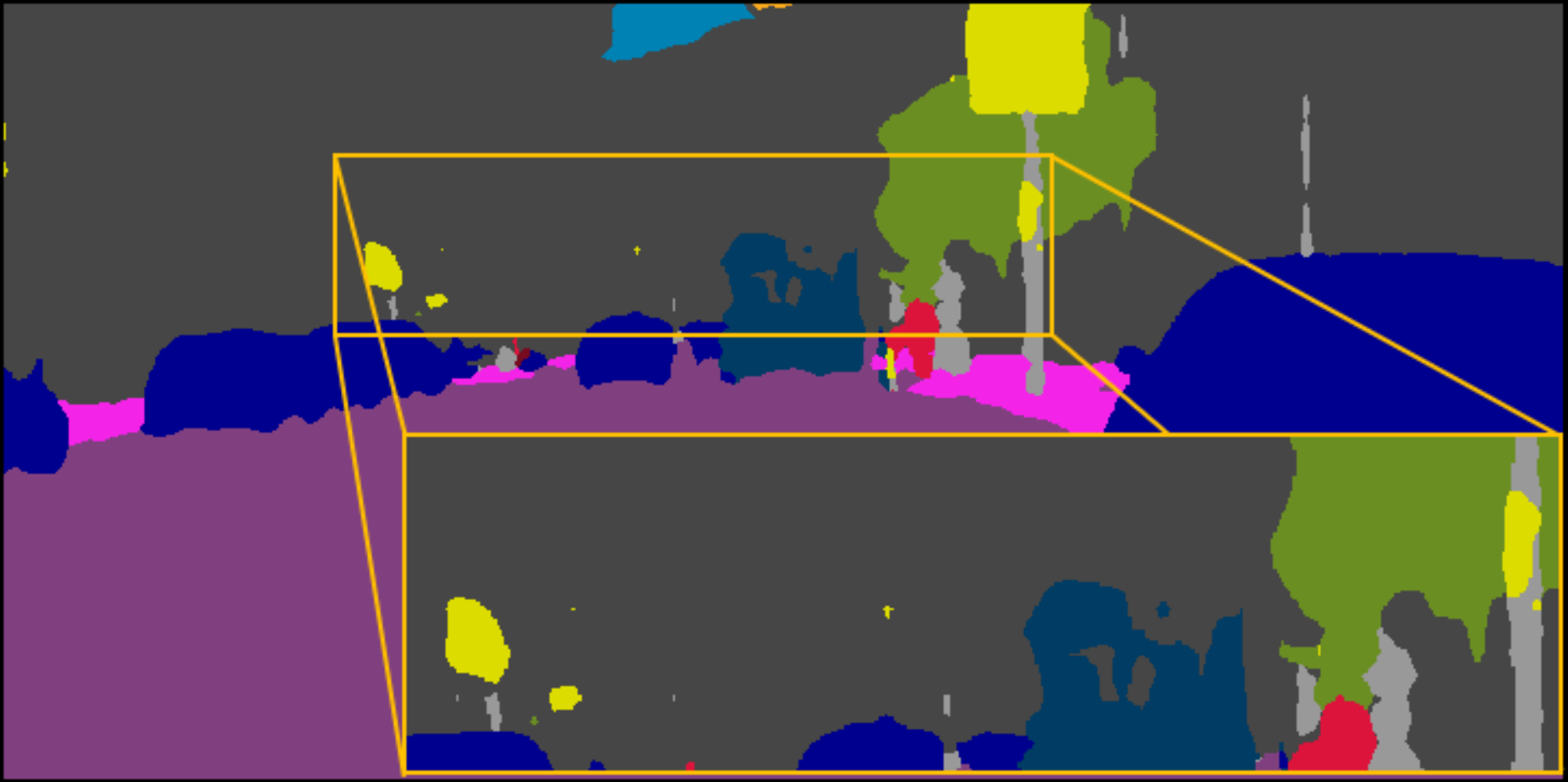}}\hspace{0.1mm}
    \subfloat{\includegraphics[width=4.2cm, height=2.5cm]{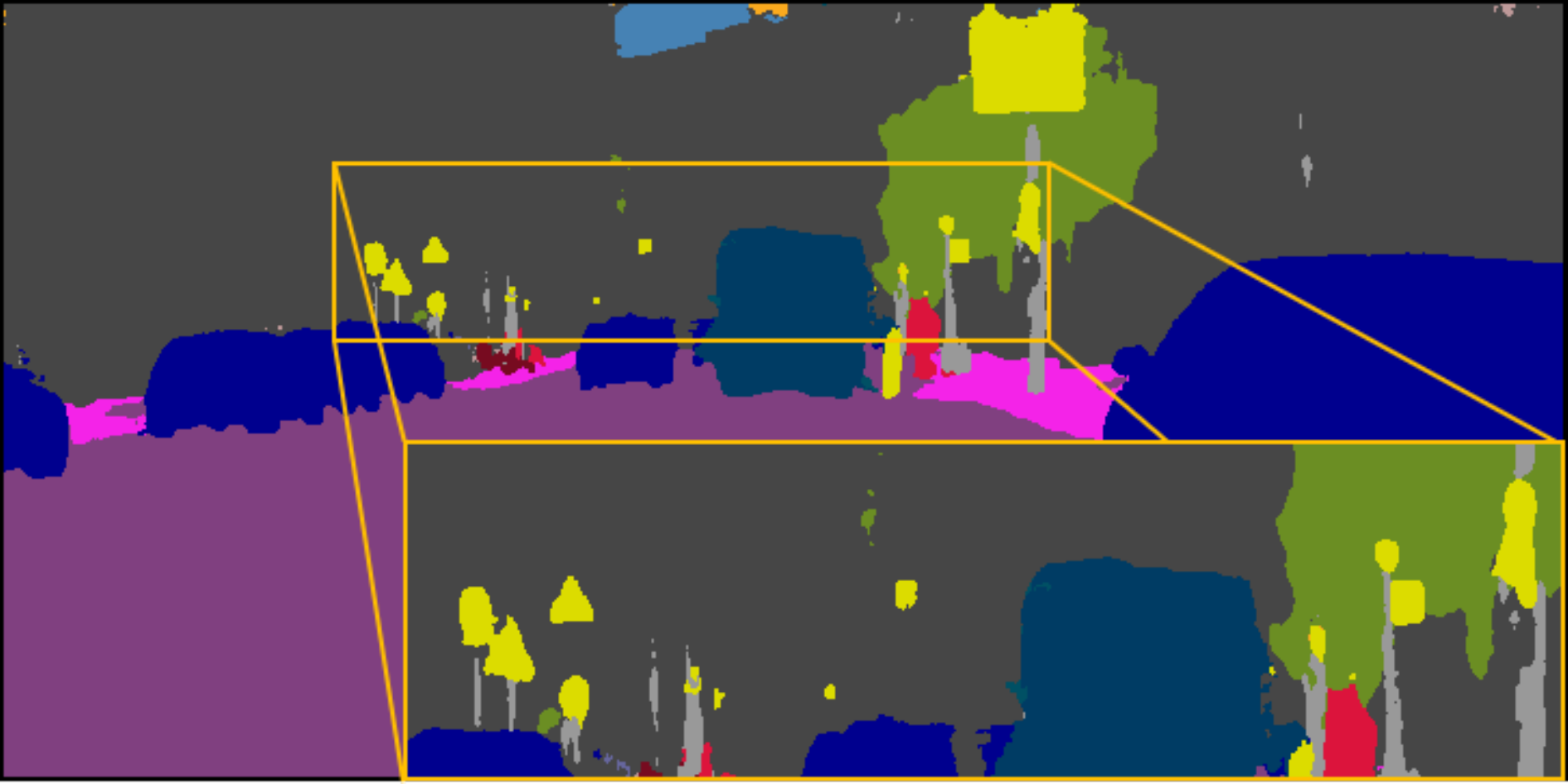}}\\\vspace{0.3mm}
    \subfloat{\includegraphics[width=4.2cm, height=2.5cm]{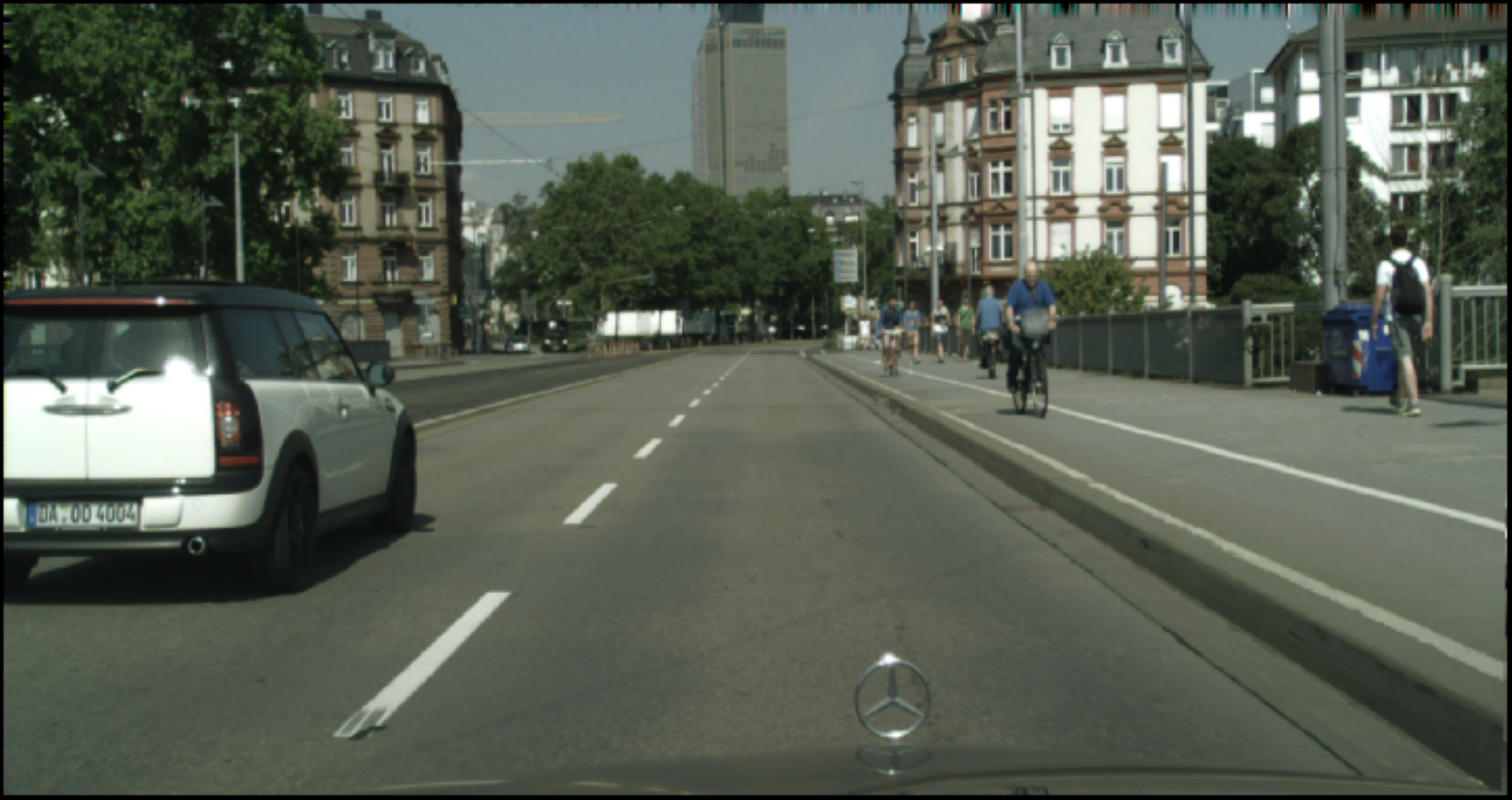}}\hspace{0.1mm}
    \subfloat{\includegraphics[width=4.2cm, height=2.5cm]{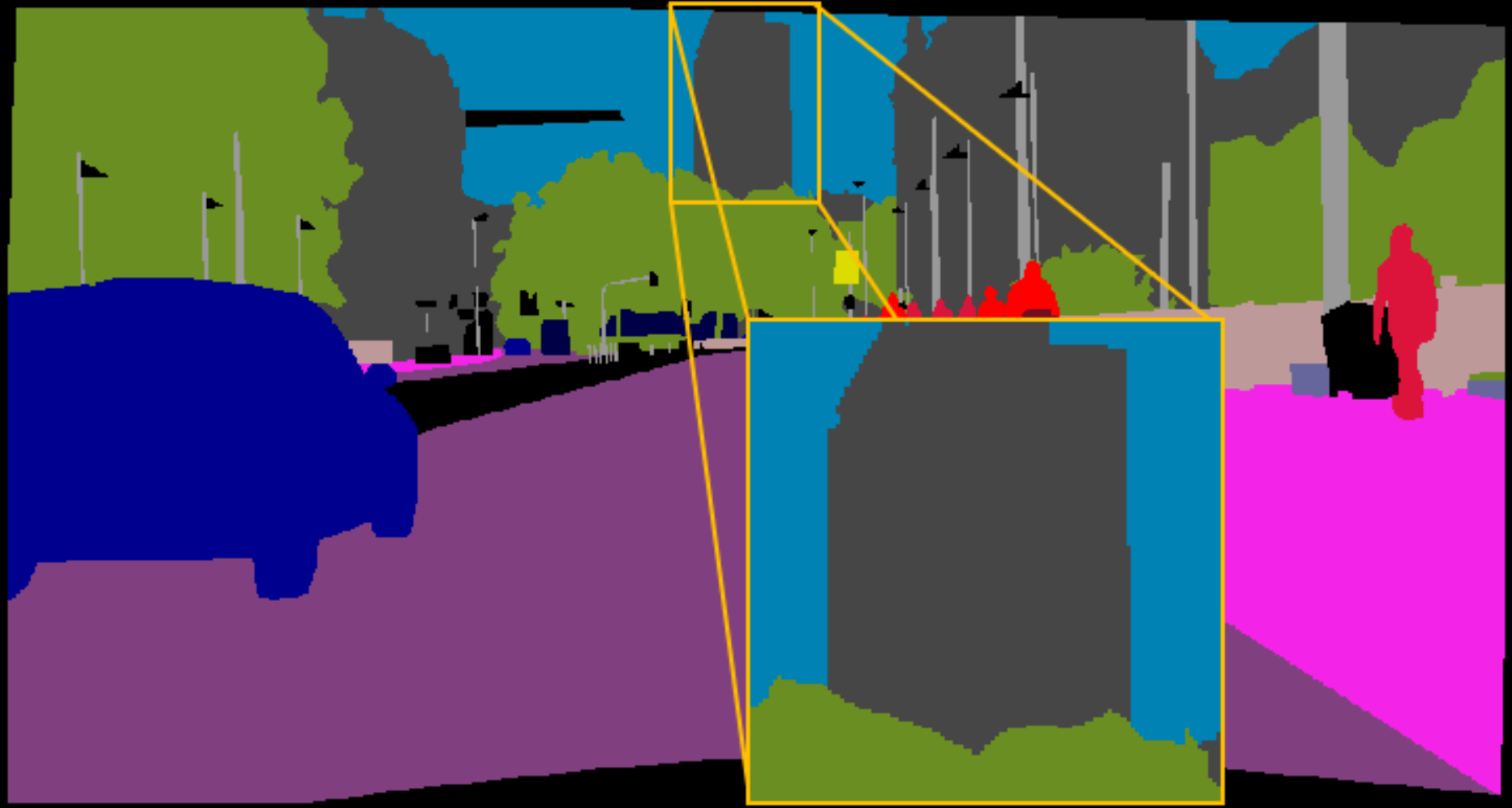}}\hspace{0.1mm}
    \subfloat{\includegraphics[width=4.2cm, height=2.5cm]{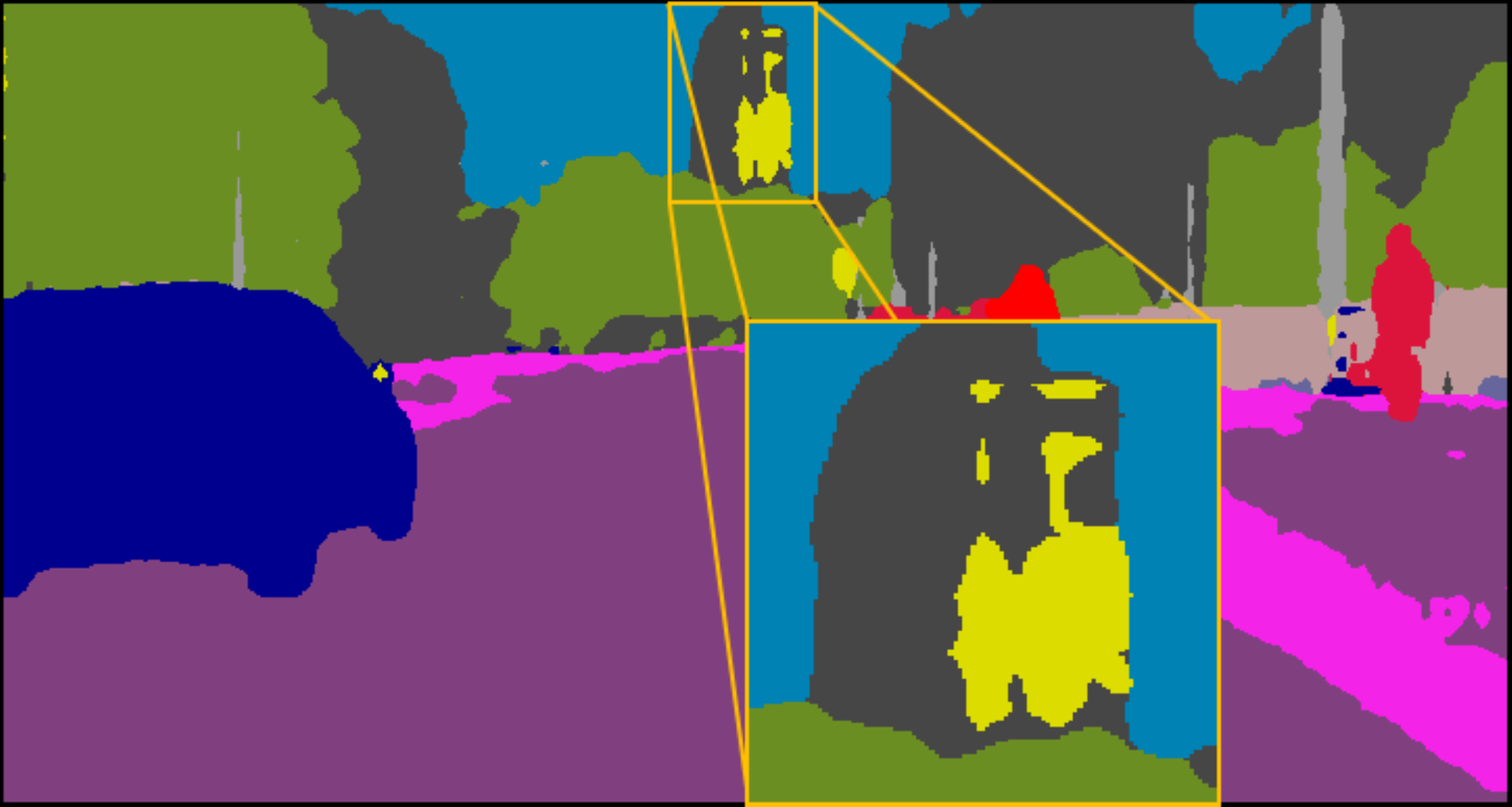}}\hspace{0.1mm}
    \subfloat{\includegraphics[width=4.2cm, height=2.5cm]{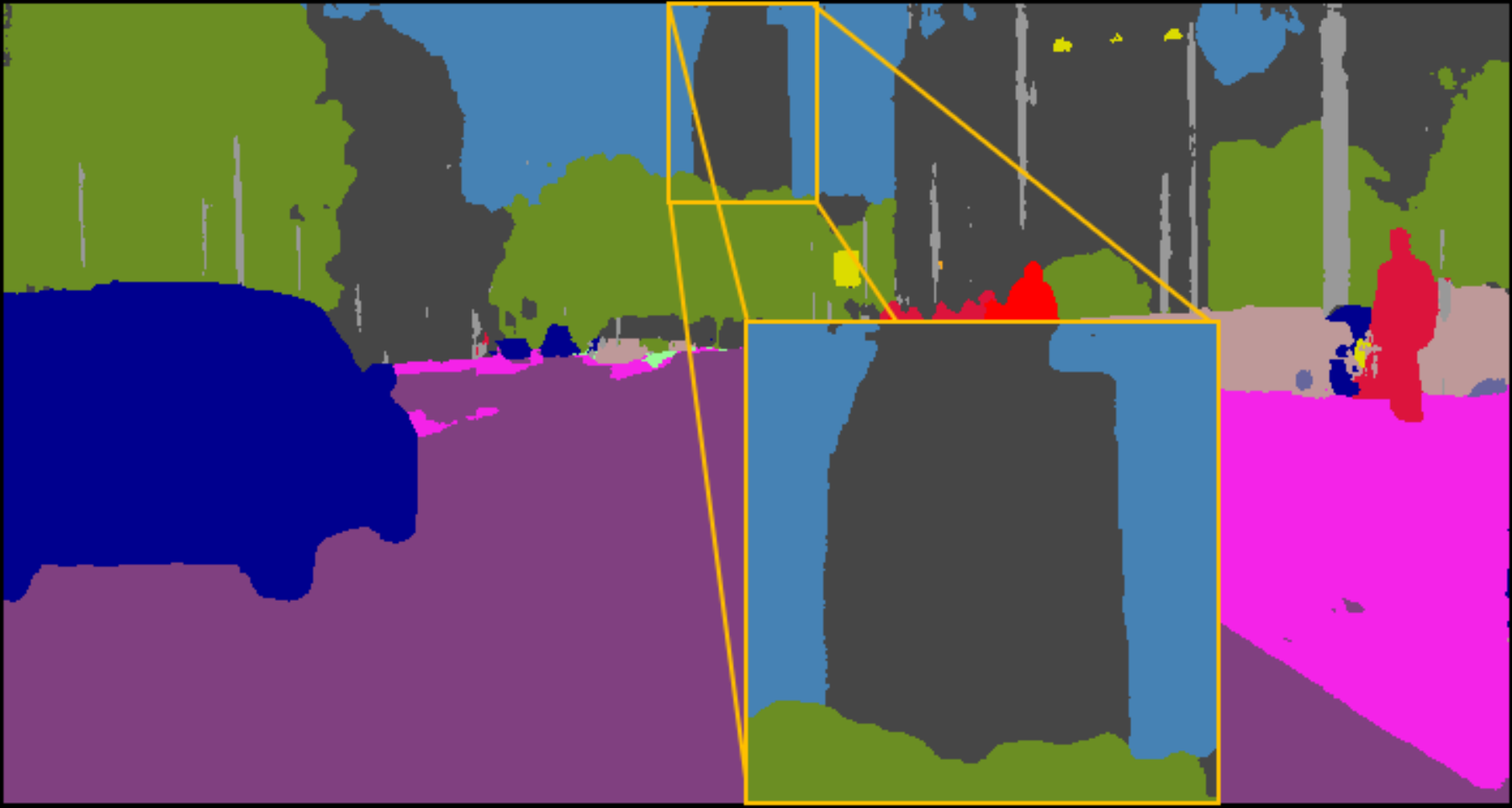}}\\\vspace{0.3mm}
    \setcounter{subfigure}{0}
    \subfloat[Input]{\includegraphics[width=4.2cm, height=2.5cm]{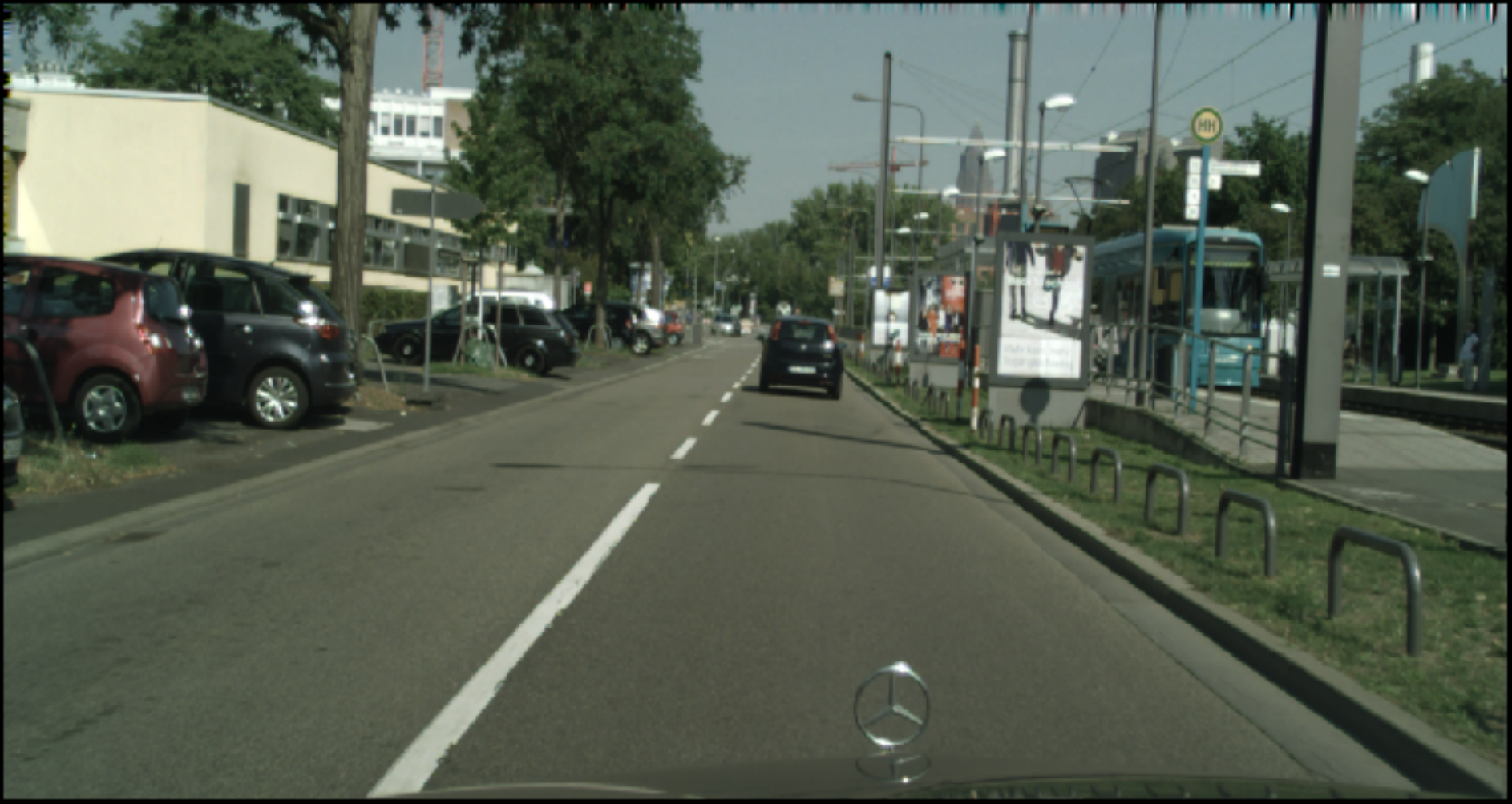}}\hspace{0.1mm}
    \subfloat[GT]{\includegraphics[width=4.2cm, height=2.5cm]{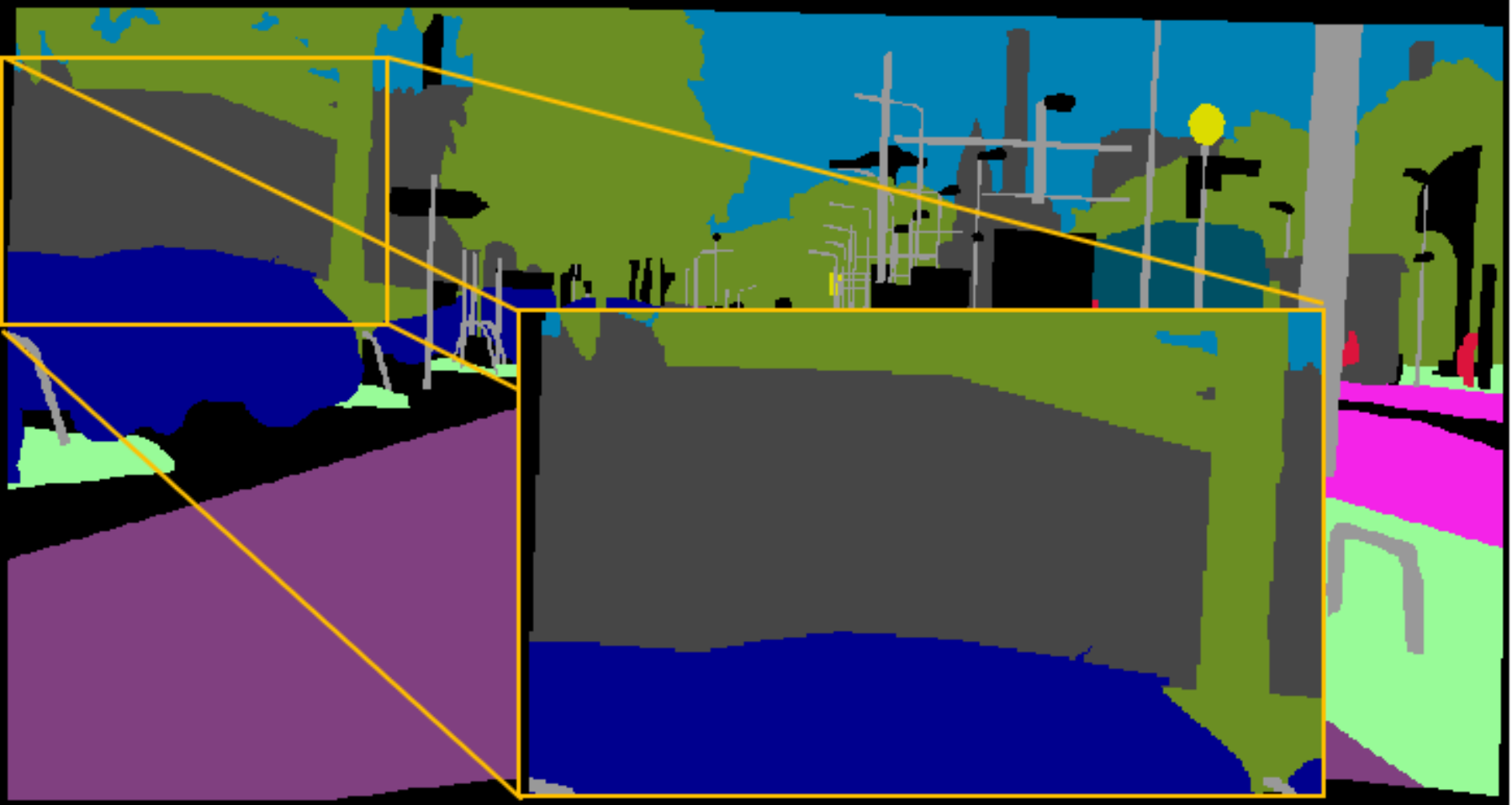}}\hspace{0.1mm}
    \subfloat[ClassMix]{\includegraphics[width=4.2cm, height=2.5cm]{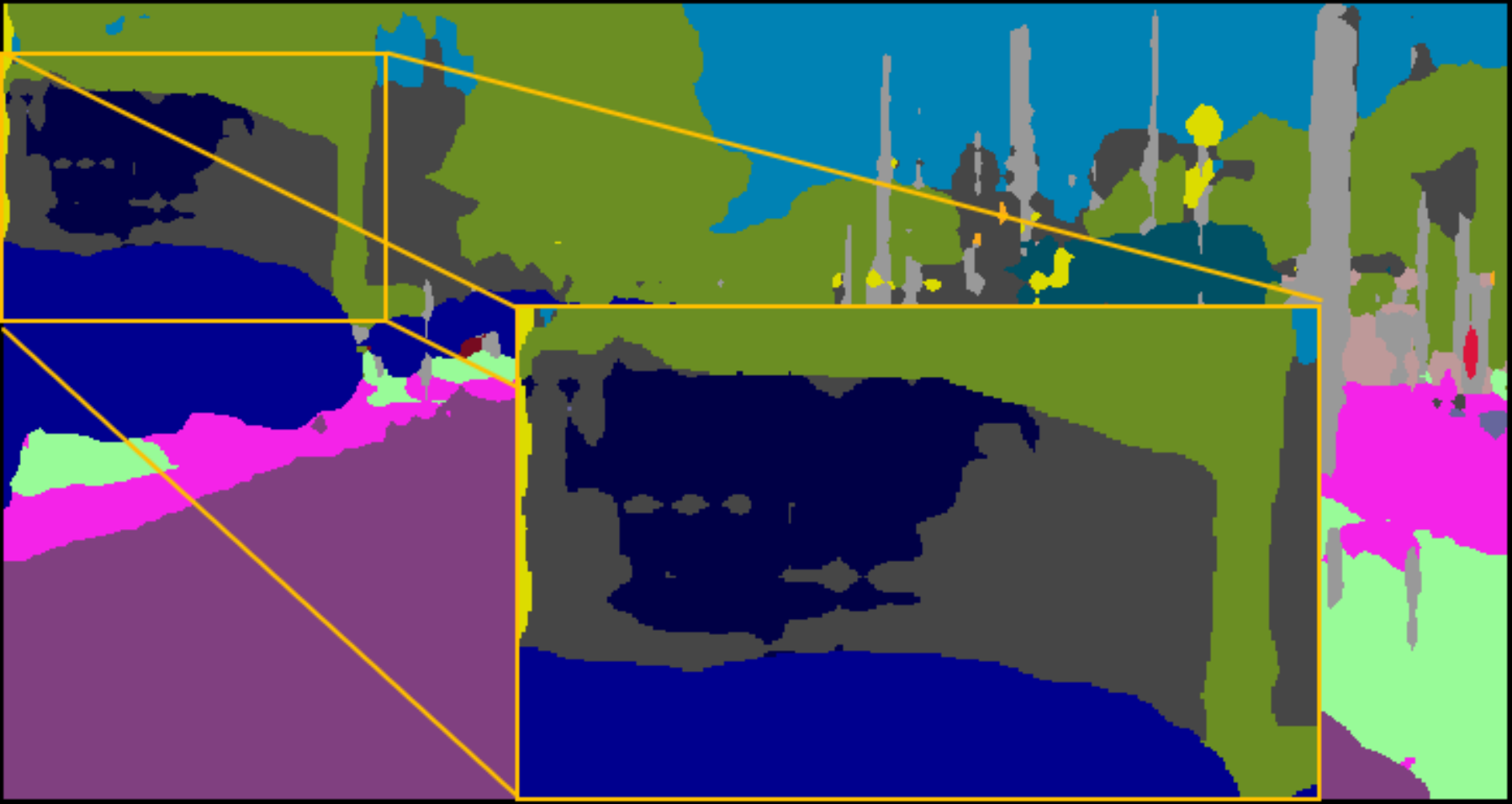}}\hspace{0.1mm}
    \subfloat[Ours]{\includegraphics[width=4.2cm, height=2.5cm]{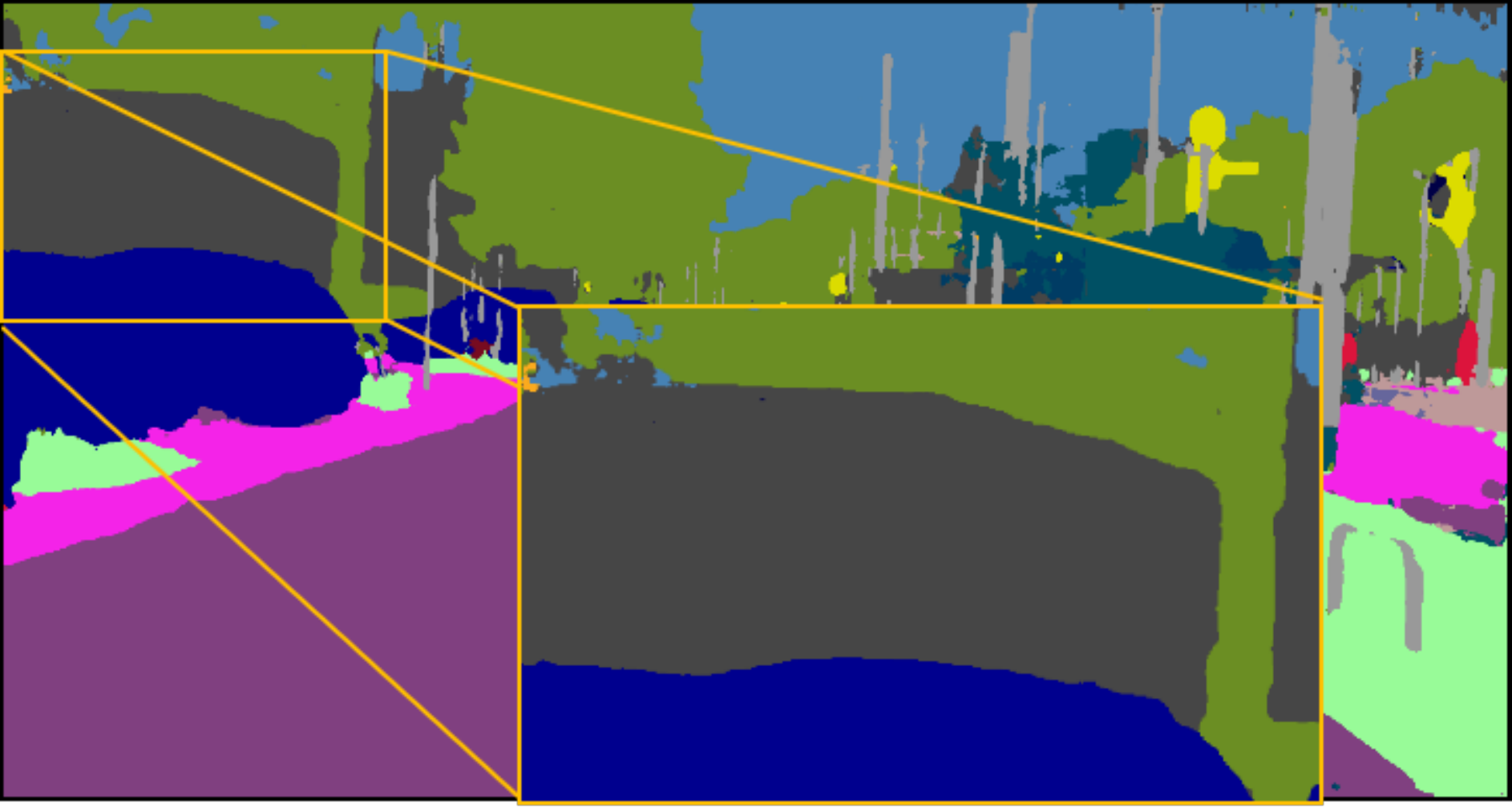}}\\
    \caption{Visual results on Cityscapes using 1/8 labeled examples. 
    The proposed semi-supervised approach generates improved results compared to ClassMix \protect\cite{olsson2021classmix}.
    (a) Input image. 
    (b) Ground-truth.
    (c) Segmentation results of ClassMix.
    (d) Segmentation results of GuidedMix-Net.}
    \label{qr_city}
    \end{figure*}

The visualization is shown in Fig. \ref{qr_city}.

\begin{quote}
\begin{small}
\bibliographystyle{aaai}
\bibliography{egbib}
\end{small}
\end{quote}

\end{document}